%% file: main.tex
\documentclass[10pt,twocolumn,letterpaper]{article}

\usepackage[pagenumbers]{cvpr} 

\input{preamble}
\definecolor{cvprblue}{rgb}{0.21,0.49,0.74}
\usepackage[pagebackref,breaklinks,colorlinks,allcolors=cvprblue]{hyperref}

\def\paperID{14143} 
\def\confName{CVPR}
\def\confYear{2026}
\def\confYear{2026}

\title{Structured Uncertainty Similarity Score (SUSS): Learning a Probabilistic, Interpretable, Perceptual Metric Between Images}

\author{Paula Seidler\\
University of Sussex, UK\\
{\tt\small ls749@sussex.ac.uk}
\and
Neill D F Campbell\\
University College London, UK\\
{\tt\small neill.campbell@ucl.ac.uk}
\and
Ivor J A Simpson\\
University of Sussex, UK\\
{\tt\small i.simpson@sussex.ac.uk}
}

\begin{document}
\maketitle
\input{sec/0_abstract}
\input{sec/introduction}

\input{sec/related_work}

\input{sec/methods}

\input{sec/experiments}    
\input{sec/results}
\input{sec/discussion_new}

\input{sec/conclusion}

{
    \small
    \bibliographystyle{ieeenat_fullname}
    \bibliography{main}
}

\appendix
\include{sec/appendix}
\end{document}

%% file: sec/0_abstract.tex
\begin{abstract}
Perceptual similarity scores that align with human vision are critical for both training and evaluating computer vision models. Deep perceptual losses, such as LPIPS, achieve good alignment but rely on complex, highly non-linear discriminative features with unknown invariances, while hand-crafted measures like SSIM are interpretable but miss key perceptual properties.
We introduce the \textbf{Structured Uncertainty Similarity Score (SUSS)}; it models each image through a set of perceptual components, each represented by a structured multivariate Normal distribution. These are trained in a generative, self-supervised manner to assign high likelihood to human-imperceptible augmentations. The final score is a weighted sum of component log-probabilities with weights learned from human perceptual datasets. Unlike feature-based methods, SUSS learns image-specific linear transformations of residuals in pixel space, enabling transparent inspection through decorrelated residuals and sampling.
SUSS aligns closely with human perceptual judgments, shows strong perceptual calibration across diverse distortion types, and provides localized, interpretable explanations of its similarity assessments. We further demonstrate stable optimization behavior and competitive performance when using SUSS as a perceptual loss for downstream imaging tasks.
\end{abstract}

%% file: sec/introduction.tex
\section{Introduction}
\begin{figure}
\centering
\includegraphics[width=1.0\linewidth]{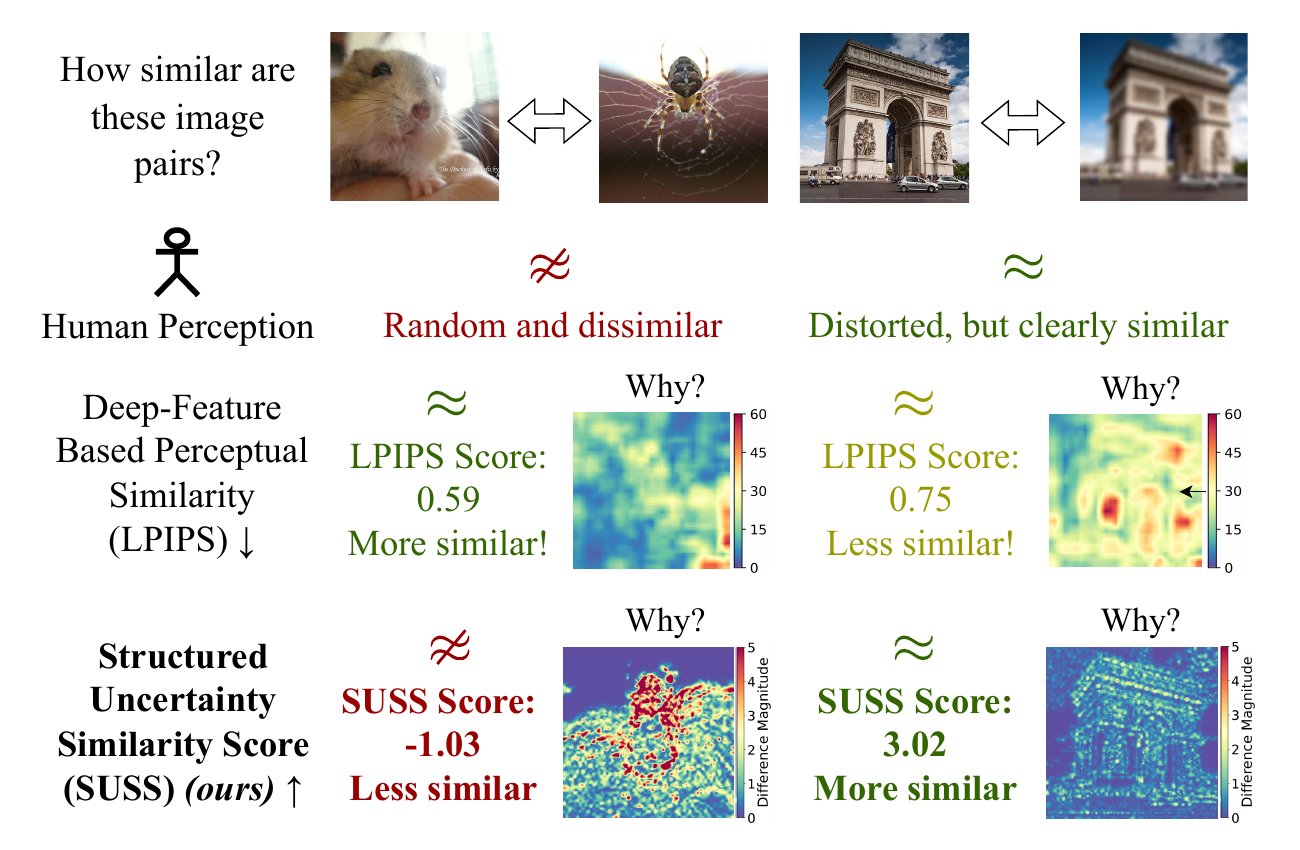}
    \caption{
    For pairs of images (top row) humans can determine the level of similarity, and offer post-hoc explanations. Deep learned perceptual metrics, such as LPIPS (third row), use features that are a complex non-linear transformation of the images and contain many invariances; making the rationale for a result hard to explain. This is illustrated by examining the magnitude of the difference of feature maps for the two example image pairs: we observe diffuse minor differences between two dissimilar images, and highly localized distinctions between closely related images. 
    Our proposed method, SUSS (bottom row) provides an interpretable pixelwise map that is based on linearly transformed residuals between the two images, these directly highlights only the differences that are considered perceivable reducing the impact of minor alterations.}
    \label{fig:problem}  \vspace{-10pt}
\end{figure}

Evaluating the similarity between pairs of images is a fundamental task in computer vision where perceptual image metrics are designed to quantify this in a way that closely aligns with human vision. These metrics form the foundation for computer vision models, both as loss objectives to guide training and fitting, or for evaluation. 

Traditional, handcrafted approaches, like the Structural Similarity Index (SSIM)~\cite{ssim} and derivatives~\cite{mssim}, capture specific perceptual aspects, but they apply fixed statistical models uniformly across all image regions and do not adapt to image content. As a result, they struggle to capture meaningful structural variations, invariance to affine transformations, and piecewise similarities, all of which are crucial for perceptual losses in downstream tasks.

In this work we make a distinction between methods that ascertain whether images contain similar content (which we term \emph{``semantic similarity''}) as opposed to images portraying the same content but with some other modification (e.g.~distortions or sensor and quantization noise); we consider the latter and refer to it as \emph{perceptual similarity}.

The shortcomings of handcrafted metrics motivate the development of (deep) learning-based perceptual losses that achieve alignment with human perception by relying on complex non-linear neural networks to extract features.
%
Importantly, these models are often based on \emph{discriminative} (e.g.~classification) models (e.g.~the Learned Perceptual Image Patch Similarity (LPIPS)~\cite{deepf}) that are trained to learn many invariances in the underlying network, some of which may not match with the subtle perceptual changes we are interested in (e.g.~spatial or textural invariances) that we hypothesize preferentially targets semantic similarity.
Identifying an explanation for a particular result is difficult, as illustrated in Fig.~\ref{fig:problem}, where the complexity of the feature representations can lead to issues of trustworthiness, but also raises questions regarding the efficacy of the metric when faced with all possible image combinations. 


To address this, we propose a new, data-driven, perceptual similarity score, SUSS, where we base the self-supervision approach on a \emph{deep, probabilistic generative model}, namely a Structured Uncertainty Prediction Network (SUPN)~\cite{ivor1}.
This model efficiently predicts a densely correlated multivariate Normal distribution over pixel space (via a sparsely structured precision matrix). 
Our approach confers four key advantages:
\begin{enumerate}[(i)]

\item Our generative formulation ensures that we avoid injecting unknown invariances that are encoded in a discriminative model by construction.

\item We build the score using a weighted sum of log Normal densities and thus we encode the score as a Mahanalobis distance; this provides a range of desirable ``norm-like'' properties that make the score robust and ideal for use as a loss in downstream tasks as well as a similarity measure.

\item The closed form Gaussian distribution allow us to \emph{introspect} and \emph{explain} the score by drawing samples and visualizing residuals all in image space (as this is the space of the metric) as opposed to the black-box nature of recent deep feature based metrics.

\item The formulation combines multi-scale structure, color and spatial uncertainty and its trained to align with human perception; we demonstrate that it obtains superior calibration in consistent quantification of similar perceptual differences across a range of different modalities (such as deformation, noise, etc\dots).

\end{enumerate}

We evaluate the efficacy of our approach through comparisons on perceptual tasks as well as quantitative evaluation of \emph{perceptual calibration}. We illustrate introspection of our score through analysis of the visualized residuals and samples from the learned perceptual distributions. Finally, we demonstrate its suitability for downstream applications through the use of SUSS as a loss term in imaging problems.



\begin{figure*}[h]
    \centering
    \includegraphics[width=0.7\textwidth]{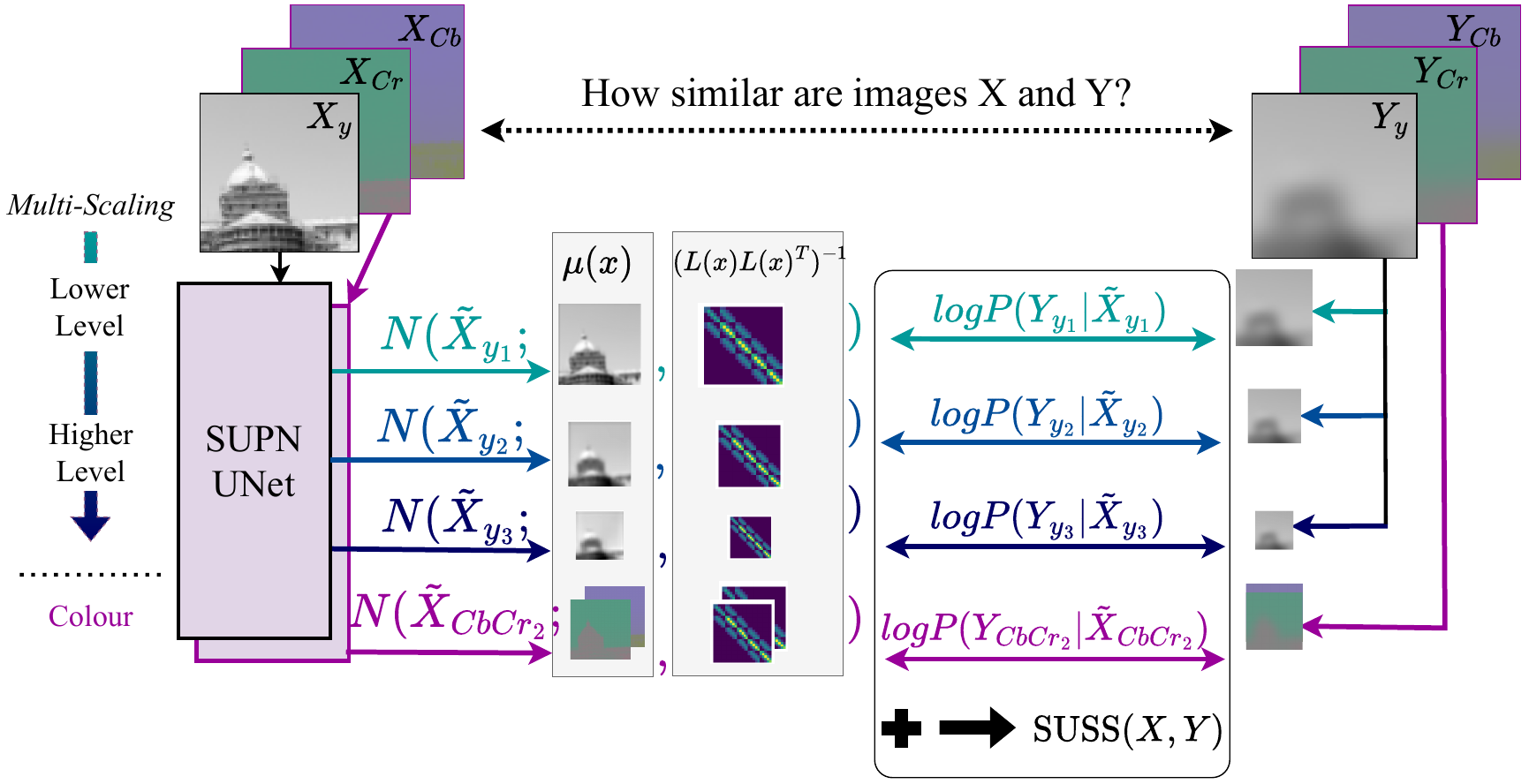}
    \caption{SUSS evaluates the similarity between images by separately constructing distributions for multi-scale structural (y) and color similarity (cb, cr). Our CNN, SUPN UNet, predicts a distribution of perceptually close augmented image data, $\tilde{X}$, as a multivariate normal distribution, with a structured Cholesky factored inverse covariance. This model is trained in a self-supervised fashion.
    SUSS is then calculated as a weighted sum of log probabilities, with weights derived from a training set using human annotations.}
    \label{fig:overview}
\end{figure*}

%% file: sec/related_work.tex
\section{Related Work}

\textbf{Handcrafted approaches}: Traditional IQA metrics incorporate psychophysical principles to approximate human visual perception. Simple pixel-based measures such as MSE and PSNR assume independent pixel errors, giving equal weight to all regions and often failing to reflect perceptual relevance. SSIM and MS-SSIM \cite{ssim,mssim} improved on this by modeling local luminance, contrast, and structural changes. FSIM \cite{zhang2011fsim} added phase congruency and gradient magnitude to emphasize salient regions, while PSNR-HVS \cite{PSNRHVS} introduced contrast sensitivity and frequency masking. GMSD \cite{gmsd} quantifies distortion through gradient magnitude similarity, VIF \cite{vif} models image information using Gaussian scale mixtures, MAD \cite{mad} combines detection and appearance models to simulate human visual response in two stages, with separate strategies for near-threshold and supra-threshold distortions, and NLPD \cite{nlpd} applies divisive normalization in a Laplacian pyramid to mimic early vision. While such measures align with insights from sensory neuron models \cite{watson1985model} and psycho-physics \cite{teo1994perceptual}, they have shortcomings: either they struggle to model human perception accurately, failing to predict preferences in certain IQA tasks \cite{deepf}; or their formulation, such as in MAD with multi-stage processing or in VIF with complex statistical modeling, leads to non-differentiability.
This limits their application as a perceptual objective in downstream tasks.

\noindent\textbf{Supervised Approaches}: 
Learned perceptual metrics measure similarity in deep-feature spaces extracted from networks pre-trained for classification. The Learned Perceptual Image Patch Similarity (LPIPS) \cite{deepf} metric computes feature-space distances from VGG or AlexNet activations, while DISTS \cite{Ding2020ImageQA} combines feature correlations and variances to represent both structure and texture similarity. ST-LPIPS extends LPIPS with spatial-temporal consistency, while PieAPP \cite{pieapp} directly learns a perceptual preference function by predicting the probability that one image appears closer to a reference than another.

These deep-learning approaches achieve higher correlation with human perception than pixel-based metrics, but they have key limitations. They require large, annotated datasets (e.g., BAPPS) and are computationally expensive to train. Their feature embeddings are non-linear, high-dimensional, and optimized for classification rather than perception, leading to a lack of interpretability. They are also sensitive to imperceptible perturbations \cite{ghazanfari2023r} and provide little insight into which features drive their judgments.

\noindent\textbf{Unsupervised Approaches}: Metrics based on information theoretic objectives aim to address some of these limitations. The Perceptual Information Metric (PIM) \cite{bhardwaj2020unsupervised} employs contrastive learning as an objective to model perceptual information in a probability space, while metrics like Deep Wasserstein Distance (DeepWSD) \cite{liao2022deepwsd} compare feature distributions rather than individual points.  The probabilistic perspective is validated by the link between probability-based sensitivity and the Contrast Sensitivity Function  \cite{hepburn2023disentangling}.  However, these methods still rely on large-scale training datasets.

%% file: sec/methods.tex
\section{Methodology}
To provide an interpretable, and robust perceptual similarity score, our method learns the probabilistic structure across several perceptual components of an image. To capture multi-scale, uncertainty-aware perceptual similarity, we frame the problem as probabilistic density estimation over four perceptual components \( C \) that can be explicitly inspected, and learned in a self-supervised manner.  

The final similarity score is computed as a weighted sum of the log-likelihoods of these components, with weights optimized to align with human perceptual judgments on a supervised dataset. Given an image $X$, we learn to predict the distributions of augmented version of each image component \( \tilde{X}_c \) and evaluate the likelihood of \( Y_c \) with respect to each of these, calculating the final score as a weighted sum:

\begin{equation}
    \mathrm{SUSS}(X, Y) = \sum_{c \in C} w_c \log p(Y_c | \tilde{X}_c)
\end{equation}
where \( \tilde{X}_c \) is modelled as a multivariate normal:
\begin{equation}
    \mathcal{N}(\tilde{X}_c: \mu_c(X), (L_c(X)L_c(X)^\top)^{-1})
\end{equation}
with the covariance is represented via a sparse Cholesky of the precision, $L_c(X)$, described in \cref{sec:supn}. 

The distribution of \( \tilde{X} \) is trained in a self-supervised manner to represent a perceptual invariance space formed through small image augmentations, capturing variability in a way that reflects human visual sensitivity. This formulation is grounded in psychophysical principles such as Weber’s Law and the Efficient Coding Hypothesis \cite{barlow1961possible, fechner1860elemente, weber1834pulsu}, which suggest that the human visual system emphasizes structurally relevant information while remaining invariant to small, imperceptible changes (see Section~\ref{sec:aug}).
Our approach follows these principles by assigning probability mass  to perceptually irrelevant structured pixel variations. 
Since the Normal log-likelihood data term \\
$-\frac{1}{2} (Y - \mu(X))^T\Sigma^{-1}(Y - \mu(X))$
can be expressed as:
\begin{equation}
-\frac{1}{2} R^\top L(X)L(X)^\top R = -\frac{1}{2} \sum (L(X)^\top R)^2,
\end{equation}
with $R = Y - \mu(X)$ (and dropping $c$ subscripts for clarity), our method effectively learns to perform a ``whitening", decorrelating and rescaling, operation on the residuals, suppressing perceptually irrelevant differences.

By learning probability distributions over perceptually close variations of each component, SUSS yields higher likelihoods for perceptually similar pairs than dissimilar ones. Conceptually, SUSS acts as a weighted sum of Mahalanobis-like distances, with a learned, structured, data-dependent precision matrix. This allows each component to retain Lipschitz continuity and local convexity properties \cite{boyd2004convex}, which enables smooth and stable optimization when SUSS is used as a loss function \cite{bertsekas1997nonlinear}.

\subsection{Structured Uncertainty Prediction Networks \label{sec:supn}} 

The core of our learned metric is the inference of multivariate probability distributions over image components, capturing both structural features and uncertainty in a computationally efficient way. Following \cite{ivor1}, we implement Structured Uncertainty Prediction Networks (SUPN) to produce these distributions.
To make learning the multivariate distribution tractable, we employ a sparse approximation of the Cholesky decomposition of the precision matrix, predicting $\Sigma$ as a lower-triangular matrix $L$. Sparsity is enforced by restricting predictions to local neighborhoods around each pixel, allowing efficient computation of log probabilities as convolutions. The model estimates both the mean $\mu(X)$ and the log-diagonal and sparse off-diagonal entries of $L(X)$.

We implement this as a UNet-based architecture (Figure \ref{fig:overview}) with a single encoder and two decoder heads, one for $\mu(X)$ (mean) and one for $L(X)$, connected via skip links between corresponding scales. For the luminance (Y) channel, outputs are predicted at three resolutions to capture fine-to-global structures, using 8×8 neighborhoods for full- and half-size images and 5×5 for quarter-size. Chrominance channels (Cb, Cr) are modeled at a single quarter-size resolution with 5×5 neighborhoods, explicitly encoding cross-channel correlations to capture perceptual dependencies.





\subsection{Defining Perceptually Invariant Augmentations\label{sec:aug}}

Central to characterizing the perceptual aspects within the SUPN distributions is learning the distributions of augmented versions of each image component, $\tilde{X}_c$, that capture the variability humans tolerate while preserving structural content. Following psychophysical findings that human perception is more sensitive to structural details than to fine color variations \cite{watson1985model, teo1994perceptual} we model these aspects  separately.
We adopt the YCbCr color space, separating luminance (Y) from chrominance (Cb, Cr). Motivated by prior works, such as MS-SSIM, we decompose the luminance channel into three scales to capture multi-scale structural information. Chrominance channels are modeled at lower spatial resolution, reflecting the reduced color sensitivity. 

To learn distributions aligned with human perception, we construct an augmented dataset capturing natural perceptual variability. Since perception is largely invariant to small transformations that preserve structure \cite{alabau2024invariance}, we apply affine (translation, rotation, scaling) and color (brightness, contrast, saturation, hue) transformations at five intensity levels, ranging from minimal variation to Just Noticeable Difference (JND) thresholds, summarized in Table~\ref{tab:affine}.

For color augmentations, we operate in the perceptually uniform CIELAB space \cite{hill1997comparative}.
Brightness, contrast, saturation, and hue transformations are applied at five perceptually calibrated levels corresponding approximately to perceptual thresholds $\Delta E$, shown in Table~\ref{tab:color}. This allows the model to learn meaningful chromatic variability without overemphasizing imperceptible differences.



\subsection{Choice of Objective Functions\label{sec:opt}}
Training of the SUPN-based architecture is self-supervised, designed to capture the distribution of structurally similar images under perceptual invariances. UNet parameters are optimized to minimize the negative log-probability of augmented images \(T(X)\) at scale \(s\) and augmentation level \(l\), ensuring alignment with the inferred SUPN distribution \(\tilde{X}\):

\begin{equation}
\mathcal{L}_{\text{SUPN}} = \sum_{s=1}^{S} \sum_{l=0}^{L} w_l \cdot \left( -\log p\left( T(X)_{s, l} \mid \tilde{X}_{s, l} \right) \right),
\end{equation}
where \(S\) denotes the number of scales, \(L\) the highest augmentation level, where \(w_l = \frac{1}{l + 1}\)
assigns higher weights to smaller transformations to reinforce perceptual invariance.

Training is done through two one-epoch schedules over the ImageNet training set with corresponding transformed variants (25 augmentations per image; as Sec. ~\ref{sec:aug}), with the unaugmented image serving as input to the network. The first schedule uses a learning rate of \(1\times10^{-4}\), followed by \(1\times10^{-5}\) in the second. Optimization employs AdamW with a weight decay of \(0.01\), \(\beta_1 = 0.9\), and \(\beta_2 = 0.99\), optimized for stable convergence and consistent learning across scales.

The SUPN head predicts the Cholesky factor \(L(X)\), whose structure is asymmetric and small perturbations in its lower-triangular entries can disproportionately affect reconstructed covariances $\Sigma$ of the learned distributions. Given this sensitivity to gradient scaling, separate optimizer settings are used for the final layers of the SUPN head. Thus, the final layers predicting \(L(X)\) uses a smaller weight decay (\(0.001\)) and a slightly higher learning rate (\(1.5\times10^{-4}\)), with \(\beta_1 = 0.78\) and \(\beta_2 = 0.9\) to stabilize learning of SUPN mapping. All other network parameters use the base configuration, where specific parameters were selected through hyperparameter tuning.


\subsection{Learning the Score Component Weightings}
To combine the perceptual components into a single similarity score that aligns with human judgments, we learn the component weights, \(w_c\), from training data. Following \cite{deepf}, weight learning is formulated as a binary classification task using the 2AFC setting, where each triplet consists of a reference image \(X\) and two candidates \(Y_1, Y_2\). 

Weight optimization uses the cross-entropy, comparing the model output differences \(\sigma(\mathrm{SUSS}(X, Y_1) - \mathrm{SUSS}(X, Y_2))\) with human judgments indicating whether \(Y_1\) was perceived as closer to \(X\) than \(Y_2\).

Given differences in component scales, and to ensure positivity, we optimize the log-weights. Optimization proceeds in two stages. First, a coarse grid search estimates the relative weight magnitudes. The best configuration initializes fine-tuning, where weights are optimized by Adam.

\subsection{Dataset Adapted SUSS Variants}
\label{sec:variants}
We propose extensions of SUSS that incorporate the human perceptual judgments alongside self-supervision. In addition to our own augmented dataset, we fine-tune several SUSS variants on benchmark perceptual training datasets. These variants fit bespoke score component weightings, but also adapt to the specific data and augmentation distributions, $\tilde{X}$. We also introduce a regularizing ranking loss in addition to the negative log-probability loss with a weighting of 0.1. 
The first variants, denoted by the dataset name and \textit{-R}, use the reference training images of each dataset but apply our own augmentations to generate $\tilde{X}$. They include a simple pairwise ranking prior, similar to ranking-based IQA formulations, e.g.~\cite{rankiqa}, penalising violations of the desired ordering so that higher distortion levels receive lower similarity likelihoods, enforcing monotonic ranking across transformation levels. 
The second variants, denoted \textit{-RH}, use the dataset-specific transformed images as $\tilde{X}$ together with their human judgment labels, such as pairwise preference probabilities, to derive perceptual levels for the ranking loss. This encourages the model to maximise correlation with human preferences, using a standard correlation-based ranking regulariser, common in IQA models trained directly on MOS scores~\cite{ranking2}. Full details of learned weights for all SUSS variants are provided in Appendix~\ref{app:weights}.

%% file: sec/experiments.tex
\section{Experiments\label{sec:exp}}
\textbf{Alignment with Human Perception Scores}:
We assess SUSS'  performance as a similarity measure aligned with human perceptual judgment by evaluating relevant properties for perceptual training objectives including adaptability to unseen distortions, smoothness across distortion levels, and consistency with subjective visual quality judgment. In all experiments, we compare to both traditional handcrafted, as well as deep-learning based approaches.

We benchmark on three major datasets for perceptual evaluation in human classification tasks: the BAPPS 2-AFC dataset \cite{deepf}, the PieAPP 2-AFC dataset \cite{pieapp}, and the PIPAL dataset \cite{pipal}. BAPPS consists of 36.3K triplets, each containing a reference image and two distorted images with human vote proportions indicating which is more similar to the reference. Triplets are grouped into six categories: traditional photometric and geometric transformations(Trad), CNN-based distortions(CNN), super-resolution(SuperRes), deblurring(Deblur), colorization(Color), and frame interpolation(FrameInterp). The small patches and subtle distortions in this dataset primarily assess fine-grained similarity.
The PieAPP 2-AFC dataset \cite{pieapp} contains 200 reference images paired with more than 20k distorted image pairs with human preference labels, covering a range of low- and high-level distortions. The images are mostly natural scenes with more complex, higher-level distortions compared to BAPPS.
The PIPAL dataset \cite{pipal} contains 29K distorted images from 250 reference images, with human judgments collected via an Elo rating system. Distortions include traditional noise, blur, compression artifacts, GAN-based restoration outputs, with more noticeable, diverse distortions, providing a wider range of non-trivial distortions compared to the other datasets.
We compare our ImageNet trained model, SUSSBase, but also fine-tuned variants to assess the influence of dataset-specific human judgment data on our models performance on these tasks(see Section~\ref{sec:variants}). We assess the performance of individual model components independently, as well as the final combined model.

We further evaluate core properties required for perceptual loss applicability, focusing on perceptual calibration and robustness of the score function on an addition dataset, KADID-10k \cite{kadid10k}. This includes 81 pristine images degraded by 25 distortion types across five intensity levels with corresponding human mean opinion scores (MOS). Scores are analyzed across distortion types and MOS levels to assess whether they vary smoothly with increasing distortion, are separable across perceptual quality ranges, and remain consistent across transformation categories-indicating calibration to perceptual  similarity. 
We additionally evaluate SUSS on a dataset of transformed ImageNet validation images, across five calibrated distortion levels where Level 3 corresponds to the just-noticeable difference (JND) threshold, similar to our training setup. These assess score smoothness and perceptual linearity across both geometric and appearance-based transformations, while examining the model’s learned spatial and textural invariances, its robustness to unseen distortions, and its ability to reflect perceptual similarity aligned with human judgment.

\noindent\textbf{Interpretability: How well does the model understand and make interpretable decisions?}
SUSS produces a mean and a structured covariance matrix for each image component. This corresponds to a learned, image specific linear transformation applied to the residual differences between two images, giving a distance metric that is both flexible and fully explainable by construction. Each component is parameterised by a SUPN distribution learned to align with human perception. We present three ways to inspect the model and understand which parts of SUSS contribute to specific decisions on a pixel space level:
First, by sampling from the inferred SUPN distributions, we observe the range of perceptually plausible images and separately inspect the learned distributions for color and multi scale structural components, and how each contributes to the final similarity score. 
Second, for each component we examine the whitened residuals $L(X)^{\top}R$ between compared images. These residuals reveal the learned covariance structure and the perceptual invariances captured by the model, showing which features are weighted more strongly in the component wise similarity judgment.
Third, to understand the final combined SUSS score at a pixel level, we construct SUSS maps by weighting the whitened residuals with the component wise scores and taking the square root. This produces a spatial relevance map that reflects how much each pixel contributes to the actual SUSS value.



\noindent\textbf{Loss Applicability}:
In order to investigate the performance of SUSS as a loss objective in guiding reconstruction in downstream tasks, we conducted two qualitative experiments. In the first, we compare reconstructing the pixel values of a target image starting from a random highly structured initialization. We compare against LPIPS, and for fairness, perform a sweep over learning rates for both the Adam and LBFGS optimizers, and select the runs that best minimize the perceptual distance. 

We also tested SUSS as a loss for the inverse problem of single-image super-resolution. This task is underdetermined and ill-posed, meaning there exist multiple valid high-resolution reconstructions for a given low-resolution input. As such, the choice of loss function is critical in guiding the optimization toward learning perceptually close solutions. We chose the state-of-the-art super-resolution model DRCT\cite{hsu2024drct} and evaluated how well two SUSS variants, SUSSBase and SUSSPIPAL-RH, perform in comparison to standard losses, including: L1, L2, SSIM, and LPIPS.
All models were pretrained on ImageNet pairs and fine-tuned for the same number of epochs on the DIV2K dataset\cite{div2k} using identical training schedules.

%% file: sec/results.tex
\section{Results}
\begin{figure}
    \centering
    \begin{minipage}{0.5\textwidth}
        \centering
        
        \begin{subfigure}[b]{\textwidth}
            \centering
            \includegraphics[width=\textwidth]{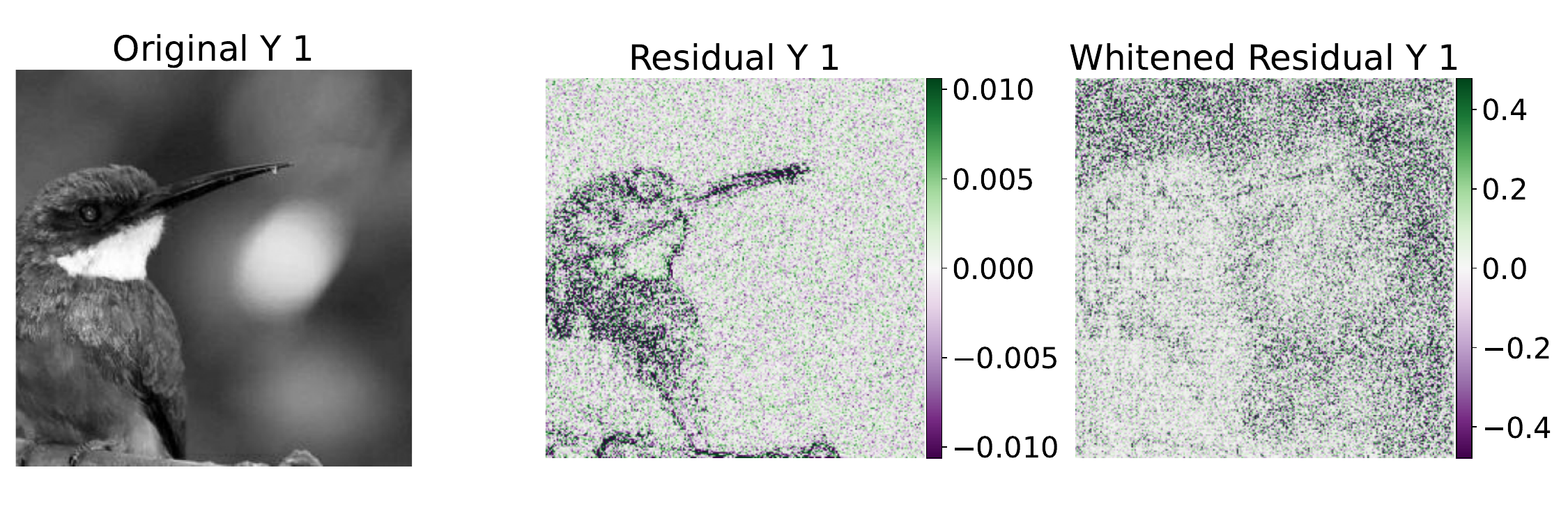}
            \caption{Residual plots for the Y channel following a perspective warp (level 3)}
        \end{subfigure}
        \vspace{0.3cm}
        \begin{subfigure}[b]{1\textwidth}
            \label{fig:pieapp_wr}
            \centering
            \includegraphics[width=\textwidth]{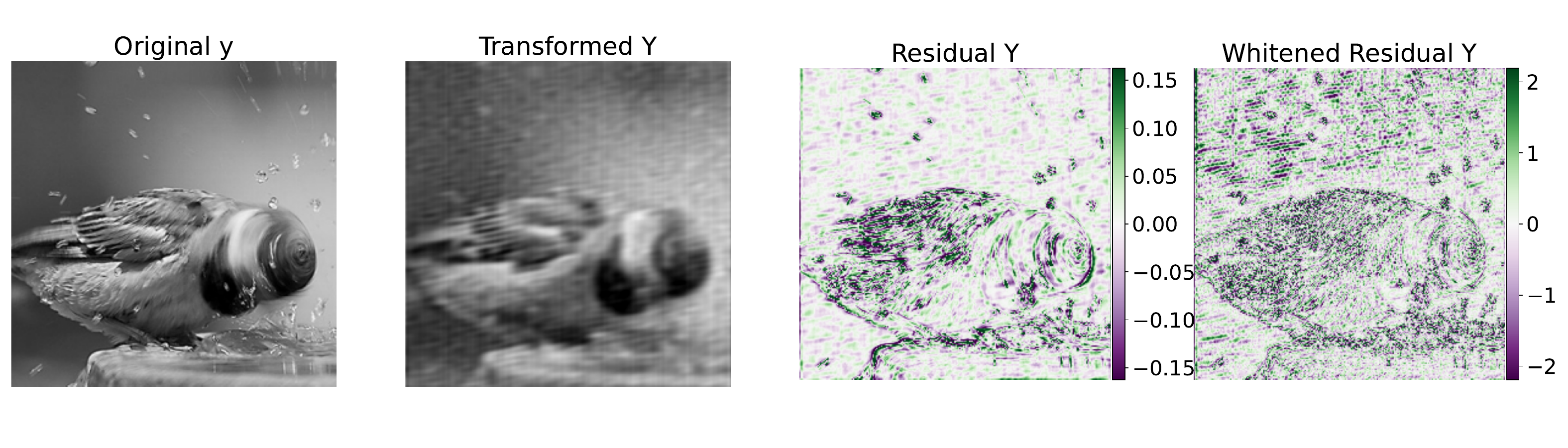}
            \caption{Residual plots for a dissimilar example from PieApp, with removed intra-frequency fourier coefficient. This has a low human preference scores(0.071).}
            \end{subfigure}
        \vspace{0.3cm}
        \begin{subfigure}[b]{\textwidth}
            \centering
            \includegraphics[width=\textwidth]{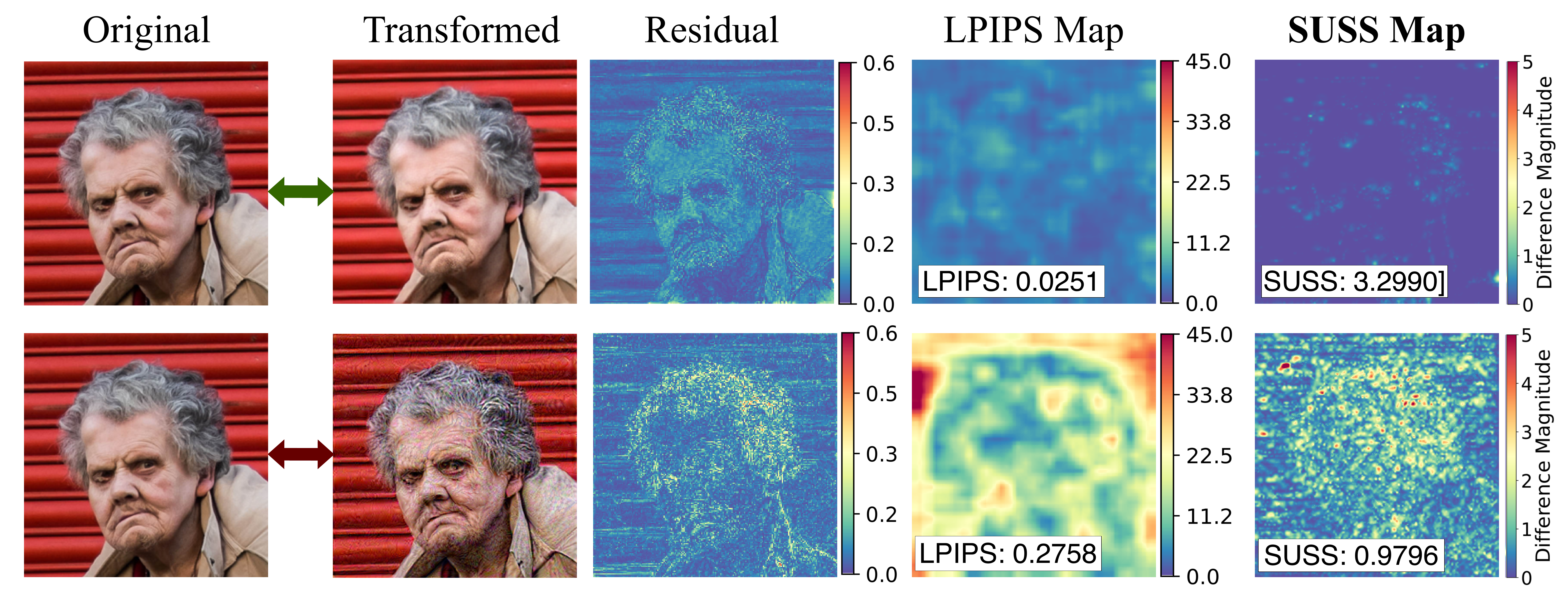}
            \caption{Residual plots and interpretability maps for 
            PIPAL 2AFC pairs. The upper row  images are judged as visually similar, the lower row less so.  
            The "Residual" is the absolute pixel-wise difference image.  
            The "LPIPS map" shows the per-pixel difference magnitude in deep feature maps of a pretrained LPIPS network.
            The "SUSS map" displays the weighted squared sum of whitened residuals, highlighting regions learned to be important by the SUSS metric.}
        \end{subfigure}
    \end{minipage}\vspace{-10pt}
    \caption{
    Inspection of perceptually relevant pixel structures learned by SUSS.
    (a–b) show small and medium transformations. The residual is the difference of feature maps, and the whitened residual $L(X)^\top R$ is a linear transform of this, as learned by SUSS. The whitened residual highlights regions of structure that is considered perceptually distant, suppressing perceptually irrelevant noise.  
    (c) illustrates the summative SUSS map.
    }
    \label{fig:whitened_residuals}
\end{figure}

\textbf{Interpretability/Explainability}:
Figures~\ref{fig:whitened_residuals} and \ref{fig:sample_colour_residu} show qualitative results illustrating how the learned covariance structures shape the residuals used in the SUSS score. In each example, the raw residual between two transformed images is shown alongside the corresponding whitened residual $L(X)^{\top}R$. Across Y-channel and color components, the whitened residuals clarify the learned score: structures that are perceptually important become more pronounced, while perceptually invariant differences, such as small shifts, are reduced. For instance, in Fig.\ref{fig:whitened_residuals} b and c, unexpected background distortions appear more clearly after whitening, whereas in a) structured regions visible in the raw residual are largely suppressed. More examples across distortion types and MOS ranges are shown in Appendix~\ref{app:inter}.

\begin{figure}
    \centering
    \includegraphics[width=1\linewidth]{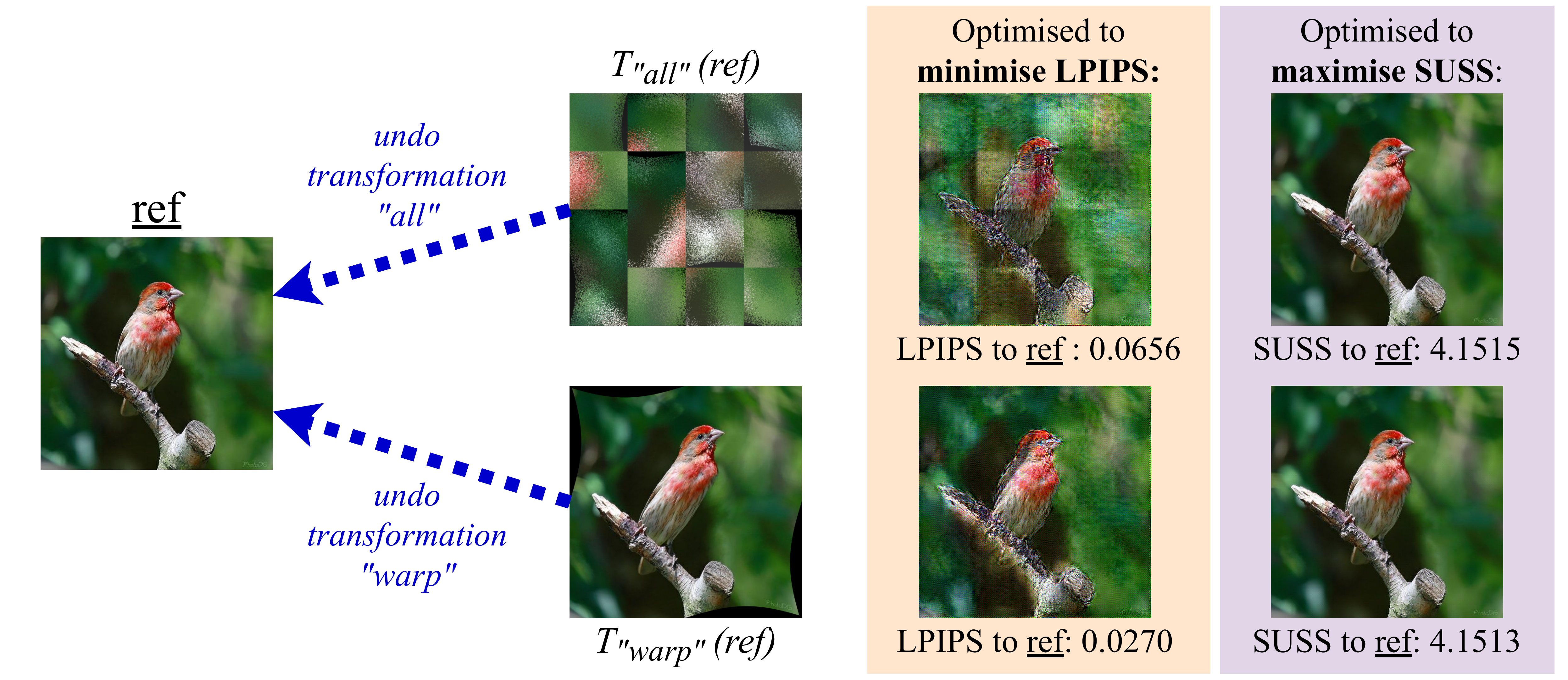}
    \caption{How well does SUSS navigate the optimization landscape as a perceptual loss objective? We perform a reconstruction task where the goal is to recover the original image (left) from distorted inputs. Using LPIPS and SUSS as loss objectives, with optimal optimization settings, we minimize the perceptual score between the reference and the input and visualise the results. Both losses converge to high similarity scores (matching MOS levels associated with imperceptible differences in KADID10K). 
    }
    
    \label{fig:reconstruction}
\end{figure}
\begin{figure}
    \centering
    \includegraphics[width=1\linewidth]{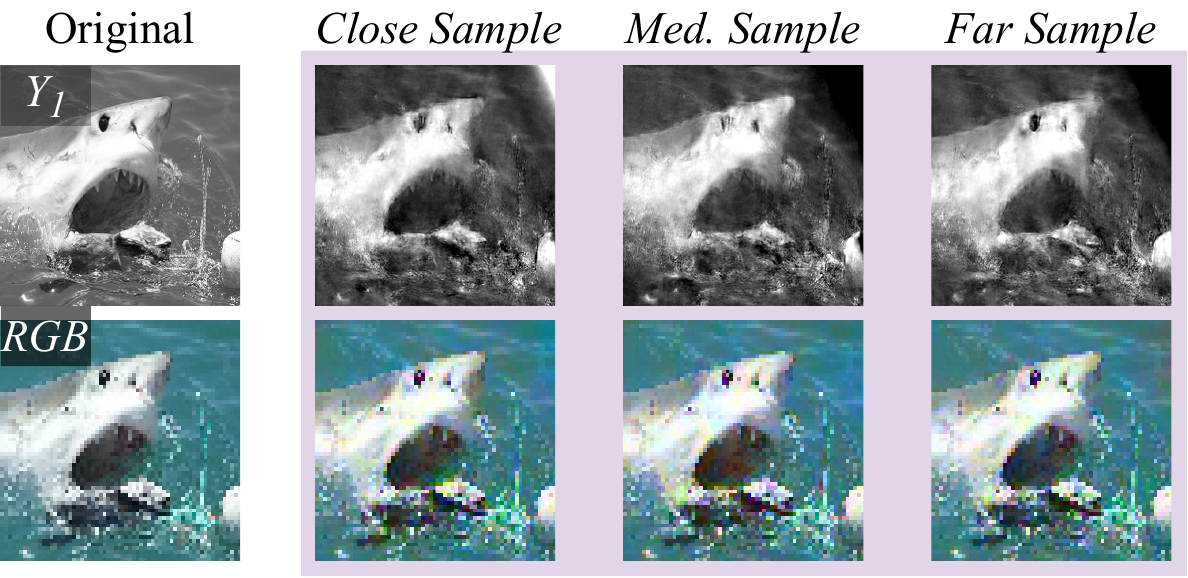}
    \caption{Samples from learned distribution (SUSSPieApp-RH model) with original image, and close, medium and far samples(left to right) for different components of the SUSS model, RGB shows CbCr samples combined with the original Y.
    Samples are drawn from the learned SUPN distributions; we select the minimum, median, and maximum probabilities from 1000 samples.
    }
    \label{fig:samples}
\end{figure}
Further insights come from sampling the learned distributions as shown in Fig. ~\ref{fig:samples} and Appendix~\ref{app:inter}.  Close samples visibly preserve characteristic structural details, such as eye contours, while medium and far samples show stronger deviations whilst still remaining perceptually similar.

\begin{table}[h!]
\centering
\begin{subtable}[t]{\columnwidth}
\centering
\resizebox{\columnwidth}{!}{%
\begin{tabular}{|l|c|c|c|c|c|c|c|}
\hline
\textbf{Method} & \textbf{Trad} & \textbf{CNN} & \textbf{Super} & \textbf{Deblur} & \textbf{Color} & \textbf{Frame} & \textbf{All} \\
\hline
Human & 80.8 & 84.4 & 73.4 & 67.1 & 68.8 & 68.6 & 73.9 \\
\hline
PSNR & 57.3 & 80.1 & 64.2 & 59.0 & 62.5 & 54.4 & 62.9 \\
SSIM & 57.5 & 76.9 & 61.6 & 58.4 & 52.2 & 55.4 & 60.3 \\
MS-SSIM & 58.8 & 76.8 & 63.9 & 58.9 & 52.3 & 57.3 & 61.3 \\
FSIM  & 62.8 & 79.4 & 66.1 & 59.0 & 57.1 & 58.1 & 63.8 \\
GMSD & 60.9 & 77.2 & 67.6 & 59.5 & 51.8 & 57.5 & 62.4 \\
$\ast$VIF & 56.3 & 75.9 & 63.4 & 58.8 & 52.6 & 54.0 & 60.1 \\
$\ast$MAD & 63.7 & 80.2 & 67.5 & 58.8 & 55.4 & 58.9 & 64.4 \\
NLPD & 55.0 & 76.6 & 63.4 & 58.5 & 52.6 & 54.0 & 60.4 \\
\hline
LPIPS (Alex) & 74.0 & 82.4 & 70.5 & 59.9 & 61.9 & 62.8 & 68.6 \\
LPIPS+ (Alex-lin) & 73.9 & 82.9 & 70.7 & 60.1 & 62.2 & 62.2 & 68.7 \\
LPIPS (VGG) & 71.4 & 81.4 & 69.0 & 57.3 & 58.8 & 62.7 & 66.8 \\
LPIPS+ (VGG-lin) & 70.5 & 81.7 & 69.2 & 57.9 & 59.1 & 62.0 & 66.7 \\
ST-LPIPS & 71.9 & 81.2 & 69.6 & 60.9 & 62.9 & 63.1 & 68.3 \\
DISTS & 74.9 & 81.3 & 69.2 & 59.9 & 61.5 & 62.8 & 68.3 \\
PieAPP & 59.4 & 58.2 & 59.8 & 68.5 & 62.6 & 62.5 & 63.2 \\
\hline
SUSSBase & 60.3 & 70.5 & 58.7 & 60.1 & 56.7 & 56.7 & 60.5 \\
SUSSBAPPS-R & 64.6 & 70.1 & 59.0 & 60.3 & 59.0 & 59.5 & 62.1 \\
SUSSBAPPS-RH & 64.5 & 72.5 & 66.7 & 64.0 & 59.1 & 58.9 & 64.3 \\
\hline
SUSSBaseY1 & 61.1 & 71.1 & 59.2 & 60.3 & 55.2 & 56.9 & 60.5 \\
SUSSBaseY2 & 60.2 & 70.2 & 58.6 & 59.9 & 56.6 & 56.6 & 60.3 \\
SUSSBaseY3 & 61.4 & 71.4 & 59.3 & 60.2 & 55.5 & 56.8 & 60.5 \\
SUSSBaseCbCr & 60.1 & 68.7 & 55.9 & 55.0 & 59.5 & 53.6 & 58.8 \\
\hline
\end{tabular}%
}
\caption{Performance on BAPPS 2AFC afc tasks in accuracy {$\uparrow$} .}
\label{tab:bapps}
\end{subtable}

\begin{subtable}[t]{0.48\columnwidth}
\centering
\resizebox{\columnwidth}{!}{%
\begin{tabular}{|l|c|c|c|}
\hline
\textbf{Method} & \textbf{KRCC}{$\uparrow$} & \textbf{PLCC}{$\uparrow$} & \textbf{SRCC}{$\uparrow$}\\
\hline

PSNR & 0.2763 & 0.4034 & 0.4066 \\
SSIM & 0.2180 & 0.2491 & 0.3022 \\
MS-SSIM & 0.2237 & 0.0350 & 0.2930 \\
FSIM & 0.2580 & 0.3144 & 0.3485 \\
GMSD & 0.1928 & 0.2173 & 0.2589 \\
$\ast$VIF & 0.1540 & 0.2459 & 0.2155 \\
$\ast$MAD & 0.2609 & 0.3682 & 0.3572 \\
NLPD & 0.1874 & 0.0832 & 0.2454 \\
\hline
LPIPS (Alex) & 0.3898 & 0.6011 & 0.5499 \\
LPIPS+ (Alex-lin) & 0.3848 & 0.5988 & 0.5423 \\
LPIPS (VGG) & 0.3461 & 0.5029 & 0.4919 \\
LPIPS+ (VGG-lin) & 0.3253 & 0.4724 & 0.4618 \\
ST-LPIPS & 0.4734 & 0.6862 & 0.6498 \\
DISTS & 0.5070 & 0.7165 & 0.6932 \\
PieAPP & 0.6279 & 0.7959 & 0.8141 \\
\hline
SUSSBase & 0.2569 & 0.3858 & 0.3615 \\
SUSSPieApp-R & 0.2728 & 0.4545 & 0.3921\\
SUSSPieApp-RH & 0.3094 & 0.4721 & 0.4031\\
\hline
\end{tabular}%
}
\caption{Results on PieAPP 2AFC.}
\label{tab:pieapp}
\end{subtable}
\hfill
\begin{subtable}[t]{0.48\columnwidth}
\centering
\resizebox{\columnwidth}{!}{%
\begin{tabular}{|l|c|c|c|}
\hline
\textbf{Method} & \textbf{KRCC}{$\uparrow$} & \textbf{PLCC}{$\uparrow$} & \textbf{SRCC}{$\uparrow$}\\
\hline
PSNR & 0.2763 & 0.4034 & 0.4066 \\
SSIM & 0.3489 & 0.4993 & 0.5041 \\
MS-SSIM & 0.3973 & 0.5622 & 0.5619 \\
FSIM & 0.4159 & 0.6088 & 0.5896 \\
GMSD & 0.4138 & 0.6216 & 0.5835 \\
$\ast$VIF & 0.3995 & 0.5598 & 0.5632 \\
$\ast$MAD & 0.4248 & 0.6073 & 0.5988 \\
NLPD & 0.3349 & 0.5034 & 0.4825 \\
\hline
LPIPS (Alex) & 0.4097 & 0.5826 & 0.5853 \\
LPIPS+ (Alex-lin) & 0.4159 & 0.5904 & 0.5931 \\
LPIPS (VGG) & 0.4042 & 0.5973 & 0.5731 \\
LPIPS+ (VGG-lin) & 0.4209 & 0.6186 & 0.5935 \\
ST-LPIPS & 0.4481 & 0.6283 & 0.6307 \\
DISTS & 0.4069 & 0.5796 & 0.5790 \\
PieAPP & 0.5148 & 0.6952 & 0.7040 \\
\hline
SUSSBase & 0.3268 & 0.4296 & 0.4682\\
SUSSPIPAL-R & 0.3443 & 0.4985 & 0.5270\\
SUSSPIPAL-RH & 0.3919  & 0.5486 & 0.5518\\
\hline
\end{tabular}%
}
\caption{Results on PIPAL.}
\label{tab:pipal}
\end{subtable}
\caption{Benchmark results on the main perceptual loss benchmark datasets. 
SUSS models consistently outperform heuristic metrics (top part of tables) and perform competitively with deep learning–based perceptual measures (below that). Models with a "$\ast$" are non-diffentiable, so not applicable as loss functions.}
\label{tab:quantitative_results}

\end{table}
\noindent\textbf{Quantitative Evaluation}:
Table~\ref{tab:bapps} presents the performance of different models on the BAPPS 2AFC dataset. Across all categories, SUSS models perform competitively with state-of-the-art deep learning-based perceptual metrics while consistently outperforming heuristic metrics such as PSNR, SSIM, and MS-SSIM. Notably, the fine-tuned SUSSBAPPS-RH model achieves the highest overall accuracy among the SUSS variants, surpassing heuristic metrics and performing on par with or slightly better than PieAPP in certain categories. 
Tables~\ref{tab:pieapp} and \ref{tab:pipal} further support this. On PieAPP, SUSS variants achieve higher correlation scores than traditional metrics and remain competitive with deep learning–based approaches. On PIPAL, the fine-tuned models outperform pixel-based metrics and show performance comparable to LPIPS, ST-LPIPS, and DISTS, indicating robustness across diverse distortion types.
 
Component analysis (Table~\ref{tab:bapps}) shows that SUSSBaseY performs best on Y-channel and deblurring distortions, while SUSSBaseCbCr performs best on color distortions. Combining these components into the full SUSS model yields the most consistent results across all datasets, particularly for color and more complex transformations.

\begin{table}[t]
\centering
\resizebox{\columnwidth}{!}{
\begin{tabular}{lcccccccc}
\toprule
\textbf{Metric} & \textbf{Blurs} & \textbf{Color} & \textbf{Compression} & \textbf{Noise} & \textbf{Brightness} & \textbf{Spatial} & \textbf{Sharp./Contr.} & \textbf{Other} \\
\midrule
SUSSPieApp & 0.1147 & 0.0481 & 0.0412 & 0.1006 & 0.0336 & 0.0222 & 0.5080 & 0.0304 \\
PSNR       & 0.3262 & 0.5220 & 0.4170 & 0.7462 & 0.3885 & 0.3180 & 0.5988 & 0.3604 \\
SSIM       & 0.1855 & 0.0526 & 0.1476 & 0.0837 & 0.2033 & 0.1386 & 0.3329 & 0.1894 \\
MS-SSIM    & 0.1440 & 0.0887 & 0.1569 & 0.1998 & 0.1320 & 0.1275 & 0.4634 & 0.2111 \\
LPIPS      & 0.1298 & 0.2210 & 0.1181 & 0.0744 & 0.2292 & 0.1771 & 0.2309 & 0.2359 \\
NLPD       & 0.2466 & 0.1151 & 0.2974 & 0.4162 & 0.2258 & 0.1374 & 0.3589 & 0.3750 \\
DISTS      & 0.1606 & 0.2517 & 0.1230 & 0.1819 & 0.3626 & 0.1897 & 0.2069 & 0.2608 \\
PieAPP     & 0.2284 & 0.3395 & 0.2395 & 0.2350 & 0.1045 & 0.0656 & 0.6763 & 0.2572 \\
\bottomrule
\end{tabular}
}
\caption{Perceptual Calibration as measured by category-wise KL Divergence between distance distributions, $\mathrm{KL}(D_{\text{category}} || D_{\text{agg}})$, where all distortions have similar perceptual magnitude. Lower values indicate more similar distributions of distances between a distortion category and the aggregated distribution.}
\label{tab:kl_divergences_categories}
\end{table}

\noindent\textbf{Perceptual Calibration}:
Table~\ref{tab:kl_divergences_categories} reports category-wise KL divergence on KADID-10k for distortions with similar perceptual magnitude. 
SUSS obtains consistently low values across  categories, implying that it mimics perceptual distance across transformations. In contrast, PSNR, SSIM, MS-SSIM and several deep metrics show much higher variability, indicating sensitivity to the distortion class.

Figures~\ref{fig:kad1} and \ref{fig:kad2} illustrate MOS-conditioned score distributions on KADID-10k. For high MOS pairs (imperceptible differences), SUSS values are tightly clustered across distortion types, indicating consistent behavior with minimal perceptual differences. For low MOS pairs the distributions remain well separated. In comparison, MSE shows large overlaps across MOS levels, and LPIPS and SSIM exhibit broader spreads for stronger distortions. 
Full MOS-conditioned distributions are included in Appendix~\ref{app:res}. Figures~\ref{fig:kandid} and \ref{fig:baaps_trans} further examine behavior across increasing transformation strength on  KADID-10k and ImageNet. 

\noindent\textbf{Image Reconstruction}: 
Results are illustrated in Fig.~\ref{fig:reconstruction} with comparison against LPIPS. We show that SUSS reliably achieves very high scores. Similarly, LPIPS, achieves a low disimilarity (indicating a high degree of perceptual similarity) but the reconstructions are clouded with artifacts.

\noindent\textbf{Super-Resolution}: 
Results on the test set can be found in Appendix~\ref{App:sr} with visual comparion with other loss functions. When used directly as a perceptual loss objective, SUSS produces results that are visually comparable to heuristic perceptual loss functions such as SSIM, with fewer artifacts than are observed in LPIPS. These findings suggest that SUSS is suitable for being integrated as a differentiable perceptual loss in training pipelines. 

%% file: sec/discussion_new.tex
\section{Discussion}
\textbf{Performance as a Perceptual Score}:
Our results show that SUSS aligns well with human perceptual judgment across the major benchmark datasets. The model captures perceptual similarity and the space of perceptual invariances around an image while remaining sensitive to actual perceptual changes and not to semantic changes, which is important for downstream use. SUSS accords better with human perception than non deep learning based approaches, except for MAD and VIF, which are not suitable as a loss function (Tables ~\ref{tab:quantitative_results}). Performance on PIPAL is lower, likely because the distortions are larger and the rating system differs from the other datasets. To learn these characteristics,  more effectively, the model likely requires further training or broader distortion coverage. 
Experiments on the KADID-10k and transformed ImageNet datasets confirm strong perceptual calibration compared to other methods. 

\noindent\textbf{Adaptability}:
Across all experiments, SUSS consistently adapts well to new data and unseen distortion types, even in settings without human labels. The strong performance of the separate SUSSBase components shows that our SUPN framework captures perceptual structure directly via the learned covariance information in $L$. The full SUSS model further improves this behavior and generalizes consistently to fully unseen distortions and transformation types, as demonstrated in the benchmark evaluations and in the controlled transformation experiments. These findings indicate that SUSS learns transferable perceptual cues rather than dataset-specific artifacts. Additional comparisons of different training regimes and their effects on the resulting SUSS maps are provided in Appendix~\ref{app:inter}.

\noindent\textbf{Interpretability}:
SUSS provides transparent and interpretable perceptual judgments through its learned, image-specific linear transform on residual differences, which limits the scope for unexpected artifacts. The resulting SUSS maps highlight regions where the differences between two images deviate from the perceptual invariances expected for that image and thus reveal which structures drive the similarity score. This behavior contrasts with deep feature map differences, which rely on long-range and highly non-linear transformations that are difficult to interpret. While MSE and SSIM can offer interpretability, they are too rigid and do not capture perceptual structure. Our approach achieves both strong perceptual performance and direct transparency, giving insight into why two images are judged as similar or different. This level of interpretability is important for applications that require explainability and trustworthiness, such as medical imaging and safety-critical vision systems. A failure case where SUSS diverges from human judgment is included in Appendix~\ref{app:inter}.

\noindent\textbf{Applicability}: For downstream perceptual loss use, the formulation of SUSS is central. By defining the score as a weighted sum of negative Mahalanobis-like distances with a learned, image-specific precision matrix, SUSS avoids injected unknown invariances that are encoded in discriminative deep models and instead inherits desirable norm-like properties. Mahalanobis distances provide Lipschitz continuity and a degree of local convexity \cite{boyd2004convex}, and weighted combinations retain these properties sufficiently for smooth optimization behavior \cite{bertsekas1997nonlinear}. Our results support this: Table~\ref{tab:kl_divergences_categories} shows low category-wise KL divergence, indicating consistent perceptual calibration across distortion types of similar perceptual magnitude. The component models in Tables~\ref{tab:bapps} capture complementary structure, and the combined SUSS model provides the most stable performance across datasets. Figures~\ref{fig:kad1} and ~\ref{fig:kad2} show compact and well-separated score distributions for high- and low-MOS pairs, and Figures~\ref{fig:baaps_trans} and ~\ref{fig:kandid} confirm smooth behavior across distortion levels on both KADID-10k and the transformed ImageNet images.
These properties carry over to optimization tasks. In the reconstruction experiment (Fig.~\ref{fig:reconstruction}), SUSS provides stable gradients from different initial transformations and produces reconstructions with far fewer artifacts than LPIPS, which often converges to visually distorted solutions despite reporting high similarity. Initial super-resolution results (Appendix~\ref{App:sr}) further show that SUSS can be used directly as a perceptual loss and produces outputs comparable to SSIM with fewer artifacts than LPIPS. Unlike LPIPS, which is typically paired with L1 or L2 to suppress artifacts, SUSS can be used on its own. Precomputing the Cholesky factors for each target image also makes evaluation efficient. Overall, the results confirm that SUSS is directly applicable as a perceptual loss, with smooth optimization behavior and reliable perceptual alignment.

\noindent\textbf{Limitations and Future Work}:
While our results show that SUSS is a promising perceptual loss, further investigation of its behavior in larger-scale application would be beneficial, including a closer analysis of optimization stability. 
A current limitation is that SUSS is not symmetric, i.e.\ $\mathrm{SUSS}(A,B) \neq \mathrm{SUSS}(B,A)$. Appendix~\ref{App:assy} provides quantitative asymmetry results. Although asymmetry does not prevent its use as a loss function, 
 its impact on different training scenarios should be examined. Future work will also explore alternative ways of capturing perceptual invariances, including expert-driven invariances and more structured perceptual priors, and how well SUSS reflects the relevant aspects of human perception in these settings.

%% file: sec/conclusion.tex
\section{Conclusion}
We introduced SUSS, a new probabilistic image similarity score, which is explainable by construction. Our results demonstrate that SUSS is better aligned with human perception than non-deep feature approaches, is suitable for use as a loss function and enables a high-degree of interpretability. \\
\textbf{Code/models} will be made available on publication.

%% file: sec/appendix.tex
\appendix

\section{Learned SUSS Weights}
\label{app:weights}

The following weights were obtained for different SUSS variants:

\begin{itemize}
  \item \textbf{SUSS (base)}: \([8.3633\times10^{-6},\, 4.1081\times10^{-8},\, 6.3725\times10^{-5},\, 6.0119\times10^{-6}]\)

  \item \textbf{SUSSBAPPS-R}: \([4.0456\times10^{-8},\, 1.6388\times10^{-5},\, 4.6524\times10^{-6},\, 1.4295\times10^{-6}]\)

  \item \textbf{SUSSBAPPS-RH}:  
  \([2.6266\times10^{-7},\, 1.4119\times10^{-6},\, 1.1266\times10^{-5},\, 4.2387\times10^{-9}]\) 

  \item \textbf{SUSSPIPAL-R}: \([3.1341\times10^{-7},\, 4.053\times10^{-6},\, 1.1602\times10^{-4},\, 7.2145\times10^{-8}]\)

  \item \textbf{SUSSPIPAL-RH}: \([4.2111\times10^{-7},\, 4.1700\times10^{-7},\, 1.0702\times10^{-5},\, 8.4151\times10^{-8}]\)

  \item \textbf{SUSSPieApp-R}: \([1.1689\times10^{-4},\, 4.3226\times10^{-8},\, 4.3802\times10^{-8},\, 4.2813\times10^{-8}]\)

  \item \textbf{SUSSPieApp-RH}:  
  \([4.2685\times10^{-6},\, 4.0278\times10^{-5},\, 1.1529\times10^{-4},\, 2.7169\times10^{-8}]\)
\end{itemize}
\section{Training Dataset Details}
\label{app:data}

\begin{table}[h!]
\centering
\resizebox{1\textwidth}{!}{
\begin{tabular}{lccccc}
\hline
Transformation & Level 0 & Level 1 & Level 2 & Level 3 (JND) & Level 4 \\
\hline
Translation & (0.008, 0.01) & (0.01, 0.03) & (0.03, 0.05) & (0.05, 0.07) & (0.07, 0.10) \\
Rotation (°) & (0.01, 0.5) & (0.5, 1.4) & (1.4, 2.3) & (2.3, 3.2) & (3.2, 4.2) \\
Scaling & (1.001, 1.005) & (1.005, 1.01) & (1.01, 1.02) & (1.02, 1.03) & (1.03, 1.04) \\
Elastic $(\alpha,\sigma)$ & (1.0, 0.5) & (5.0, 1.0) & (10.0, 2.0) & (15.0, 3.0) & (20.0, 4.0) \\
Perspective & 0.05 & 0.10 & 0.15 & 0.20 & 0.25 \\
\hline
\end{tabular}
}
\caption{Affine and geometric transformation levels, proportional to image size, based on human perceptual thresholds.}
\label{tab:affine}
\end{table}

\begin{table}[h!]
\small
\centering
\resizebox{1\textwidth}{!}{
\begin{tabular}{lccccc}
\hline
Transformation & Level 0 (\(\Delta E \approx 1\)) & Level 1 (\(\Delta E \approx 2{-}3\)) & Level 2 (\(\Delta E \approx 4\)) & Level 3 (\(\Delta E \approx 5{-}6\)) & Level 4 (\(\Delta E \approx 7\)) \\
\hline
Brightness  & (0.001, 0.002) & (0.002, 0.005) & (0.005, 0.008) & (0.008, 0.015) & (0.015, 0.03) \\
Contrast    & (0.002, 0.004) & (0.004, 0.008) & (0.001, 0.0015) & (0.015, 0.02) & (0.025, 0.035) \\
Saturation  & (0.0002, 0.0005) & (0.0005, 0.001) & (0.001, 0.015) & (0.015, 0.02) & (0.025, 0.035) \\
Hue         & (0.01, 0.02) & (0.02, 0.03) & (0.03, 0.04) & (0.04, 0.05) & (0.05, 0.06) \\
\hline
\end{tabular}
}
\caption{Color transform levels used to generate the perceptually aligned training dataset, based on CIELAB $\Delta E$ thresholds \cite{hill1997comparative}.}
\label{tab:color}
\end{table}
\clearpage\section{Interpretability Results}
\label{app:inter}

\begin{figure*}[t]
    \begin{subfigure}[b]{\textwidth}
        \centering
        \includegraphics[width=0.6\linewidth]{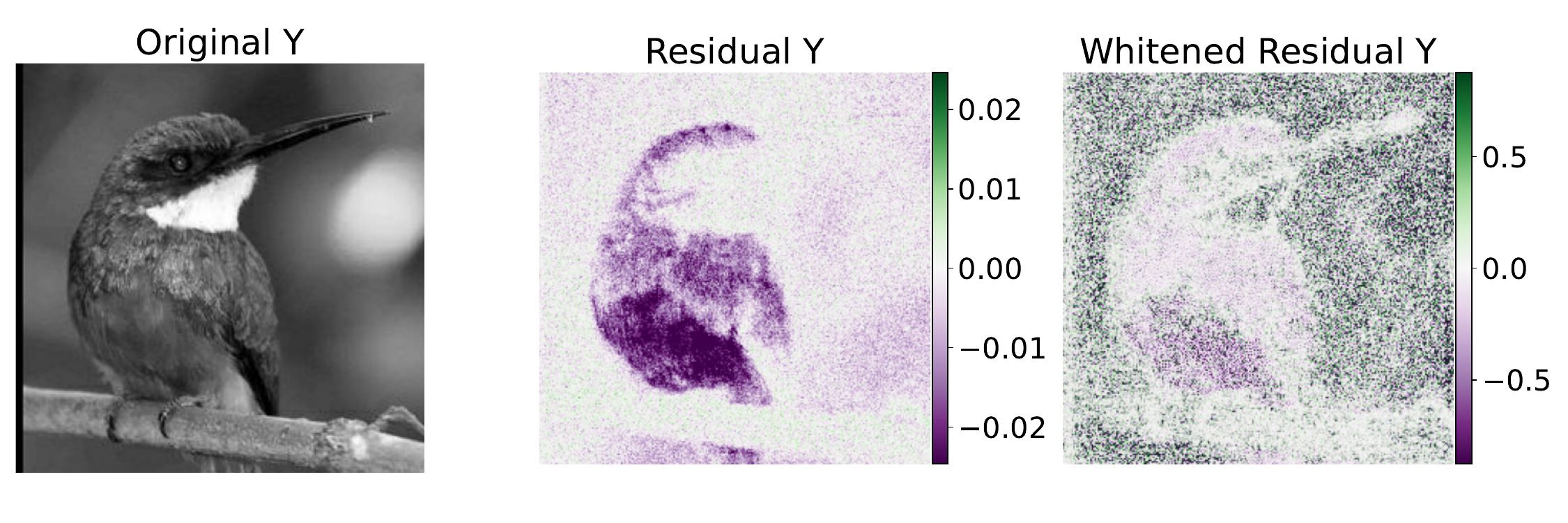}
        \caption{Y Channel Original Scale, Colour Transformation + Shift Level 3}
    \end{subfigure}

    \begin{subfigure}[b]{\textwidth}
        \centering
        \includegraphics[width=0.6\linewidth]{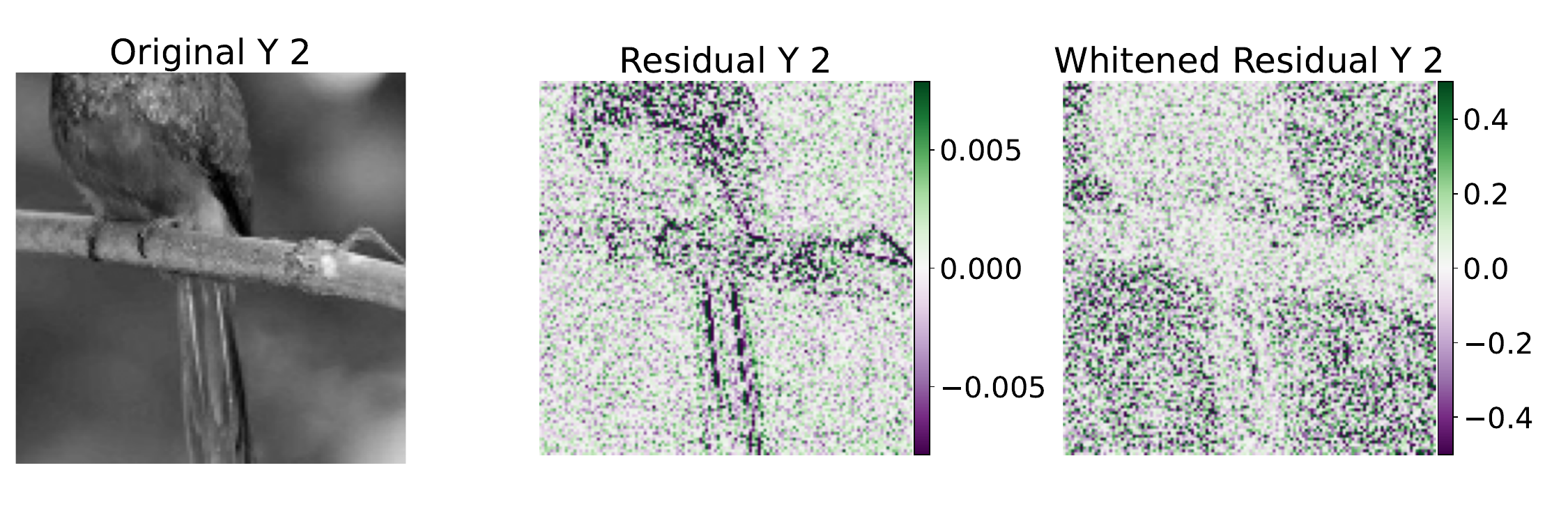}
        \caption{Whitened Residual Y Channel 1/2 Scale, Translation Level 4}
    \end{subfigure}

    \begin{subfigure}[b]{\textwidth}
        \centering
        \includegraphics[width=0.6\linewidth]{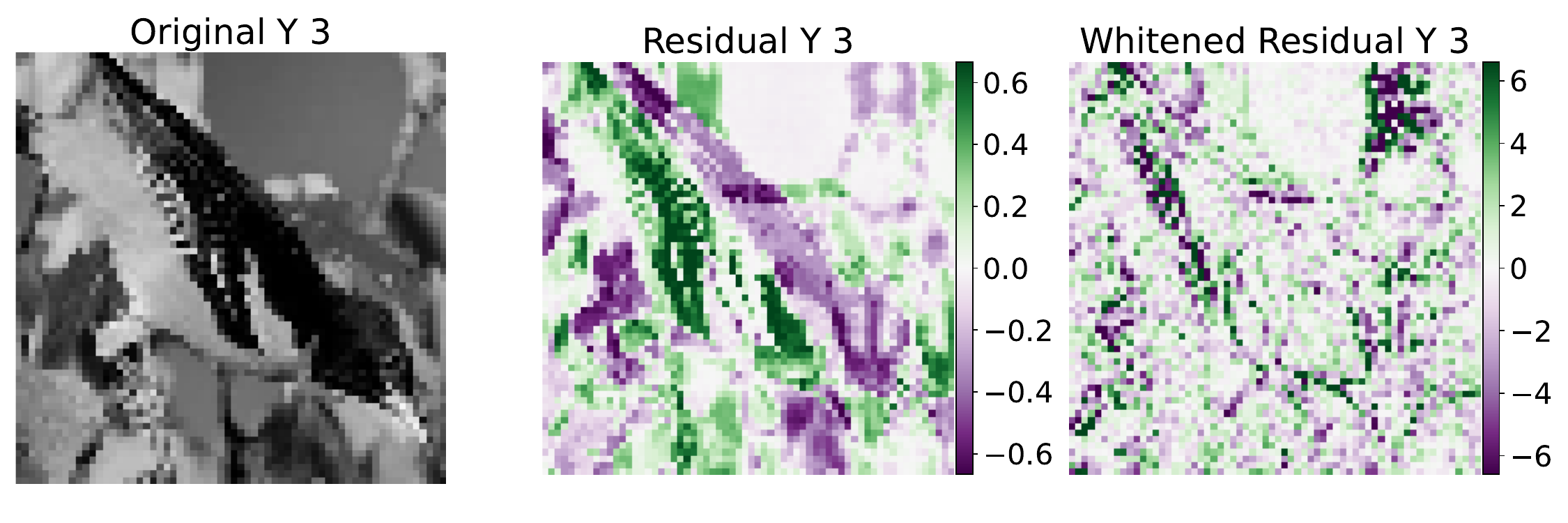}
        \caption{Whitened Residual Y Channel 1/4 Scale, Scaling + Translation + Rotation, Level 5}
    \end{subfigure}

    \begin{subfigure}[b]{\textwidth}
        \centering
        \includegraphics[width=0.6\linewidth]{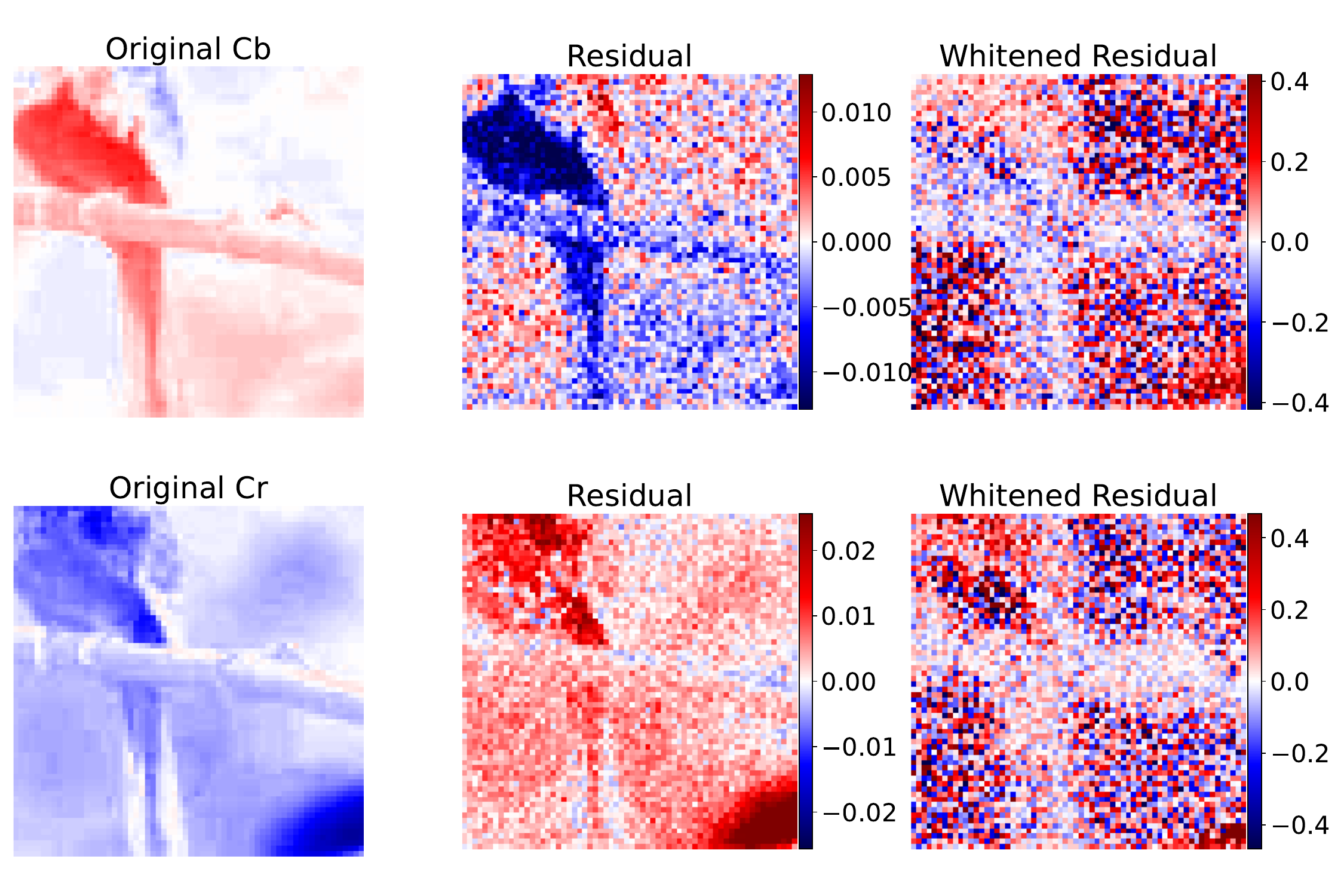}
        \caption{Example Colour Whitened Residuals, Colour Transformation Level 1}
        \label{fig:sample_colour_residu}
    \end{subfigure}

    \caption{Whitened residuals for SUSSPieapp-RH across different scales and color transformations, corresponding to all separate components of the SUSS score (see Figure \ref{fig:overview}). The input image transformations are similar to those used in the training dataset, spanning slightly beyond the just noticeable difference threshold (refer to Section~\ref{app:data}). The residual shown in the middle highlights differences relative to the original image, while the whitened residual represents the output of the learned model after removing perceptually similar structures. Examples show mainly noise in the background as the remaining differences, indicating that invariances to these transformations have been successfully learned.}
\end{figure*}

\vspace{1em}
\begin{figure*}[t]
    \begin{subfigure}[b]{\textwidth}
        \centering
        \includegraphics[width=0.7\textwidth]{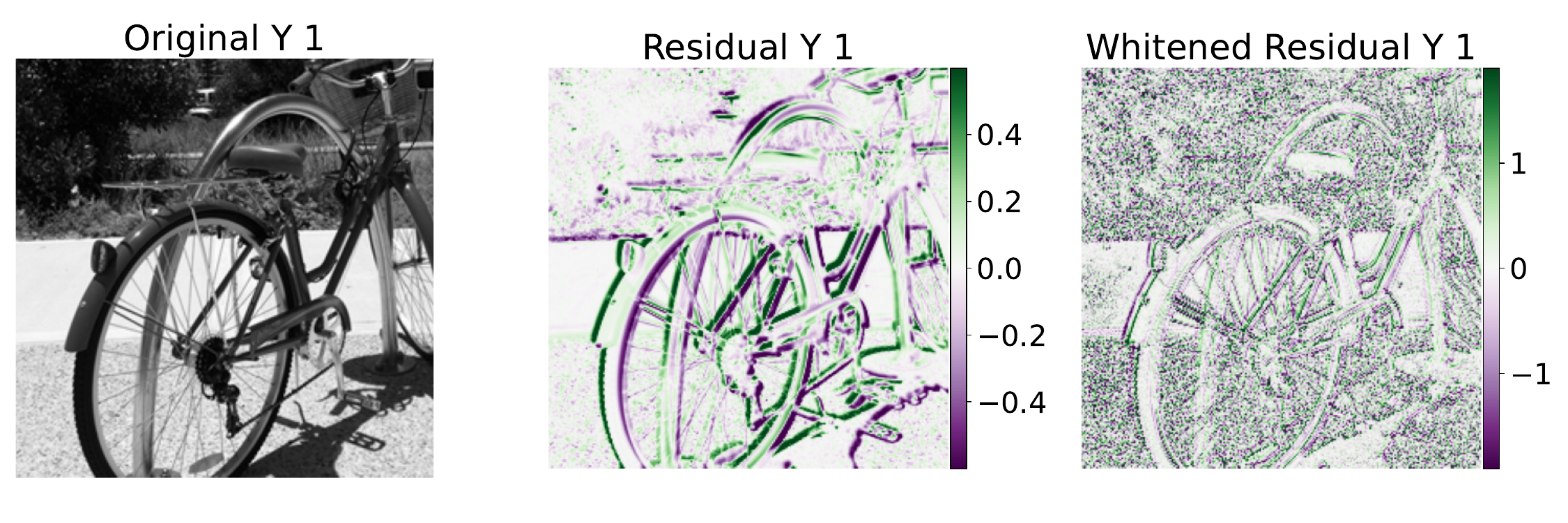}
        \caption{SUSSBase Whitened Residual}
    \end{subfigure}
    \begin{subfigure}[b]{\textwidth}
        \centering
        \includegraphics[width=0.7\linewidth]{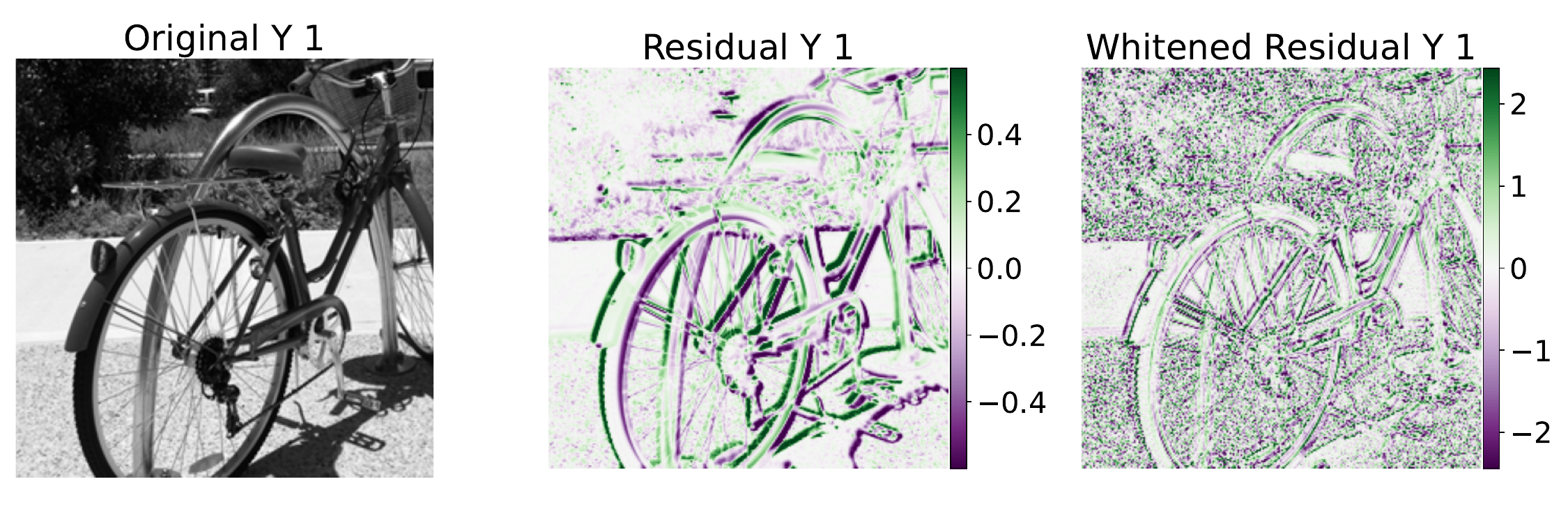}
        \caption{SUSSPieApp-RH Whitened Residual}
    \end{subfigure}
    \begin{subfigure}[b]{\textwidth}
        \centering
        \includegraphics[width=0.7\linewidth]{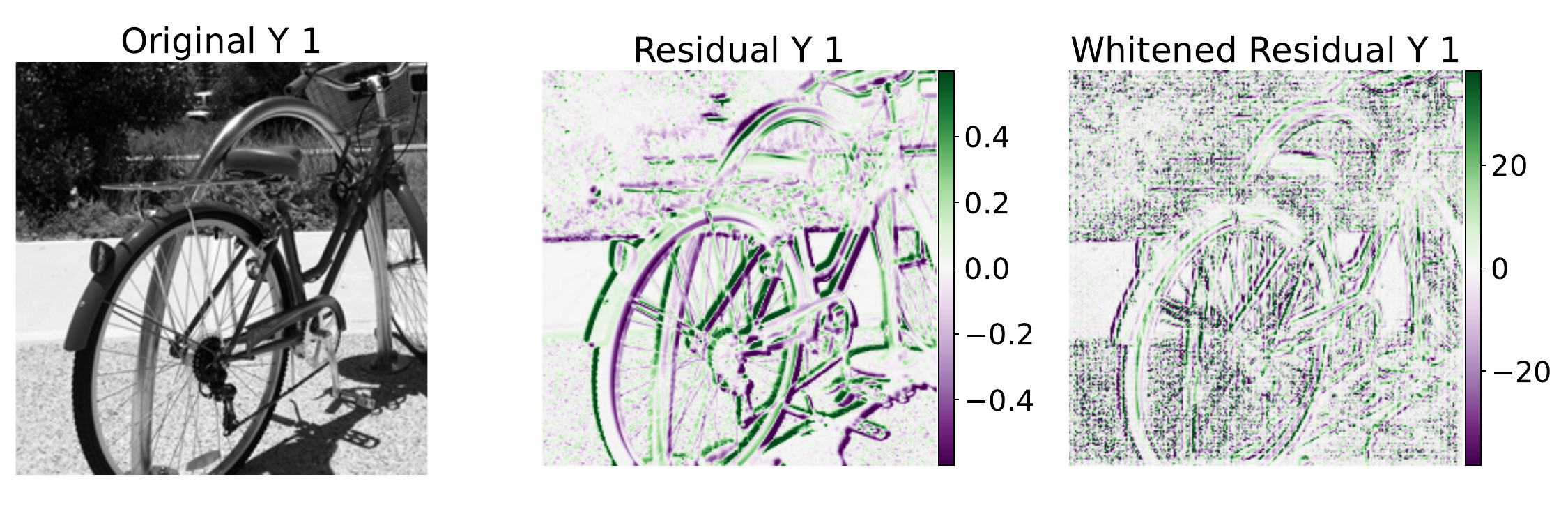}
        \caption{SUSSBAPPS-RH Whitened Residual}
    \end{subfigure}
    \begin{subfigure}[b]{\textwidth}
        \centering
        \includegraphics[width=0.7\linewidth]{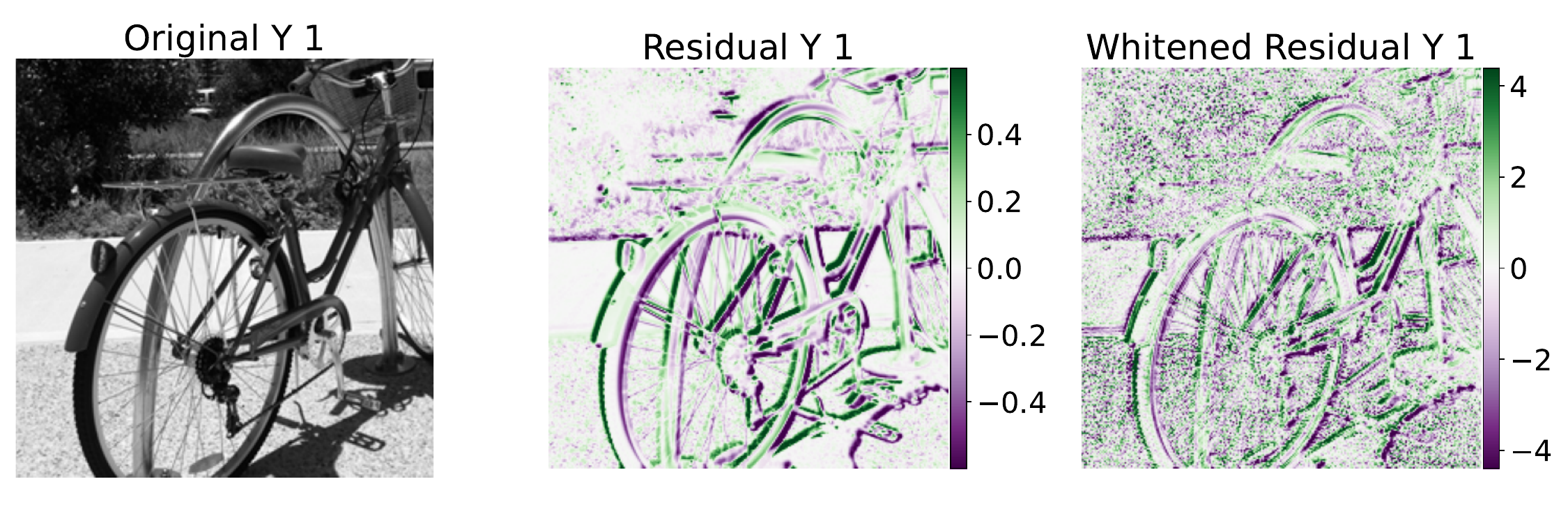}
        \caption{SUSSPIPAL-RH Whitened Residual}
    \end{subfigure}
    \caption{Whitened residual for different fine-tuned SUSS variants, illustrating the impact of ranking and human judgment losses. The transformed image(image pair from PieAPP dataset) corresponds to a chromatic aberration distortion with MOS = 0.175.}
\end{figure*}

\begin{figure*}[t]
    \begin{subfigure}[b]{\textwidth}
        \centering
        \includegraphics[width=\textwidth]{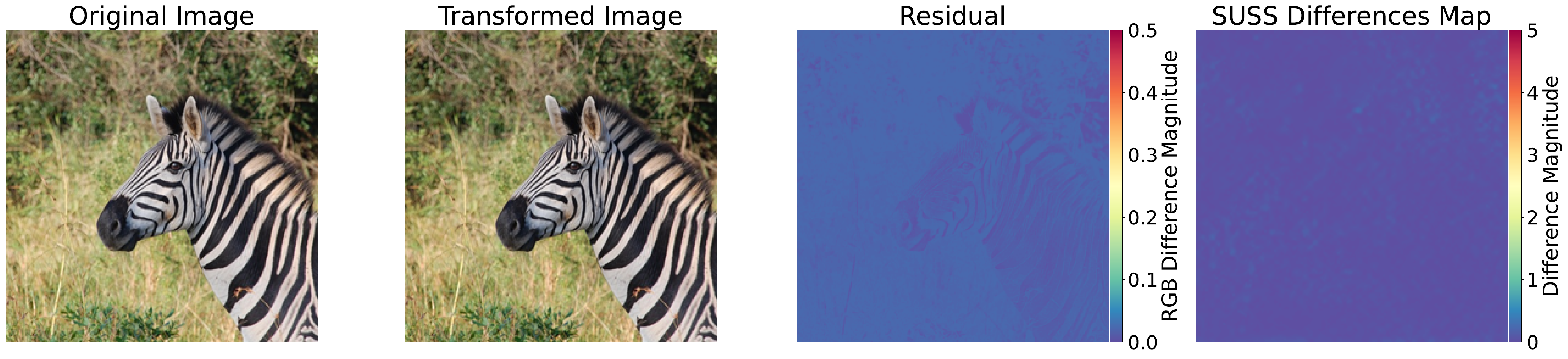}
        \caption{MOS score: 4.8, SUSS score: 3.4522}
    \end{subfigure}
    \begin{subfigure}[b]{\textwidth}
        \centering
        \includegraphics[width=1\linewidth]{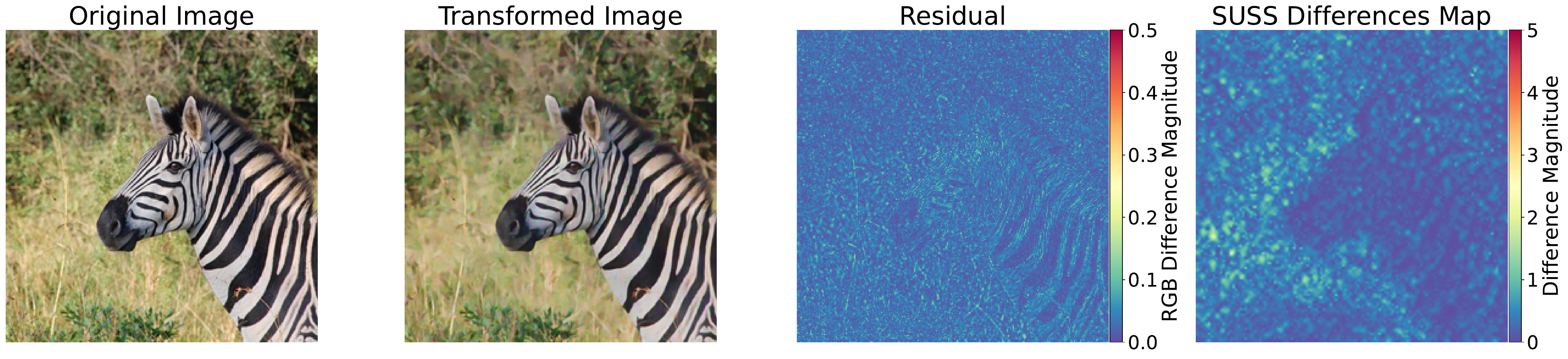}
        \caption{MOS score: 4.07, suss score: 2.6204}
    \end{subfigure}
    \begin{subfigure}[b]{\textwidth}
        \centering
        \includegraphics[width=1\linewidth]{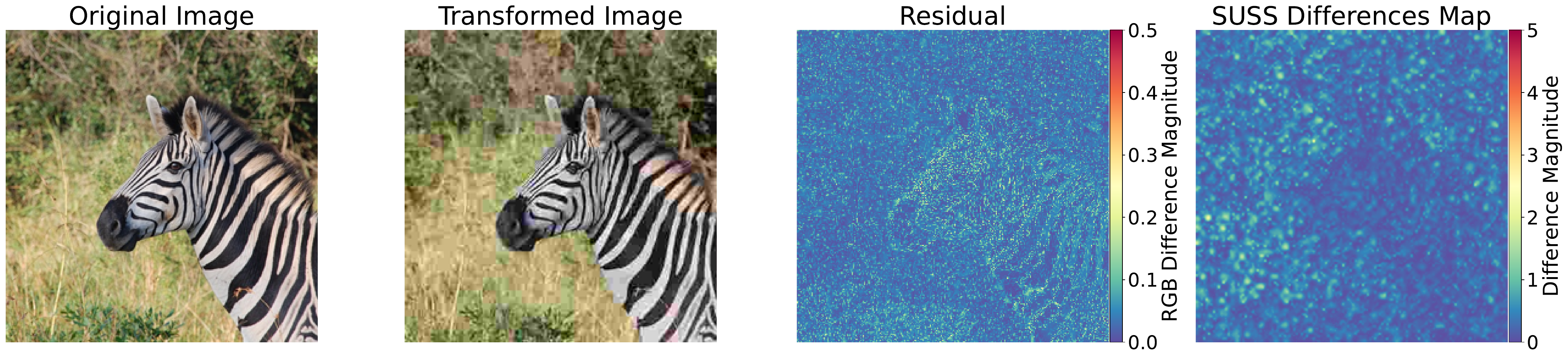}
        \caption{MOS score: 2.9, suss score: 2.2129}
    \end{subfigure}
    \begin{subfigure}[b]{\textwidth}
        \centering
        \includegraphics[width=1\linewidth]{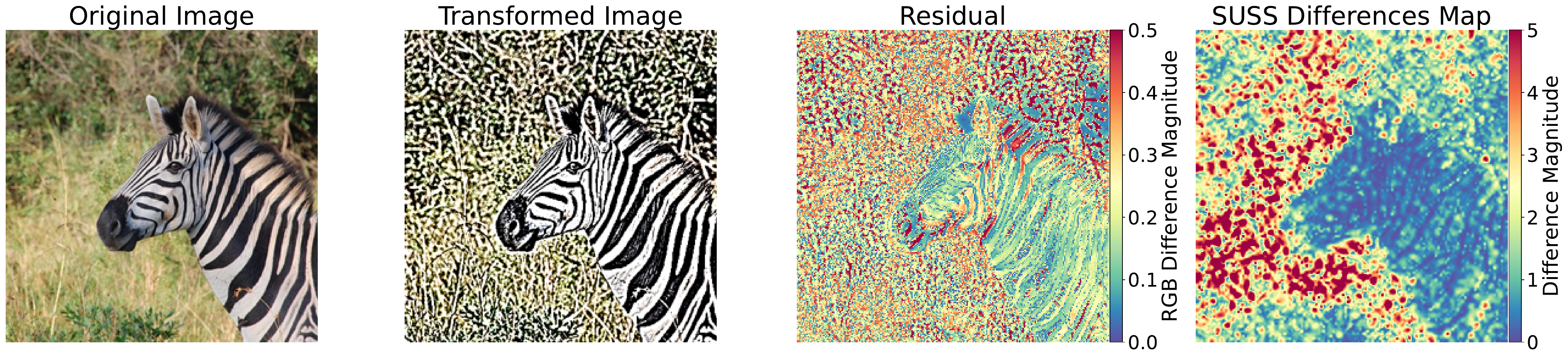}
        \caption{MOS score: 1.3, SUSS score:-7.5582}
    \end{subfigure}
    \caption{SUSS maps for distortions with MOS scores ranging from 5 (imperceptible differences) to 1 (clearly noticeable differences), using image pairs from the KADID-10k dataset. The SUSS maps (shown for SUSSPieApp-RH) represent the weighted squared sum of whitened residuals and highlight regions that the SUSS metric has learned to consider perceptually important.}
\end{figure*}

\begin{figure*}[t]
    \begin{subfigure}[b]{\textwidth}
        \centering
        \includegraphics[width=\textwidth]{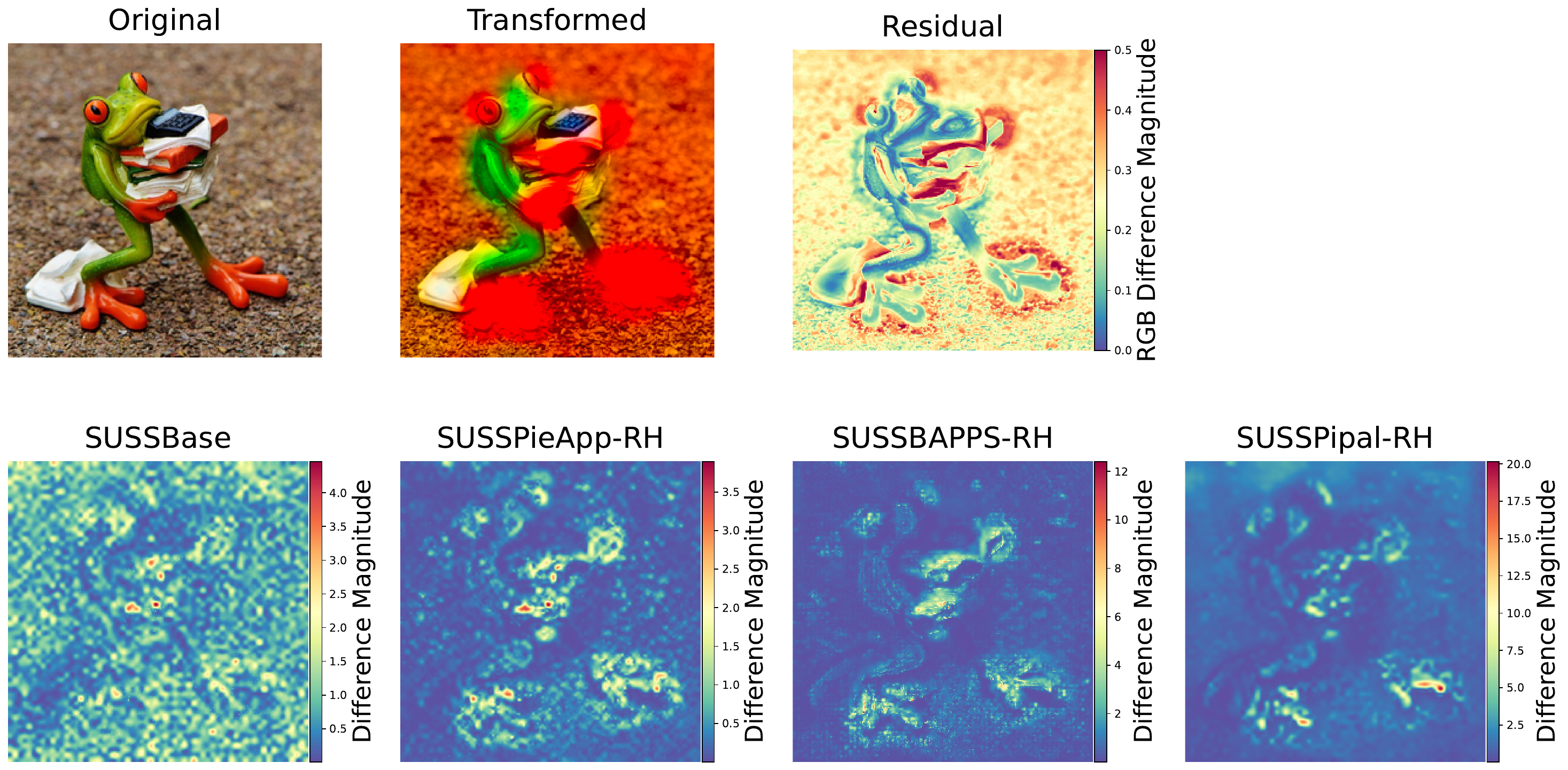}
        \caption{SUSS maps (bottom row) show that all variants correctly identify the red dots on the feet and background as the most perceptually different regions, while the frog itself remains structurally similar. Differences exist between variants: e.g., SUSSBase assigns higher relevance to background structural details compared to the other models, possibly due to a stronger weighting of the original-scale Y-channel component.}
    \end{subfigure}
    
    \vspace{5em}
    
    \begin{subfigure}[b]{\textwidth}
        \centering
        \includegraphics[width=\textwidth]{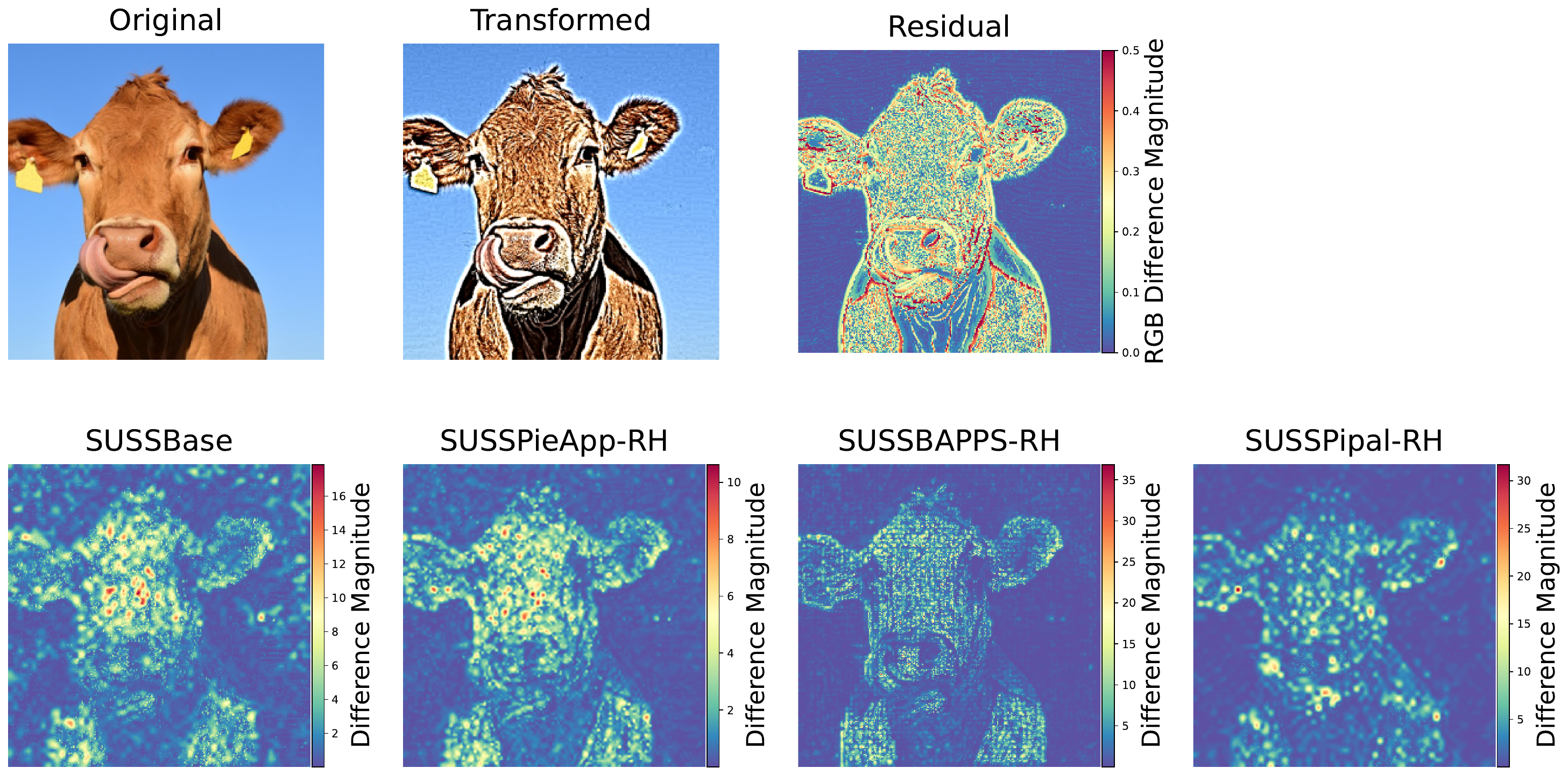}
        \caption{SUSS maps illustrate differences in learned structural emphasis across model variants. All models recognise similarity in the cow’s overall outline and key features such as the eyes; however, they differ in the degree to which fine texture and background details are considered perceptually relevant.}
        
    \end{subfigure}
    
    \caption{SUSS maps for different fine-tuned SUSS variants on examples from the KADID-10k dataset. While the overall spatial patterns are qualitatively comparable, the absolute magnitudes of the maps are not directly comparable due to variant-specific weighting schemes.}
\end{figure*}

\begin{figure*}[t]
    \begin{subfigure}[b]{\textwidth}
        \centering
        \includegraphics[width=\textwidth]{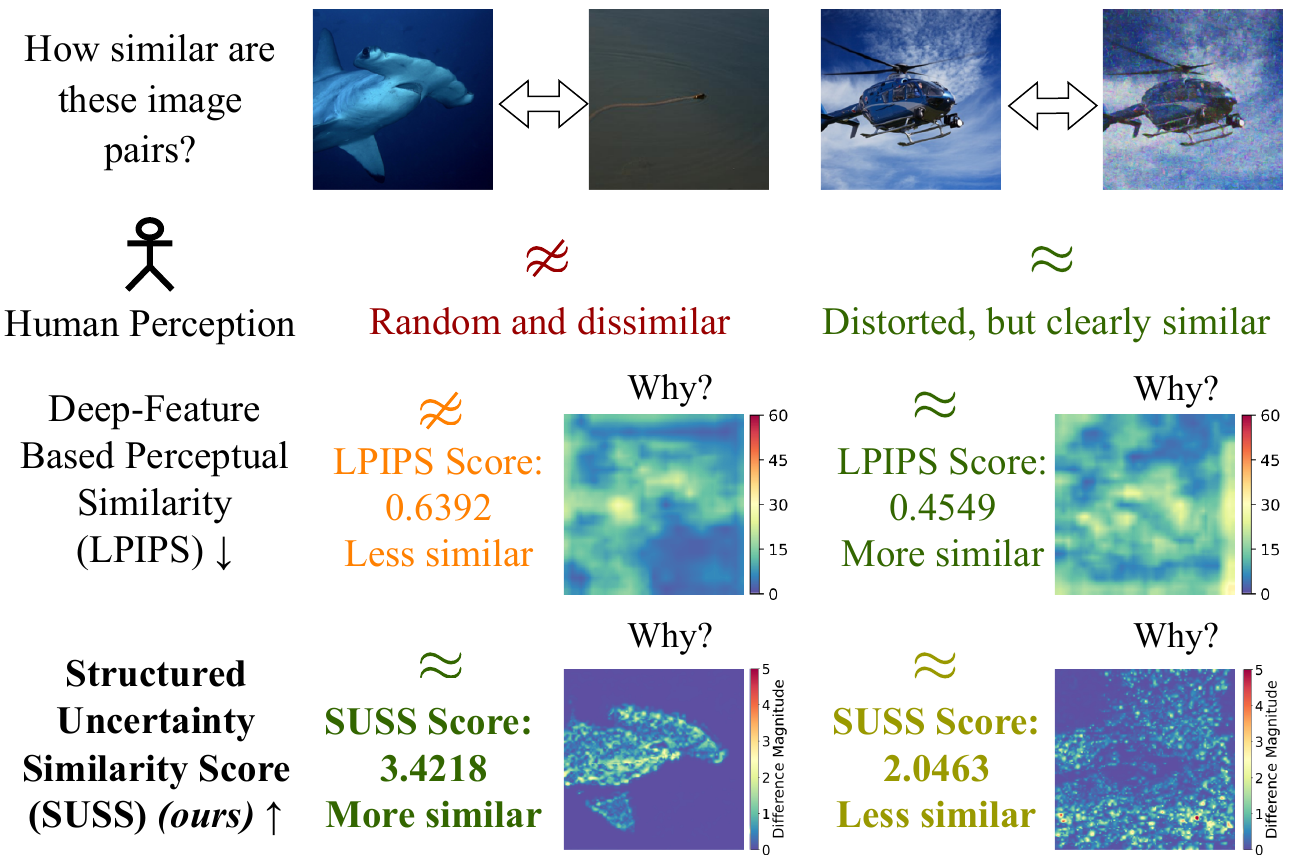}
    \end{subfigure}
    
    \caption{SUSS failure case: The SUSS judgment does not align with human perception, as the image pair on the right is judged as more similar than the one on the left. The SUSS maps explain the faulty judgment: although structural differences in the shark are detected in the left example, their magnitude is small, while background structures in the right pair contribute more strongly to the overall score. Samples were obtained using the same procedure as the LPIPS failure case in Fig.~\ref{fig:problem}: using ImageNet, a query image was compared against a large random subset of images from classes different from the reference one, and the pair with the highest similarity score was selected; for KADID-10k examples, the transformation yielding the lowest similarity score for a random reference image was chosen.}
    \label{fig:sample_fail}
\end{figure*}
\begin{figure*}[t]
    \centering
    \begin{subfigure}[b]{0.45\linewidth}
        \centering
        \includegraphics[width=\linewidth]{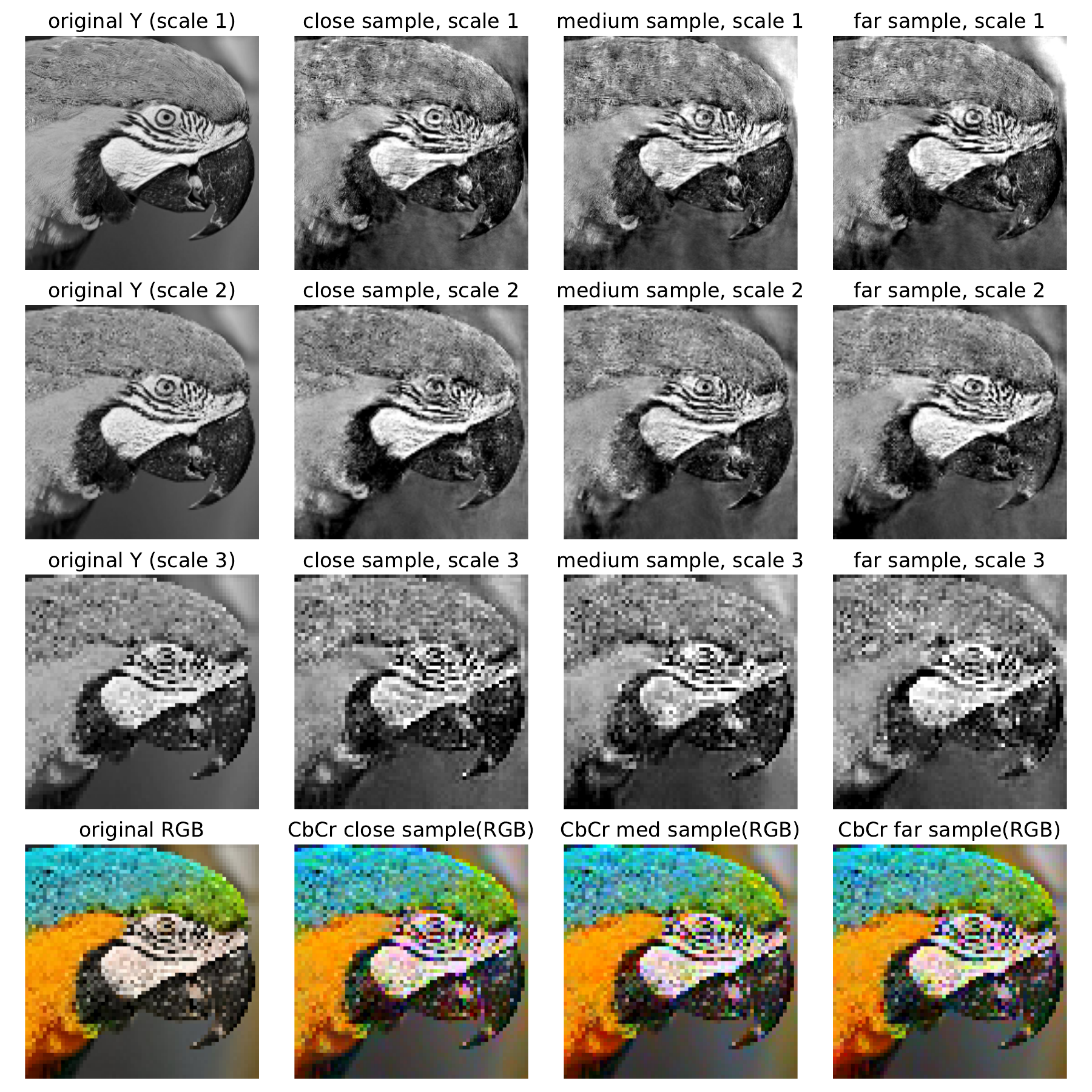}
        \caption{Example sample with far, close, medium samples and added mean}
        \label{fig:sample_mean}
    \end{subfigure}
    \hfill
    \begin{subfigure}[b]{0.45\linewidth}
        \centering
        \includegraphics[width=\linewidth]{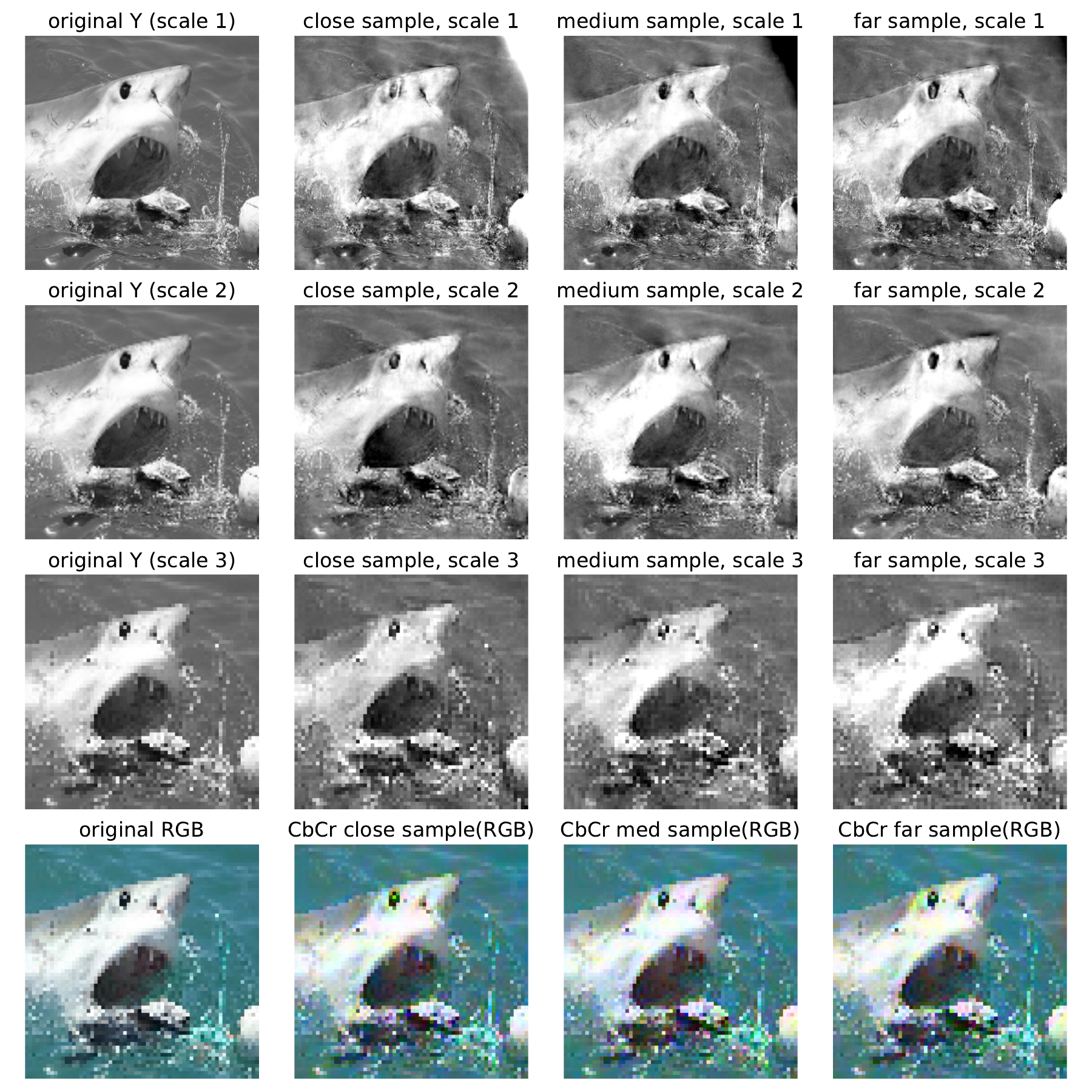}
        \caption{Example sample with far, close, medium samples and added mean}
        \label{fig:sample_mean2}
    \end{subfigure}
    \begin{subfigure}[b]{0.45\linewidth}
        \centering
        \includegraphics[width=\linewidth]{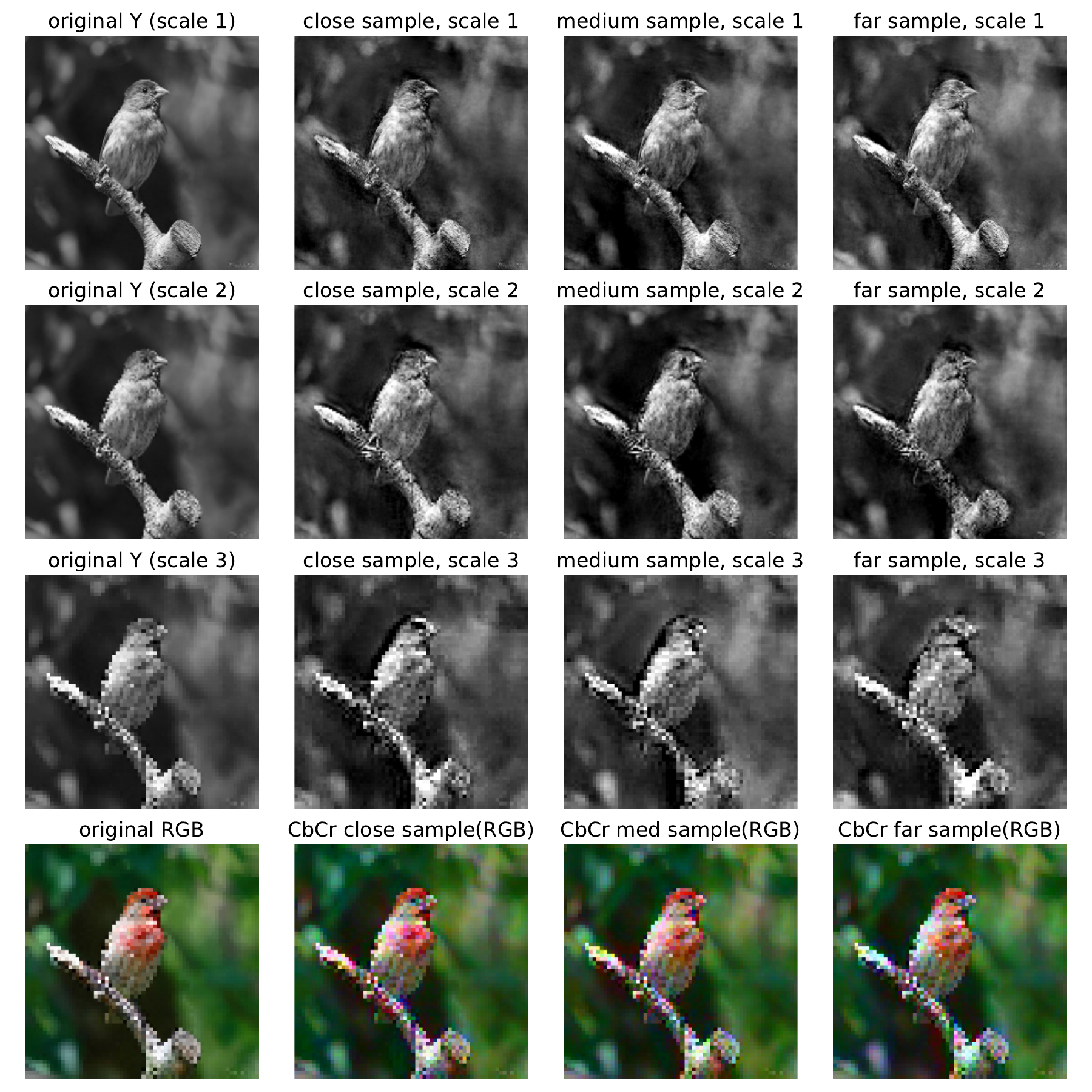}
        \caption{Example sample with far, close, medium samples and added mean}
        \label{fig:sample_mean}
    \end{subfigure}
    \hfill
    \begin{subfigure}[b]{0.45\linewidth}
        \centering
        \includegraphics[width=\linewidth]{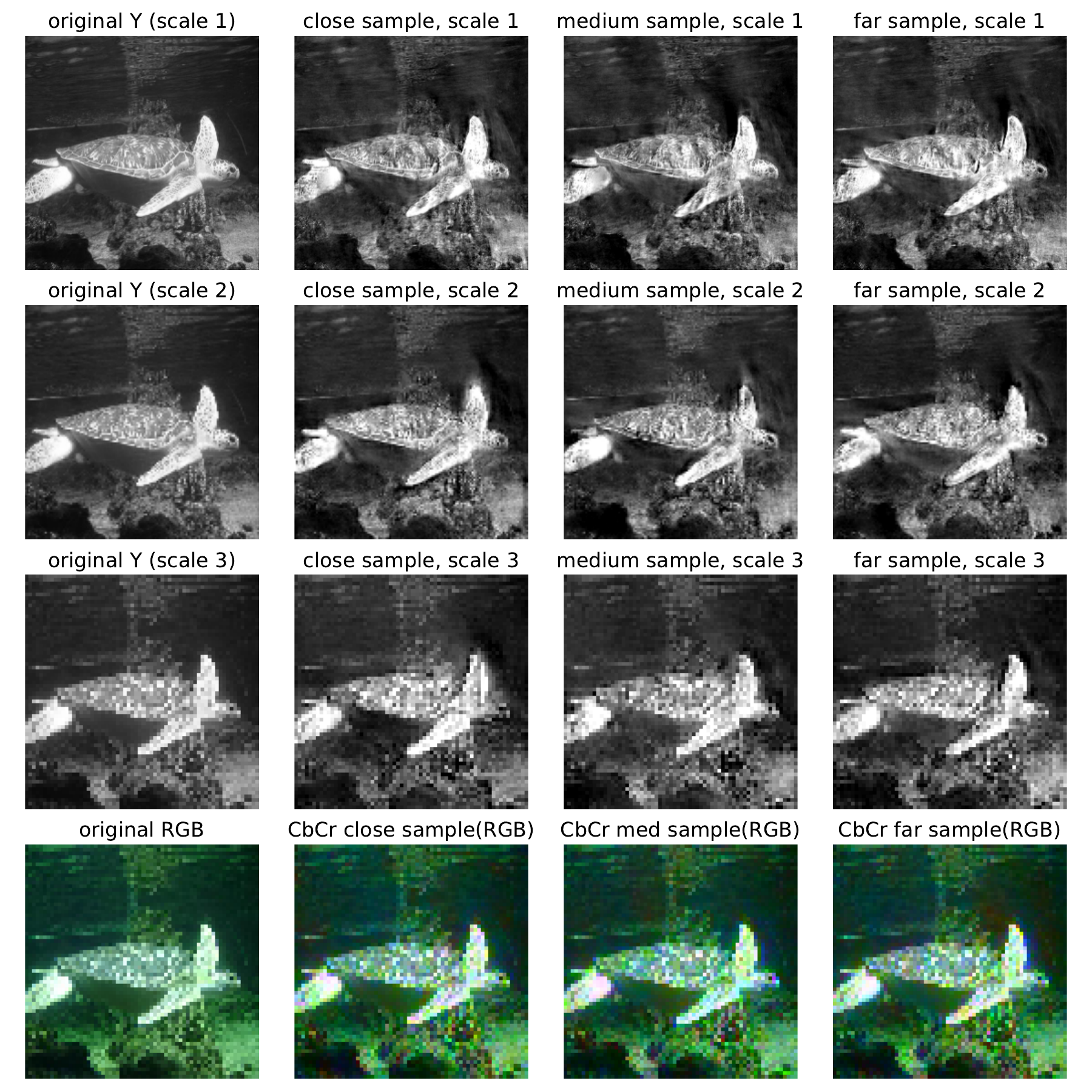}
        \caption{Example sample with far, close, medium samples and added mean}
        \label{fig:sample_mean2}
    \end{subfigure}
    \caption{Samples with \textbf{added mean} from SUSSPieApp-RH, drawn from all four components of the SUSS model (see Fig.~\ref{fig:overview}). Samples are selected based on log-$p$ scores to obtain close, medium, and far examples, illustrating the range of the perceptual space captured by the learned distributions. The bottom rows show samples generated from chrominance channels recombined with the original Y-channel at the corresponding scale to visualise the results in RGB space.}
    \label{fig:samples}
\end{figure*}
\begin{figure*}[t]
    \centering
    \begin{subfigure}[b]{0.45\linewidth}
        \centering
        \includegraphics[width=\linewidth]{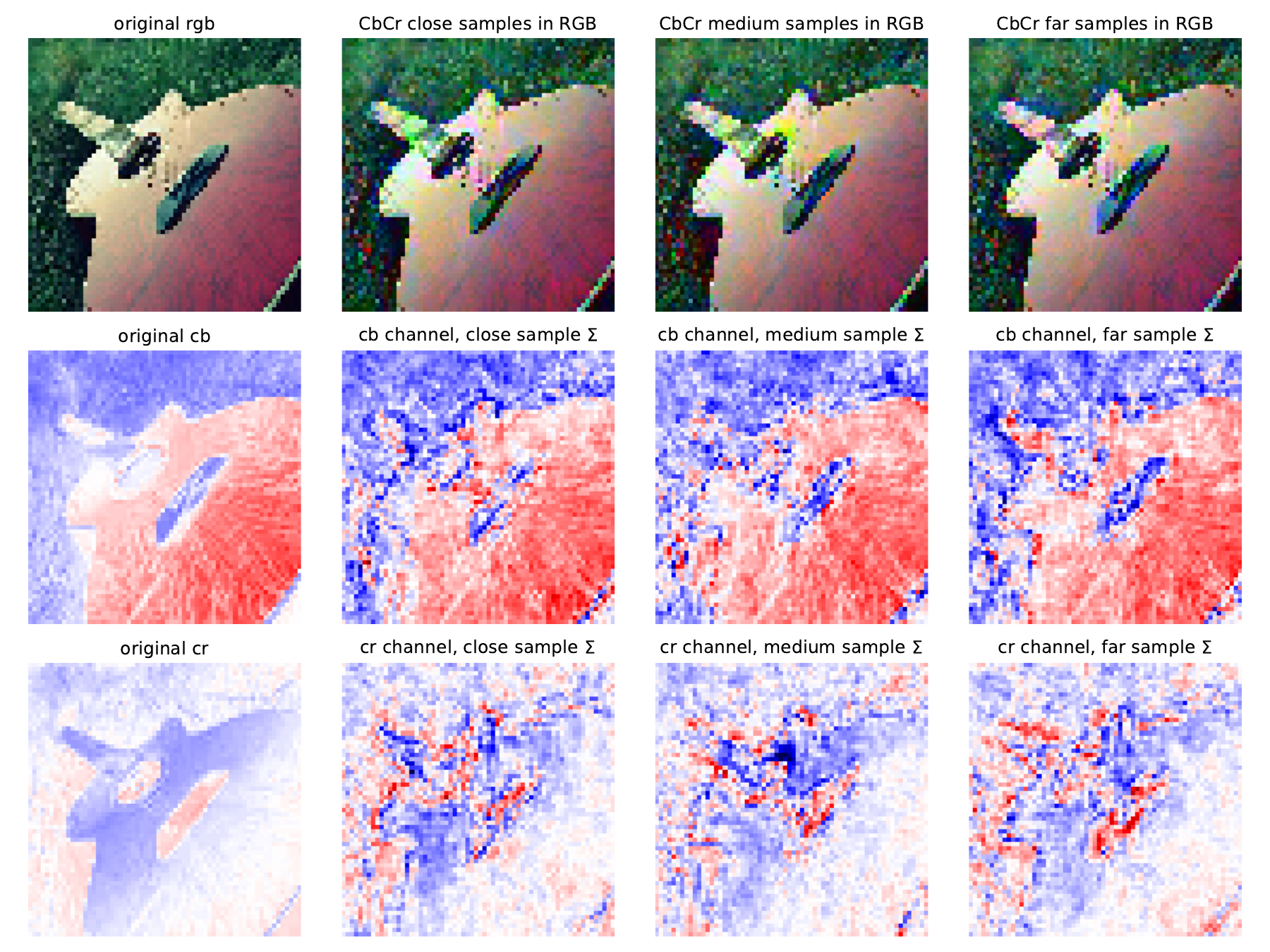}
        \caption{Example color sample with far, close, medium samples}
        \label{fig:sample_colour_a}
    \end{subfigure}
    \hfill
    \begin{subfigure}[b]{0.45\linewidth}
        \centering
        \includegraphics[width=\linewidth]{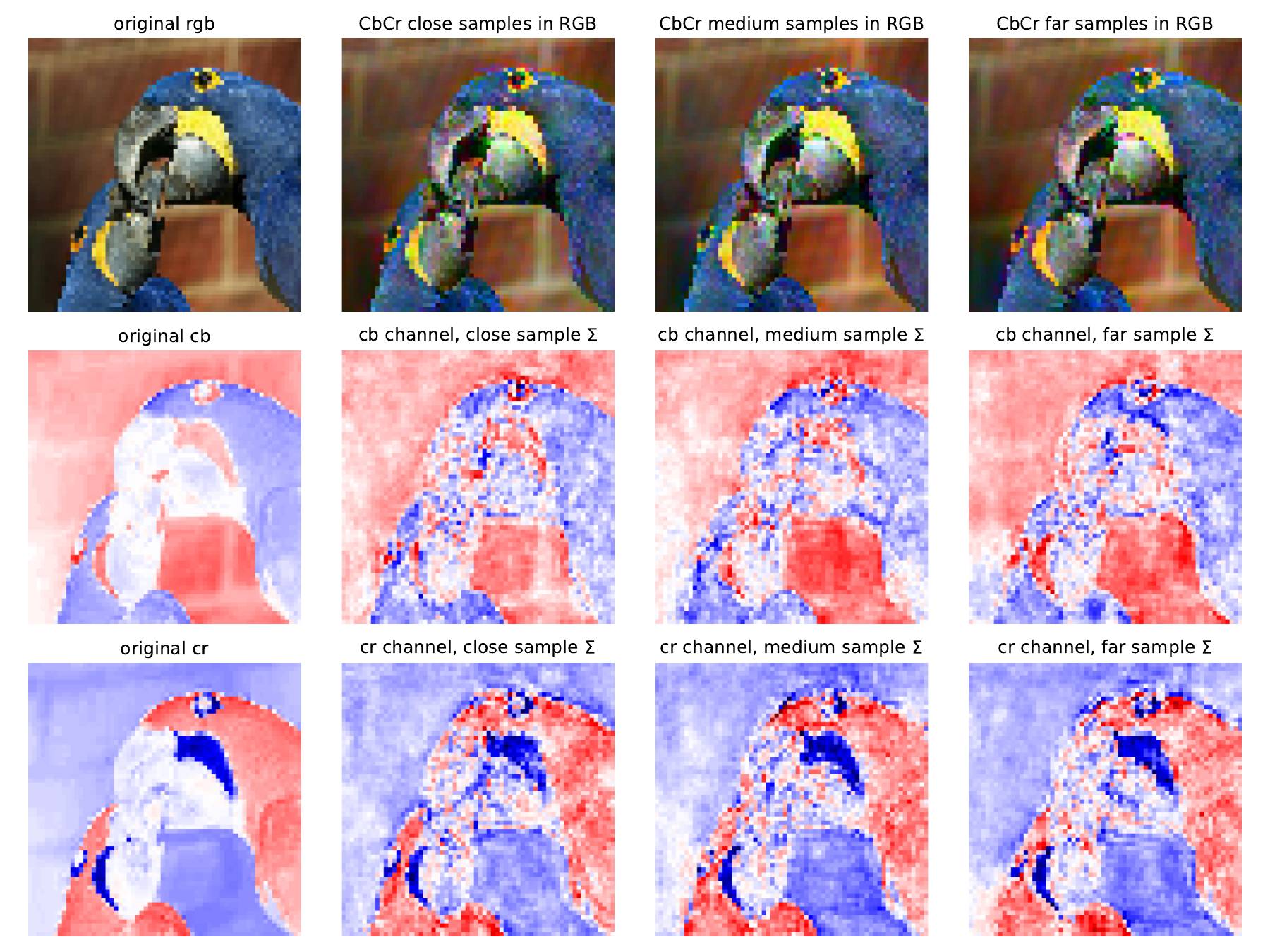}
        \caption{Example color sample with far, close, medium samples}
        \label{fig:sample_colour_b}
    \end{subfigure}

    \vspace{1em}
    
    \begin{subfigure}[b]{0.45\linewidth}
        \centering
        \includegraphics[width=\linewidth]{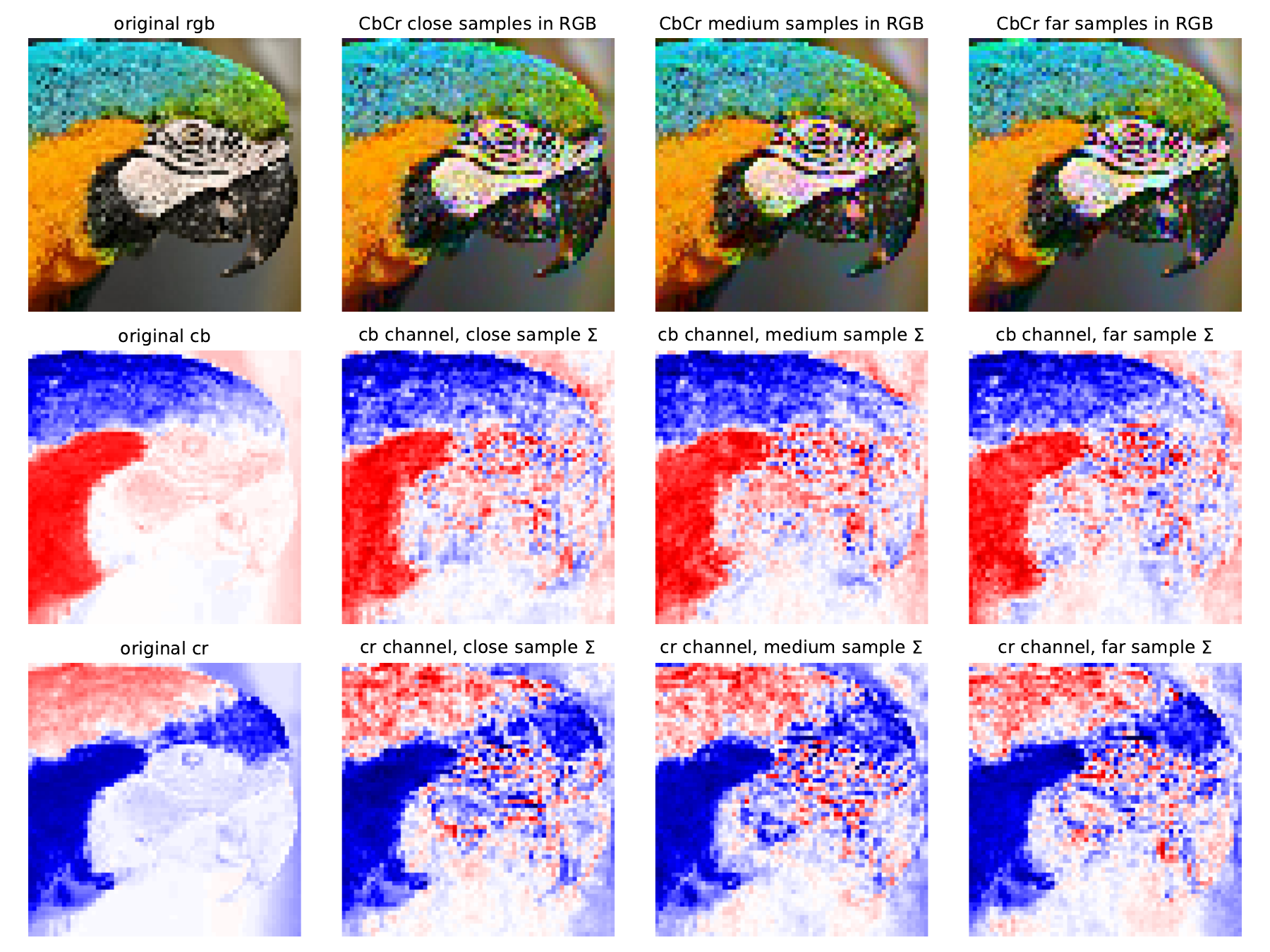}
        \caption{Example color sample with far, close, medium samples}
        \label{fig:sample_colour_a}
    \end{subfigure}
    \hfill
    \begin{subfigure}[b]{0.45\linewidth}
        \centering
        \includegraphics[width=\linewidth]{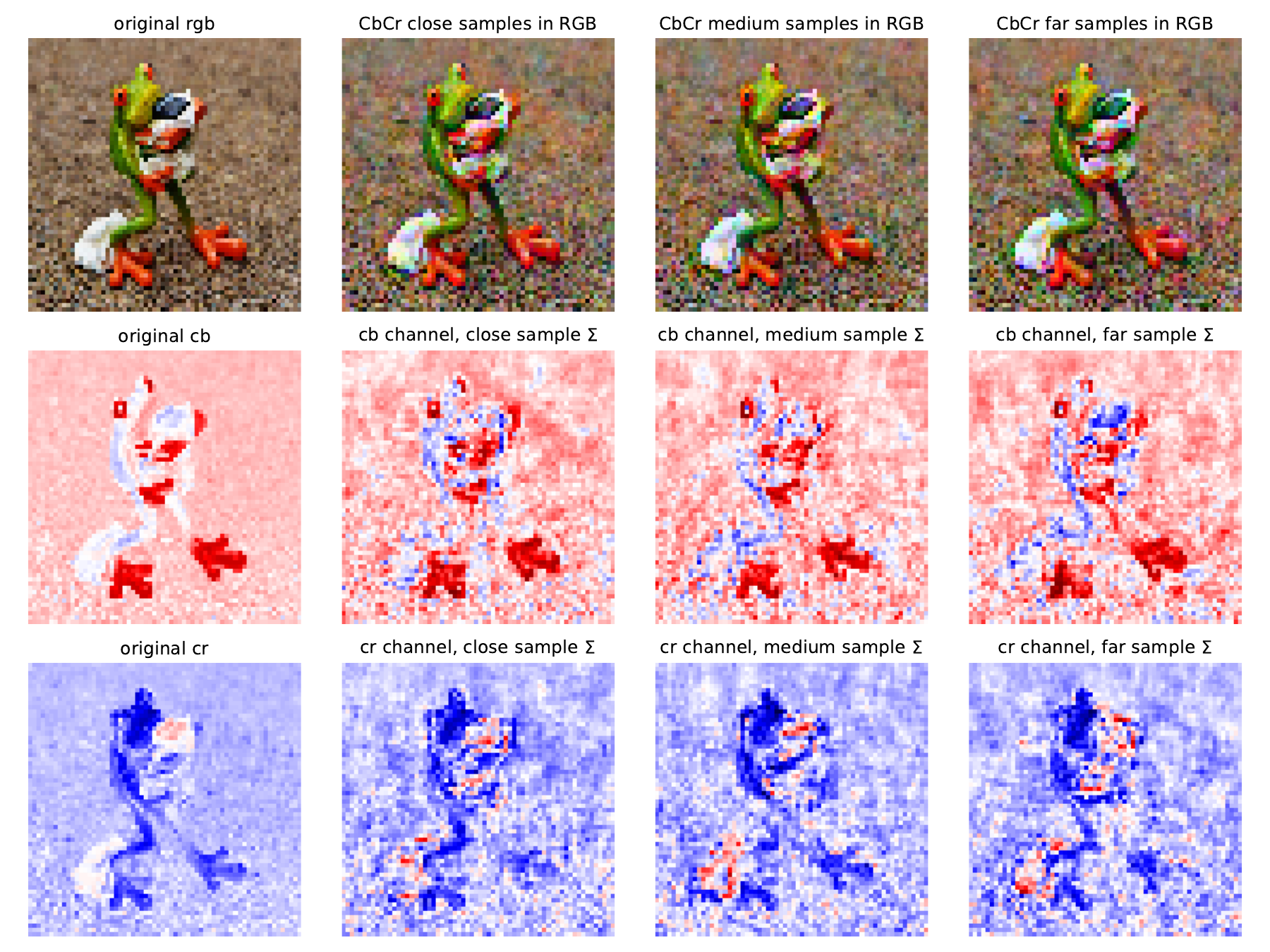}
        \caption{Example color sample with far, close, medium samples}
        \label{fig:sample_colour_b}
    \end{subfigure}

    \begin{subfigure}[b]{0.45\linewidth}
        \centering
        \includegraphics[width=\linewidth]{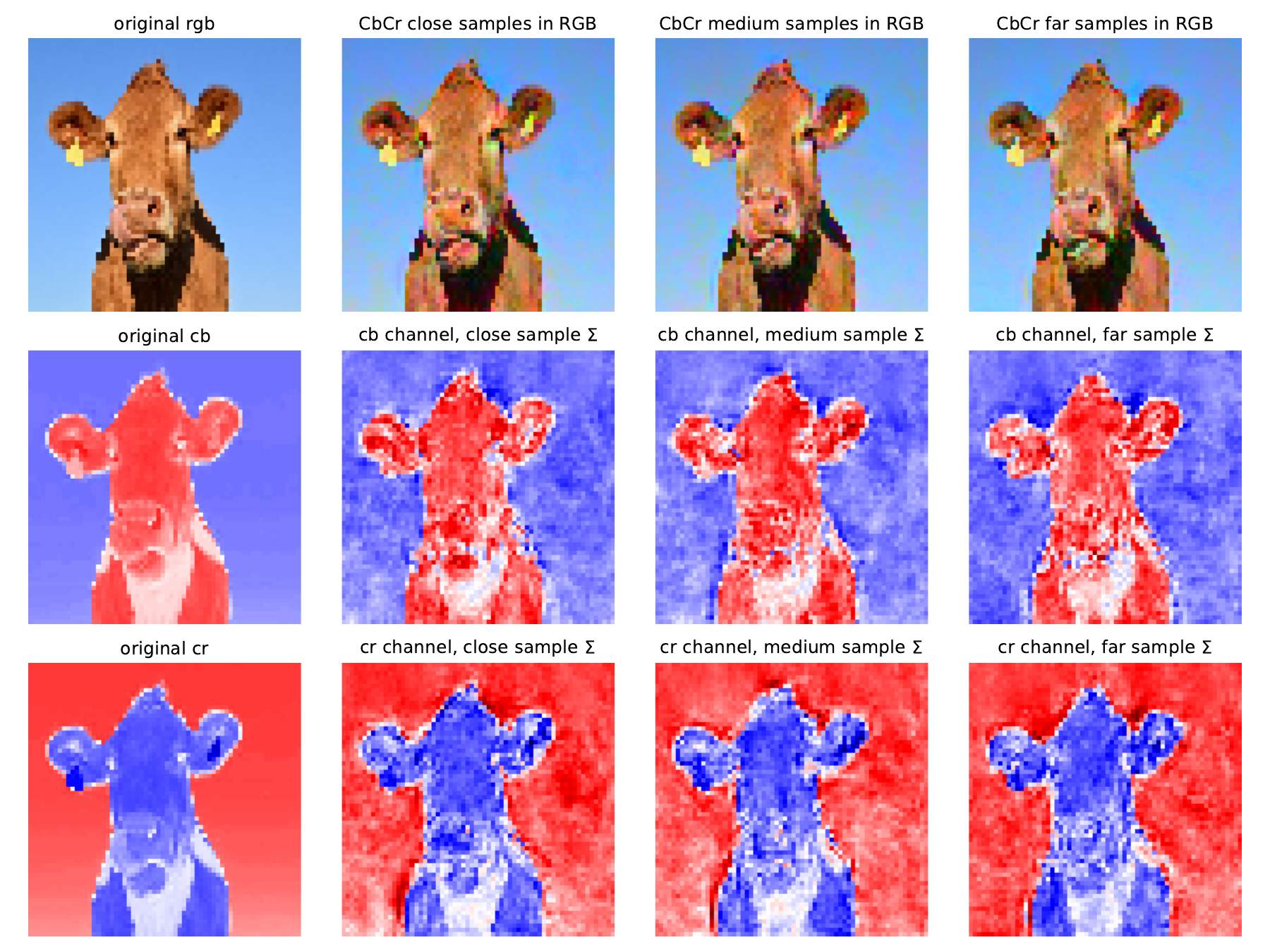}
        \caption{Example color sample with far, close, medium samples}
        \label{fig:sample_colour_a}
    \end{subfigure}
    \hfill
    \begin{subfigure}[b]{0.45\linewidth}
        \centering
        \includegraphics[width=\linewidth]{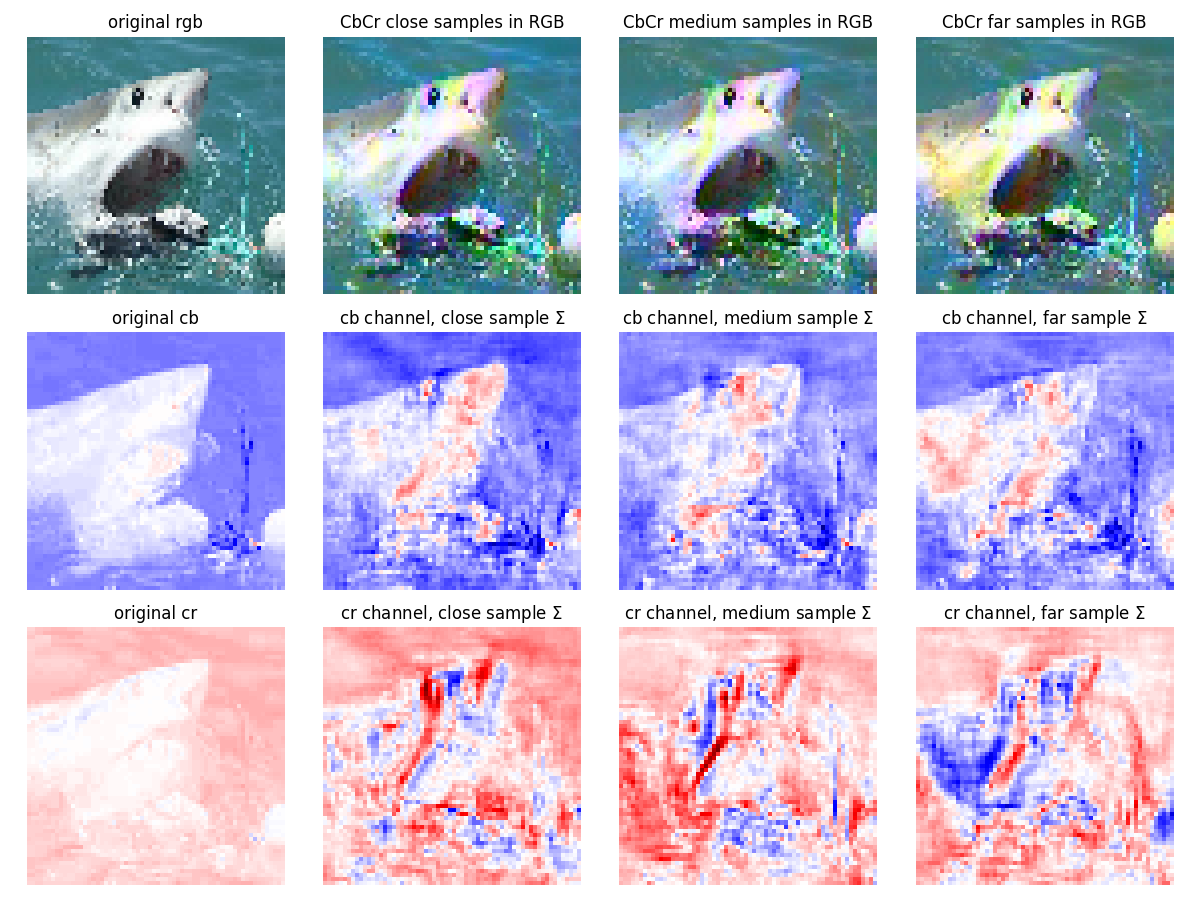}
        \caption{Example color sample with far, close, medium samples}
        \label{fig:sample_colour_b}
    \end{subfigure}
    \caption{Color Samples SUSSPieApp-RH. Here Cr/Cb samples are taken only from learned covariance components, without the mean added, illustrating which structures are learned and captured by the SUPN distributions. The top rows display Cr/Cb channel samples recombined with the original Y-channel image in RGB space.}

    \label{fig:samples}
\end{figure*}
\begin{figure*}[t]
    \centering
    \begin{subfigure}[b]{0.45\linewidth}
        \centering
        \includegraphics[width=\linewidth]{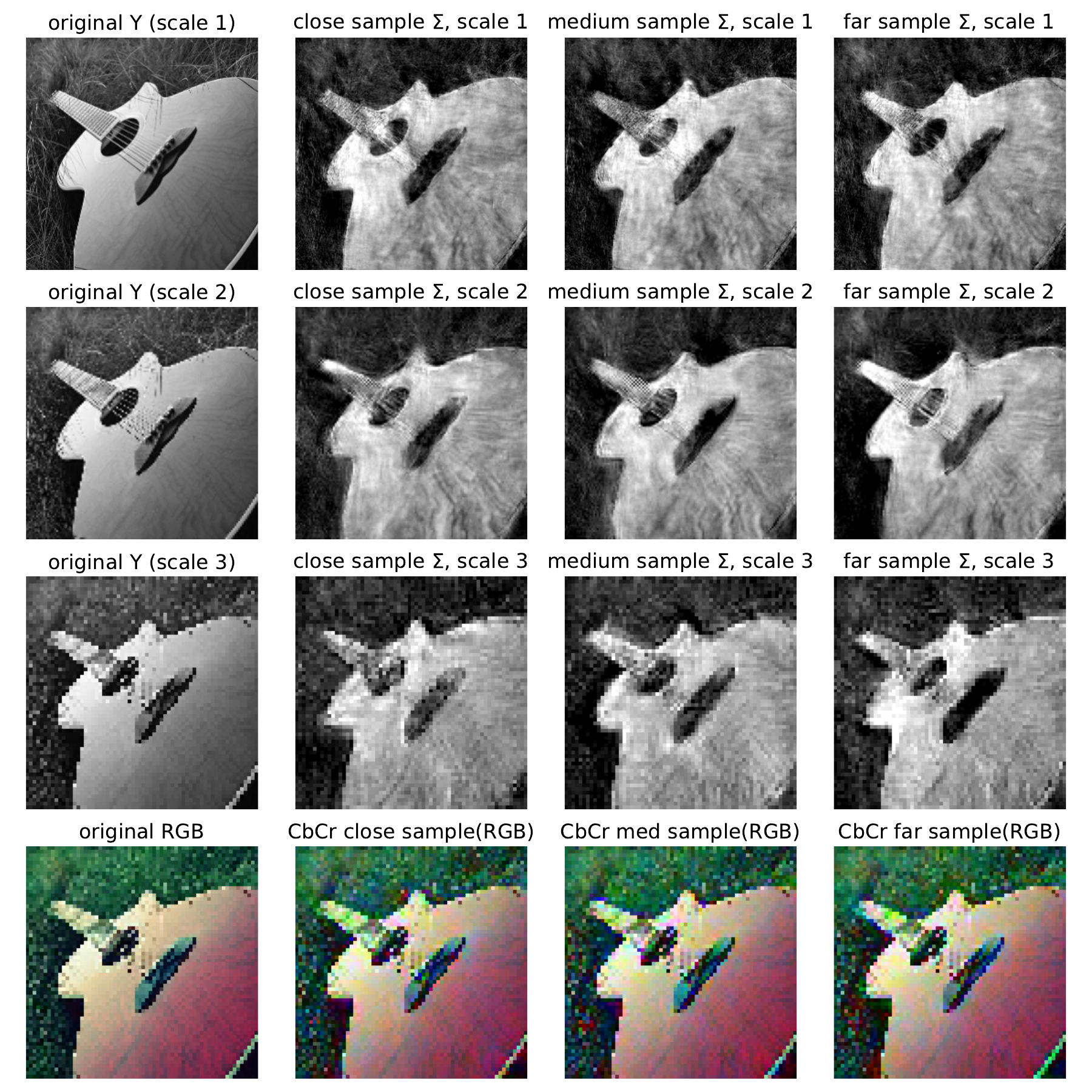}
        \caption{Example sample with far, close, medium samples and added mean}
        \label{fig:sample_mean}
    \end{subfigure}
    \hfill
    \begin{subfigure}[b]{0.45\linewidth}
        \centering
        \includegraphics[width=\linewidth]{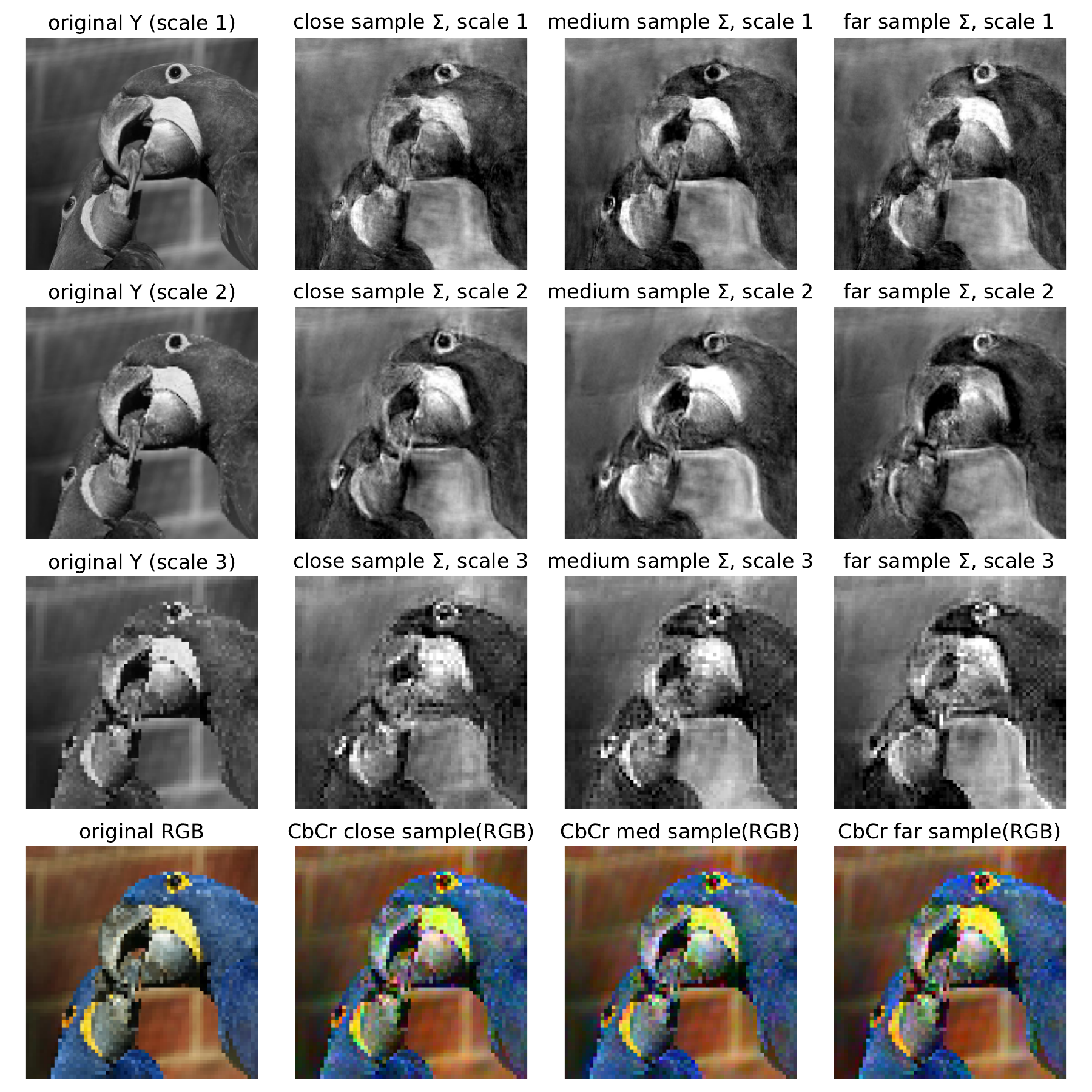}
        \caption{Example sample with far, close, medium samples(no mean)}
        \label{fig:parrotpie}
    \end{subfigure}
    \begin{subfigure}[b]{0.45\linewidth}
        \centering
        \includegraphics[width=\linewidth]{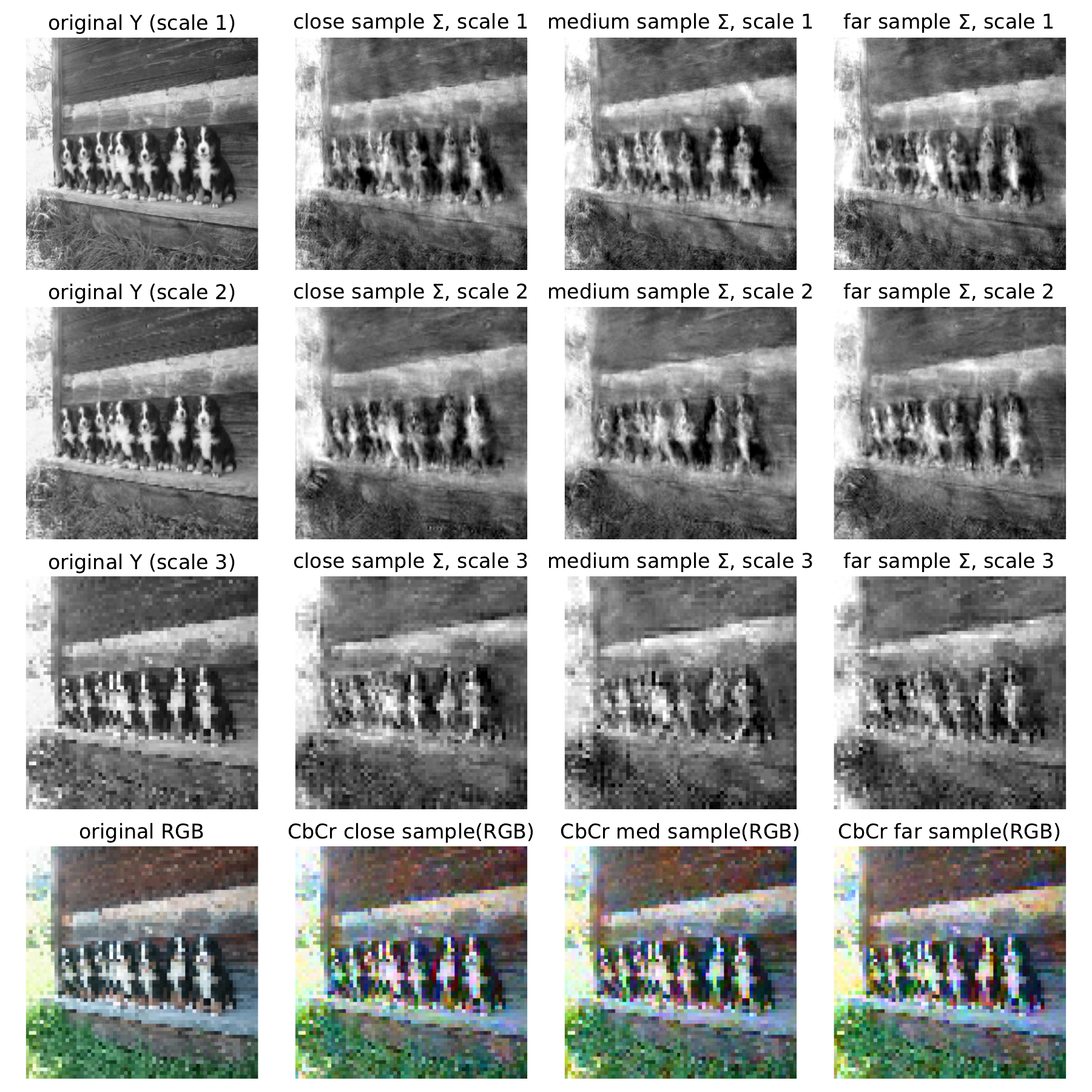}
        \caption{Example sample with far, close, medium samples(no mean)}
        \label{fig:sample_mean}
    \end{subfigure}
    \hfill
    \begin{subfigure}[b]{0.45\linewidth}
        \centering
        \includegraphics[width=\linewidth]{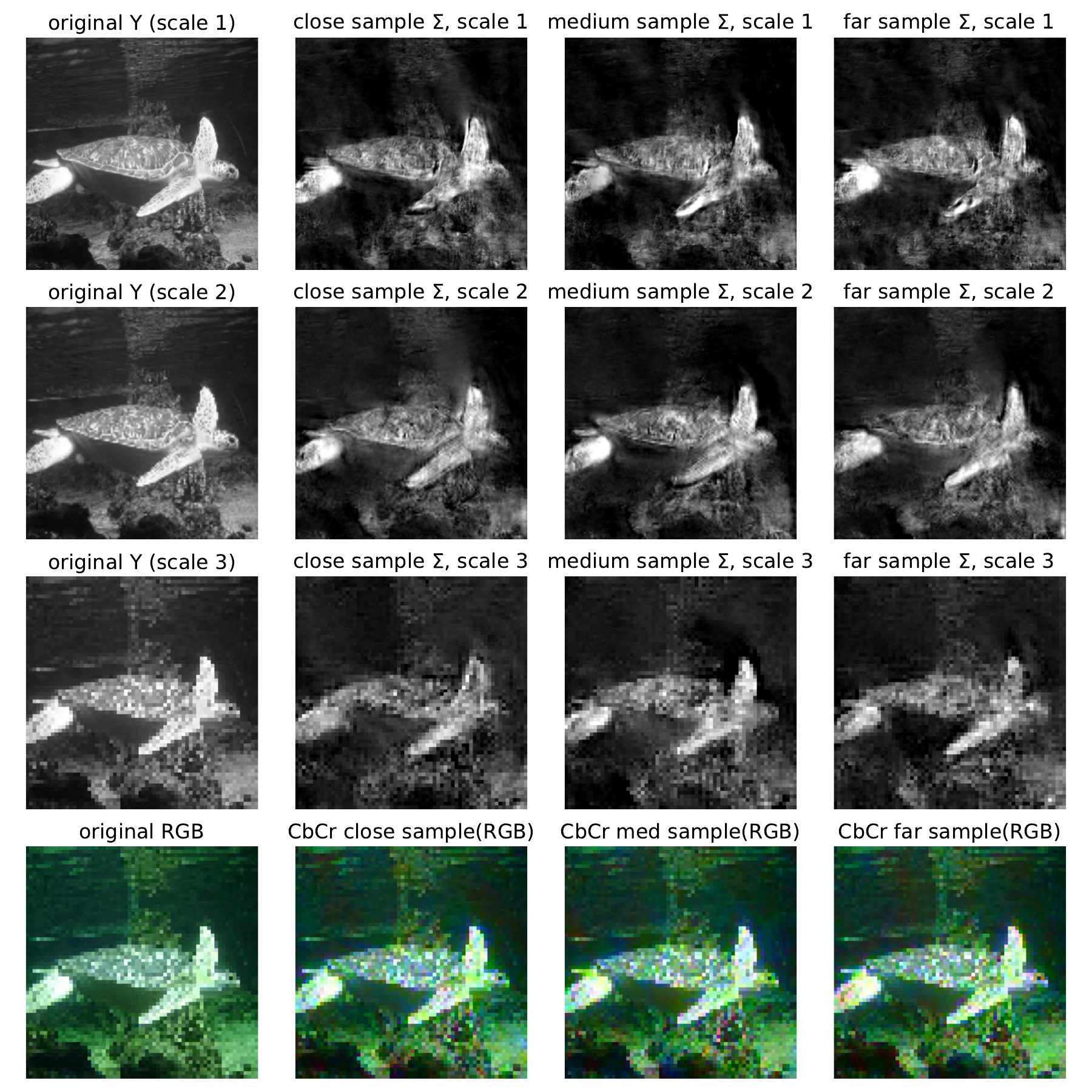}
        \caption{Example sample with far, close, medium samples(no mean) - compare with Fig. \ref{fig:sample_mean2} to see impact of mean added}   \label{fig:sample_mean2}
    \end{subfigure}

    \caption{Samples SUSSPieApp-RH: Example samples drawn from the covariance component of the inferred SUPN distributions for different SUSS components, \textbf{without the mean added}.}
    \label{fig:samples}
\end{figure*}
\begin{figure*}[t]
    \centering
    \begin{subfigure}[b]{0.45\linewidth}
        \centering
        \includegraphics[width=\linewidth]{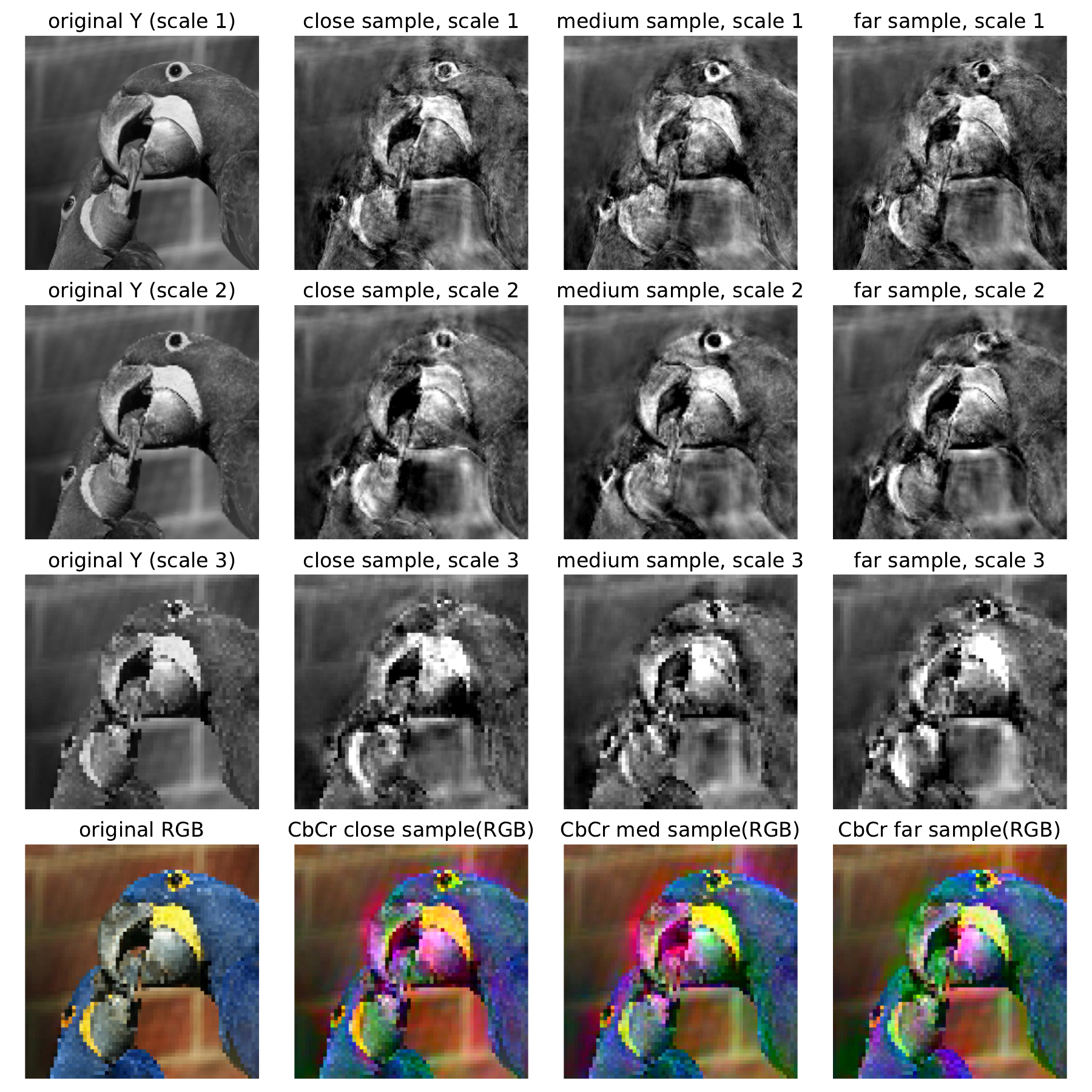}
        \caption{Example sample with far, close, medium samples(no mean added). Compare with \ref{fig:parrotpie} to see differences to PieApp-RH results}
        \label{fig:sample_mean}
    \end{subfigure}
    \hfill
    \begin{subfigure}[b]{0.45\linewidth}
        \centering
        \includegraphics[width=\linewidth]{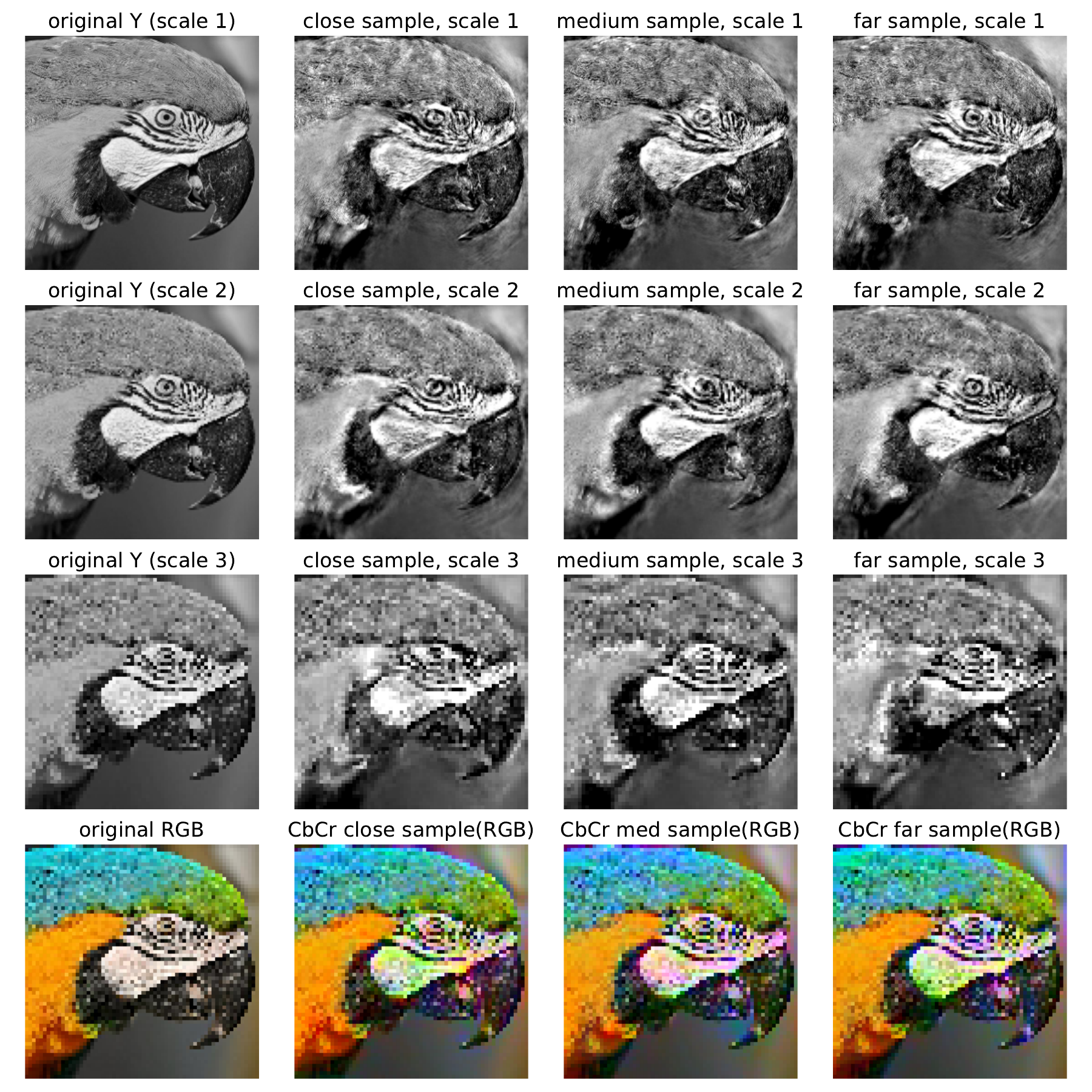}
        \caption{Example sample with far, close, medium samples(no mean added)}
        \label{fig:sample_mean2}
    \end{subfigure}
    \begin{subfigure}[b]{0.45\linewidth}
        \centering
        \includegraphics[width=\linewidth]{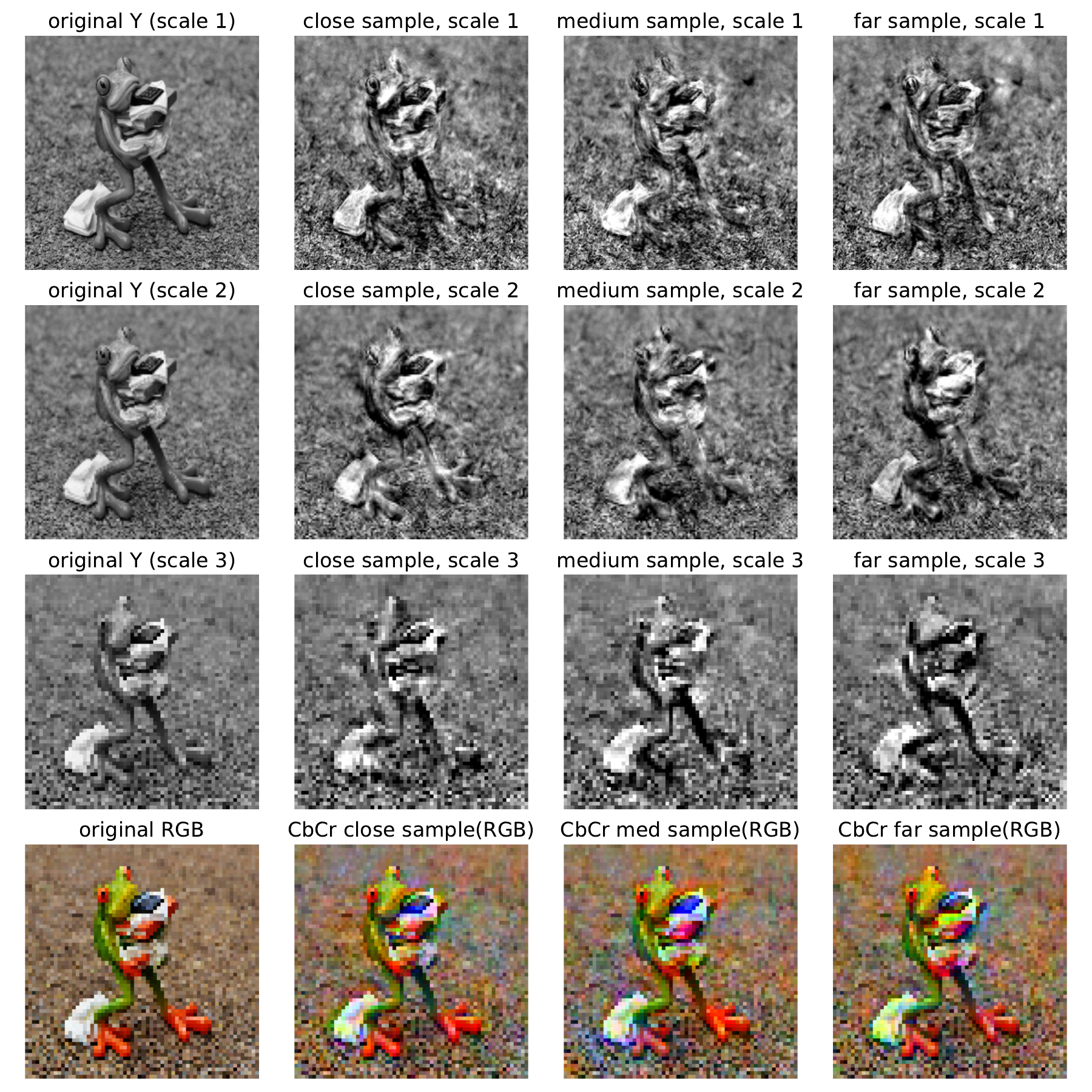}
        \caption{Example sample with far, close, medium samples(no mean added)}
        \label{fig:sample_mean}
    \end{subfigure}
    \hfill
    \begin{subfigure}[b]{0.45\linewidth}
        \centering
        \includegraphics[width=\linewidth]{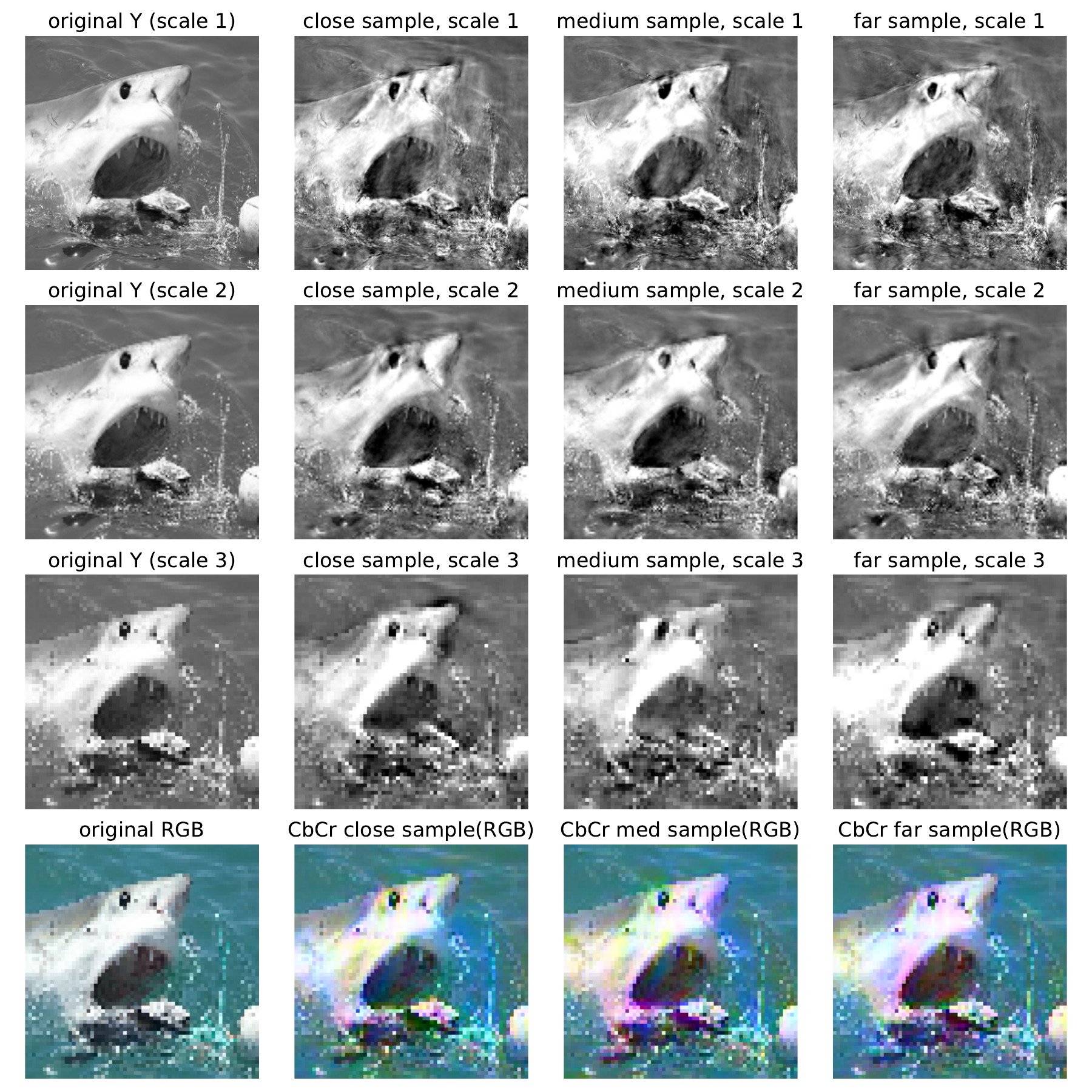}
        \caption{Example sample with far, close, medium samples(no mean added)}
        \label{fig:sample_mean2}
    \end{subfigure}

    \caption{Samples from SUSSBase. These samples are generated only from the learned covariance component of each SUPNs, \textbf{without the mean added}, in order to highlight differences in the learned structural patterns compared to other model variants.}

    \label{fig:samples}
\end{figure*}
 \clearpage
\section{Kadid10k Results}
\label{app:kadid}
\begin{figure*}[t]
    \centering
    \includegraphics[width=1\textwidth]{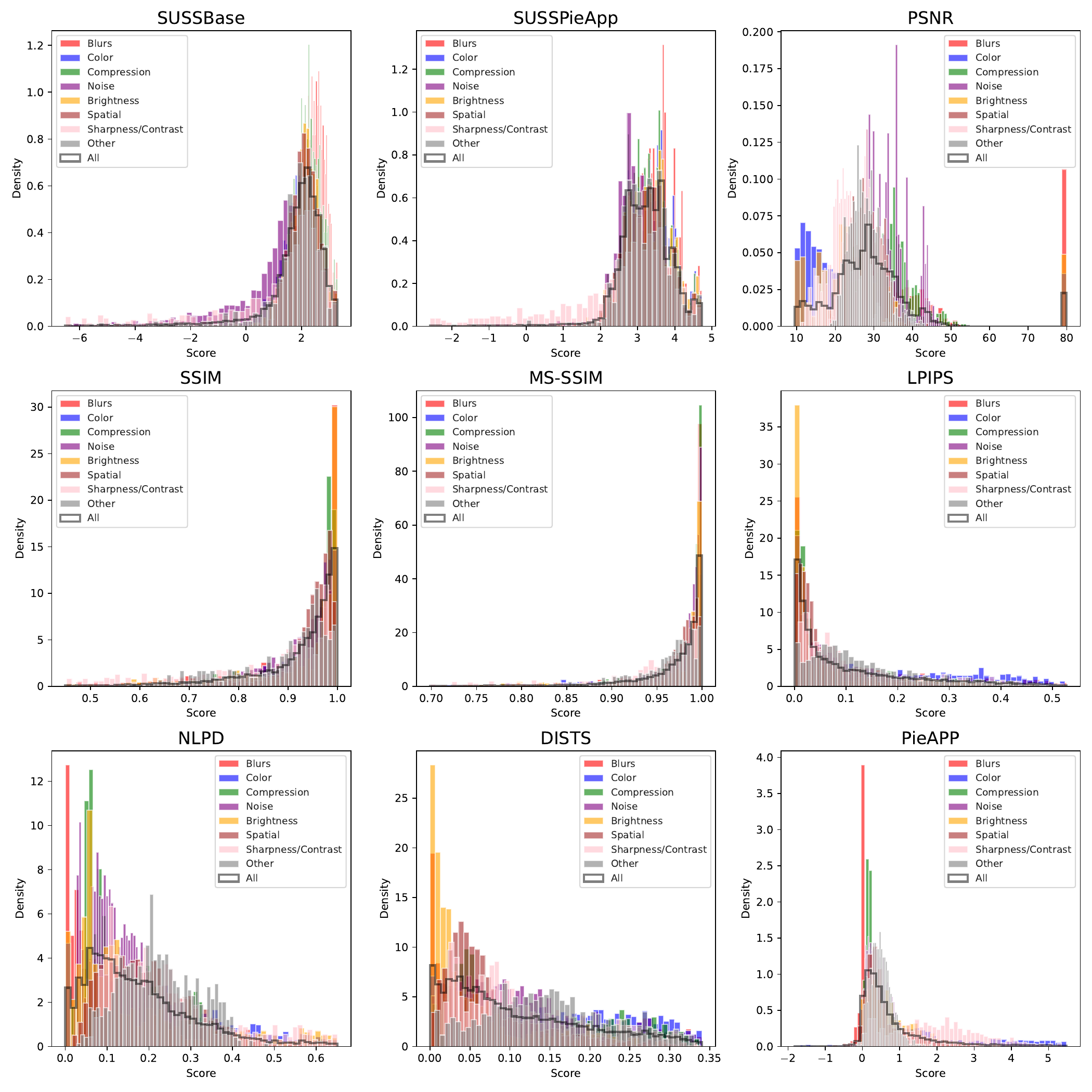}
    \caption{Distributions of values for the KADID-10k dataset for all images with a MOS of 4-5, corresponding to imperceptible to almost imperceptible differences. These distributions show score values for different distortion types across various metrics at this level.}
    \label{fig:kad1}
\end{figure*}

\begin{figure*}[t]
    \centering
    \includegraphics[width=1\textwidth]{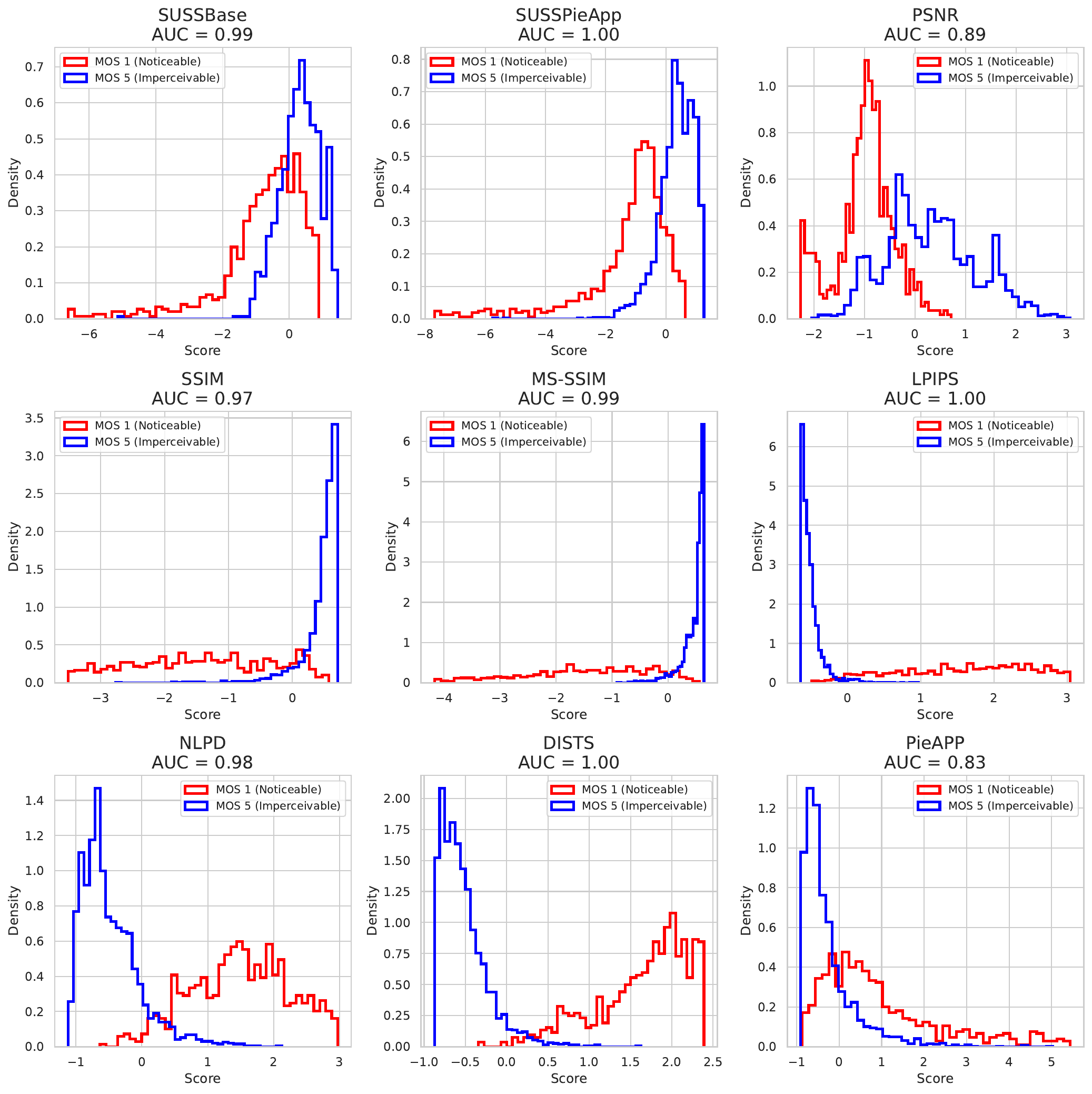}
    \caption{Distributions of values for the KADID-10k dataset for all images labeled as "imperceptible differences"(blue line) with highest MOS similarity scores, as well as "clearly noticeable differences" with lowest MOS scores(red line) for different metrics. AUC scores quantify the metric's ability to separate between these MOS groups. Normilised to enable comparison across metrics.}
    \label{fig:kad2}
\end{figure*}
 \clearpage
\section{Perceptual Space Results(Tranformed Validation Image Net)}
\label{app:res}

\begin{figure*}[t]
    \begin{minipage}{\textwidth}
        \centering
        \begin{subfigure}[b]{0.44\textwidth}
            \centering
            \includegraphics[width=\textwidth]{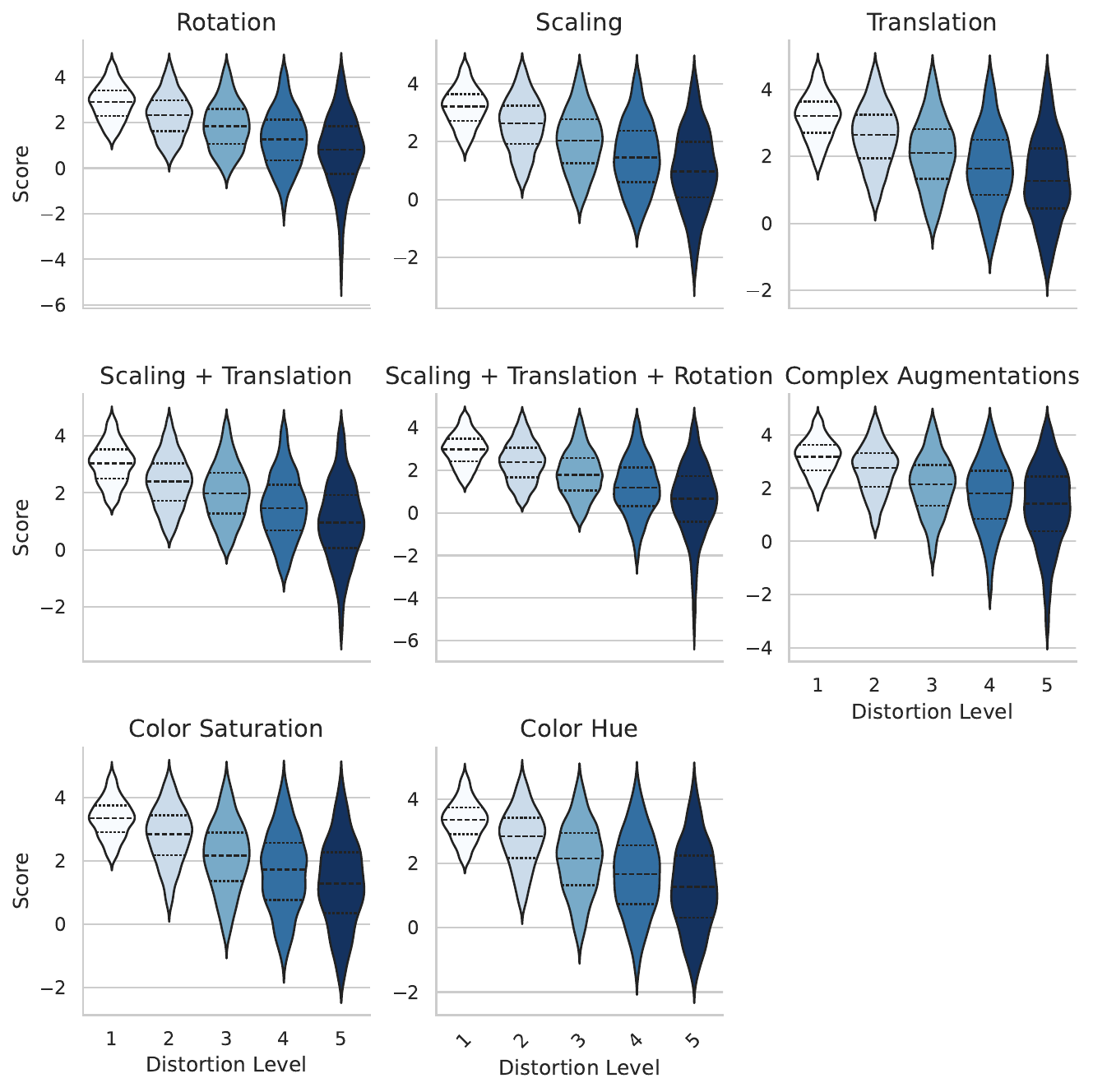}
            \caption{SUSSPieApp-RH results}
        \end{subfigure}
        \begin{subfigure}[b]{0.44\textwidth}
            \centering
            \includegraphics[width=\textwidth]{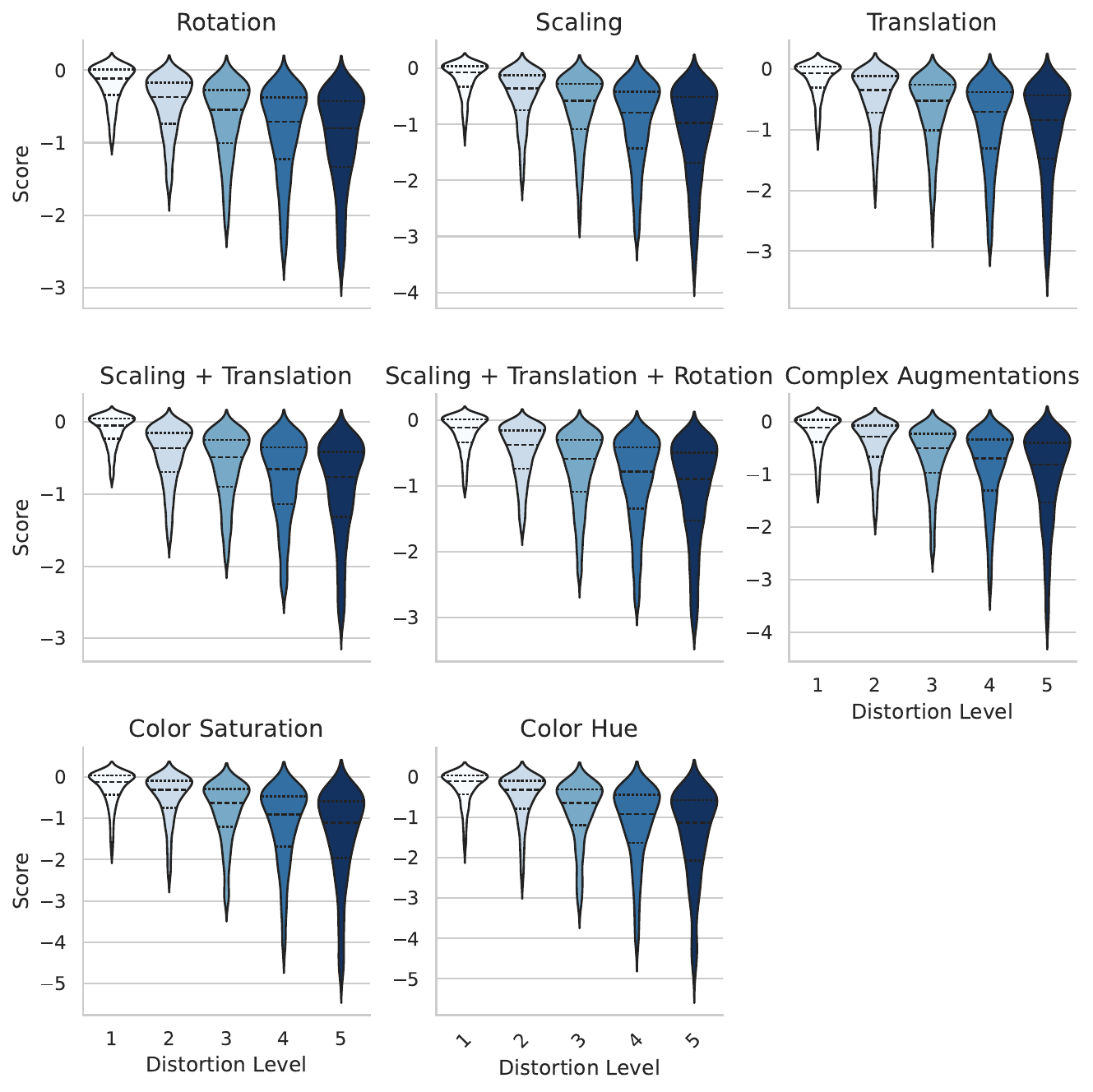}
            \caption{SUSSBase results}
        \end{subfigure}
        \begin{subfigure}[b]{0.33\textwidth}
            \centering
            \includegraphics[width=\textwidth]{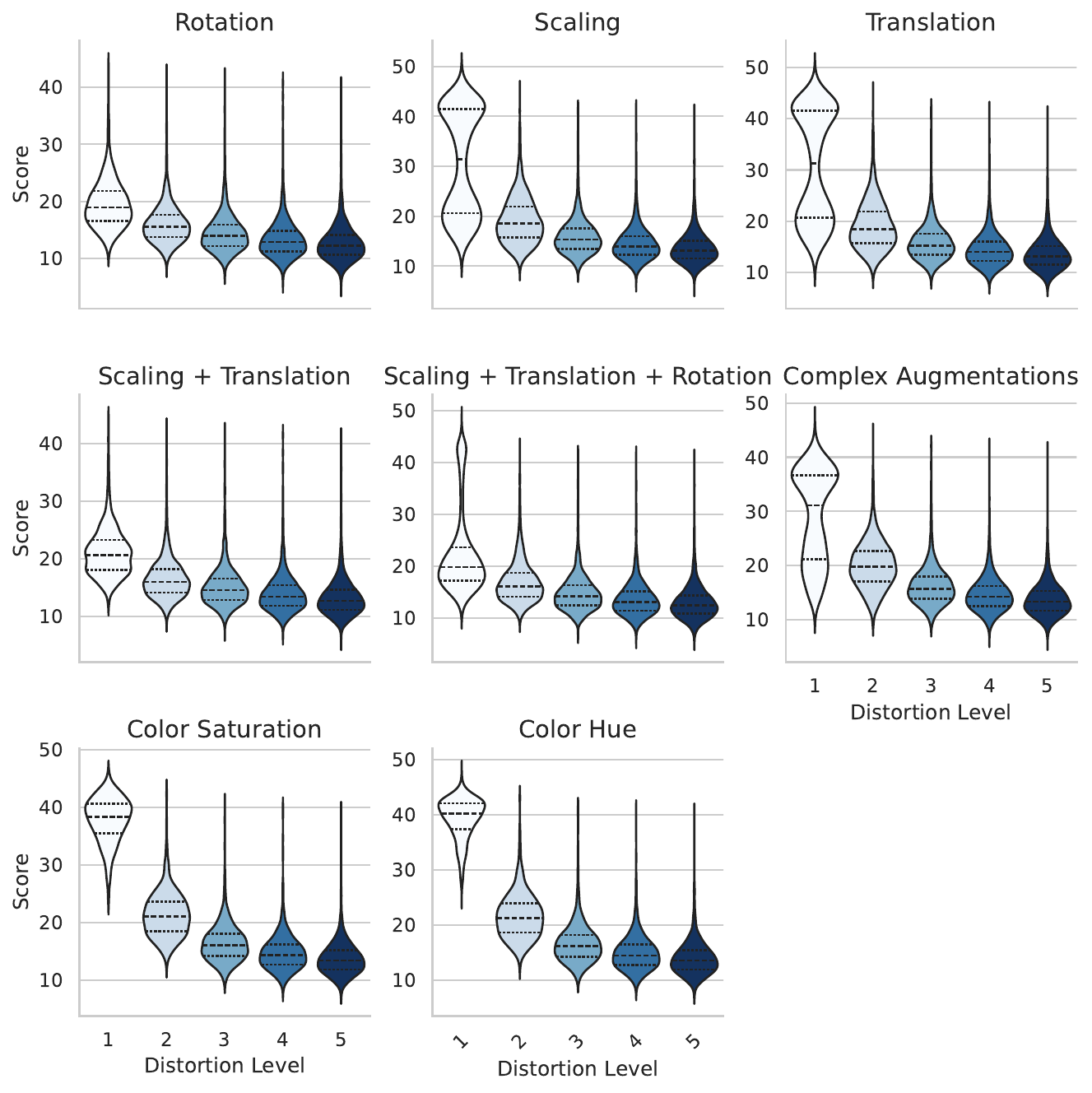}
            \caption{PSNR}
        \end{subfigure}
        \begin{subfigure}[b]{0.33\textwidth}
            \centering
            \includegraphics[width=\textwidth]{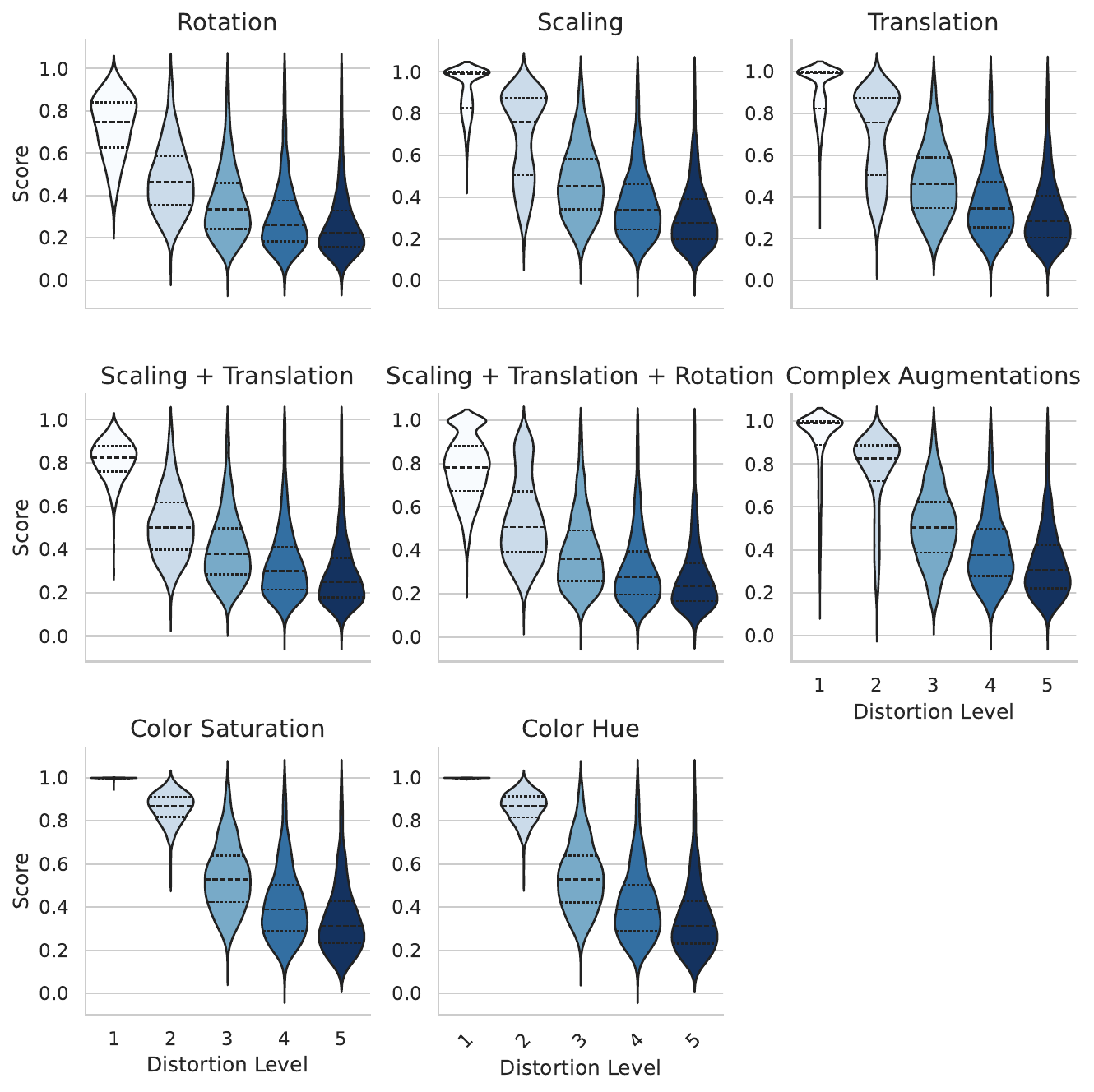}
            \caption{MSSIM}
        \end{subfigure}
        \begin{subfigure}[b]{0.33\textwidth}
            \centering
            \includegraphics[width=\textwidth]{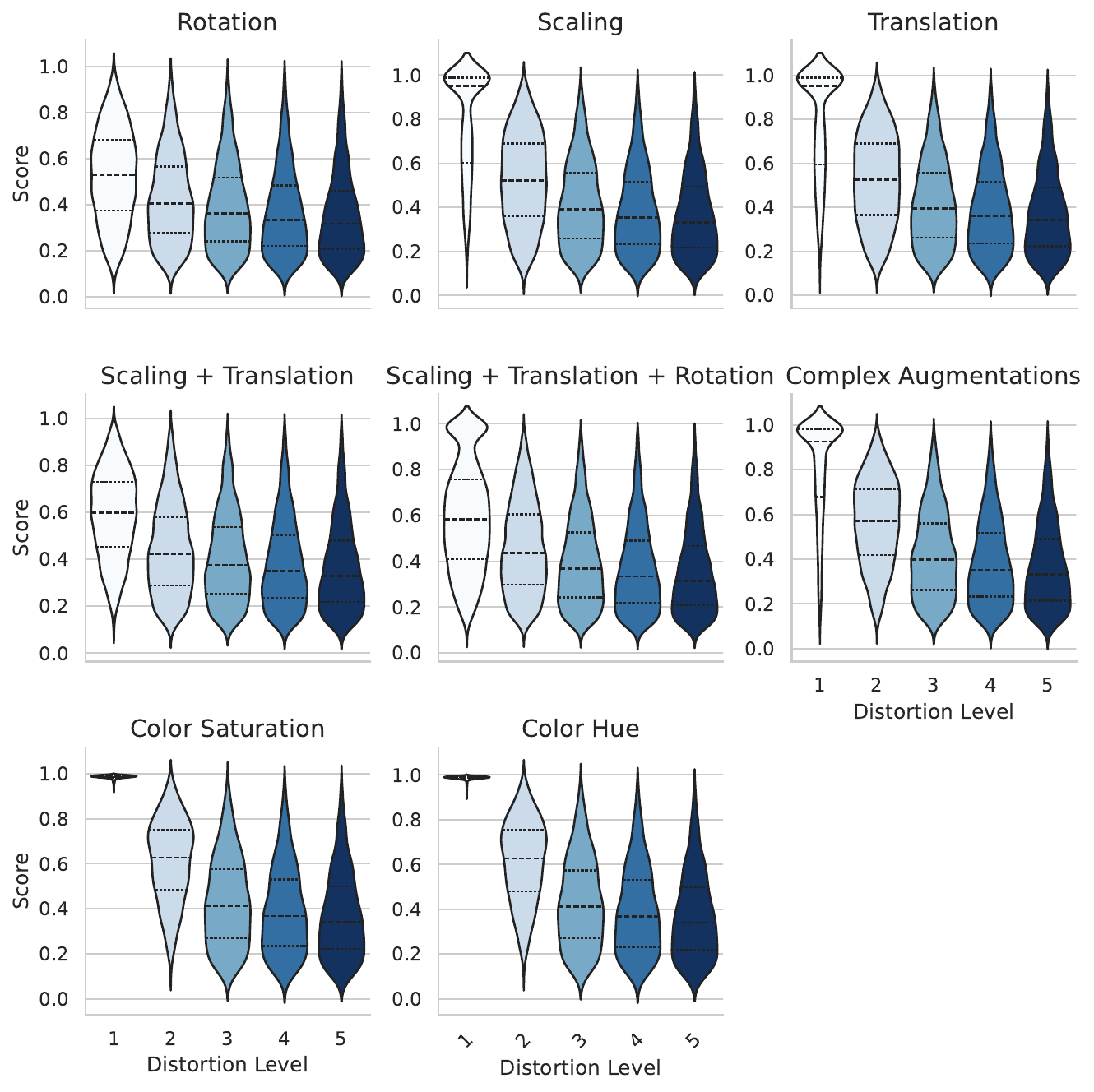}
            \caption{SSIM}
        \end{subfigure}
        \begin{subfigure}[b]{0.33\textwidth}
            \centering
            \includegraphics[width=\textwidth]{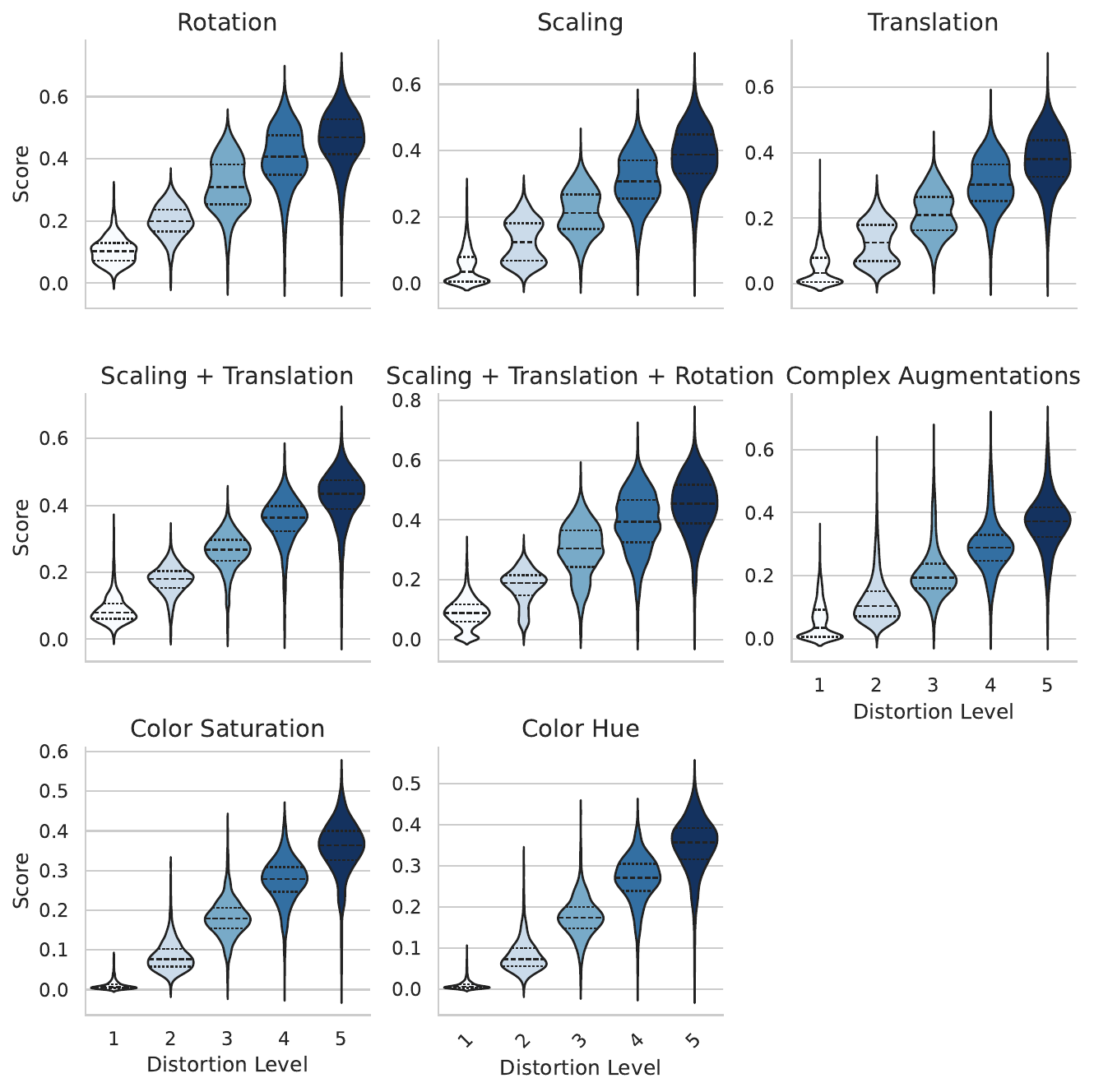}
            \caption{LPIPS}
        \end{subfigure}
        \begin{subfigure}[b]{0.33\textwidth}
            \centering
            \includegraphics[width=\textwidth]{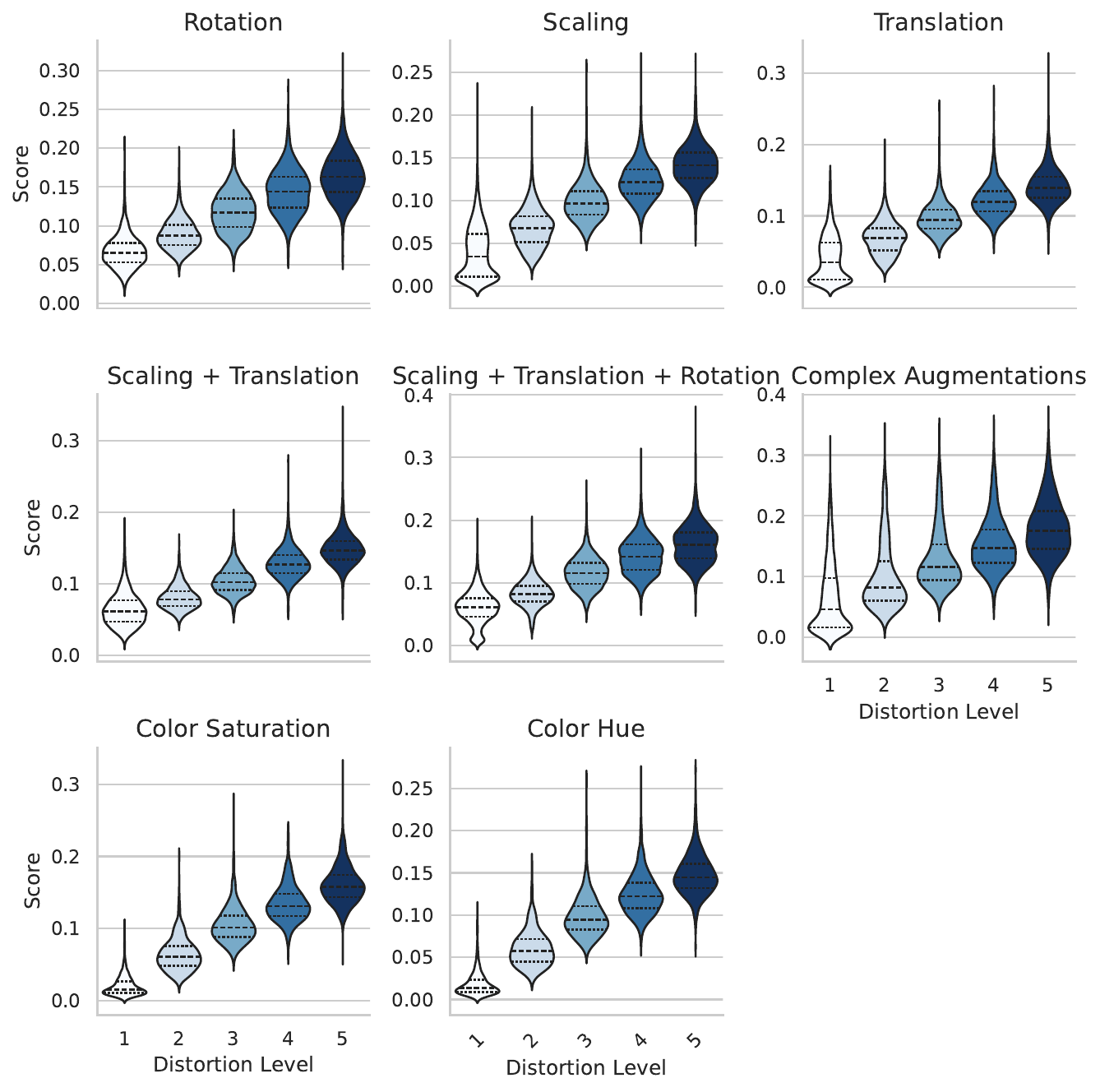}
            \caption{DIST}
        \end{subfigure}
    \end{minipage}
    \caption{Results on ImageNet transformed validation dataset with small distortions, showing how the different metrics perform across different transformation types. The results indicate a general linear increase in distortion, with some variation in response between metrics.}
    \label{fig:baaps_trans}
\end{figure*}
\begin{figure*}[t]
    \centering
    \begin{minipage}{\textwidth}
        \centering
        \begin{subfigure}[b]{0.44\textwidth}
            \centering
            \includegraphics[width=\textwidth]{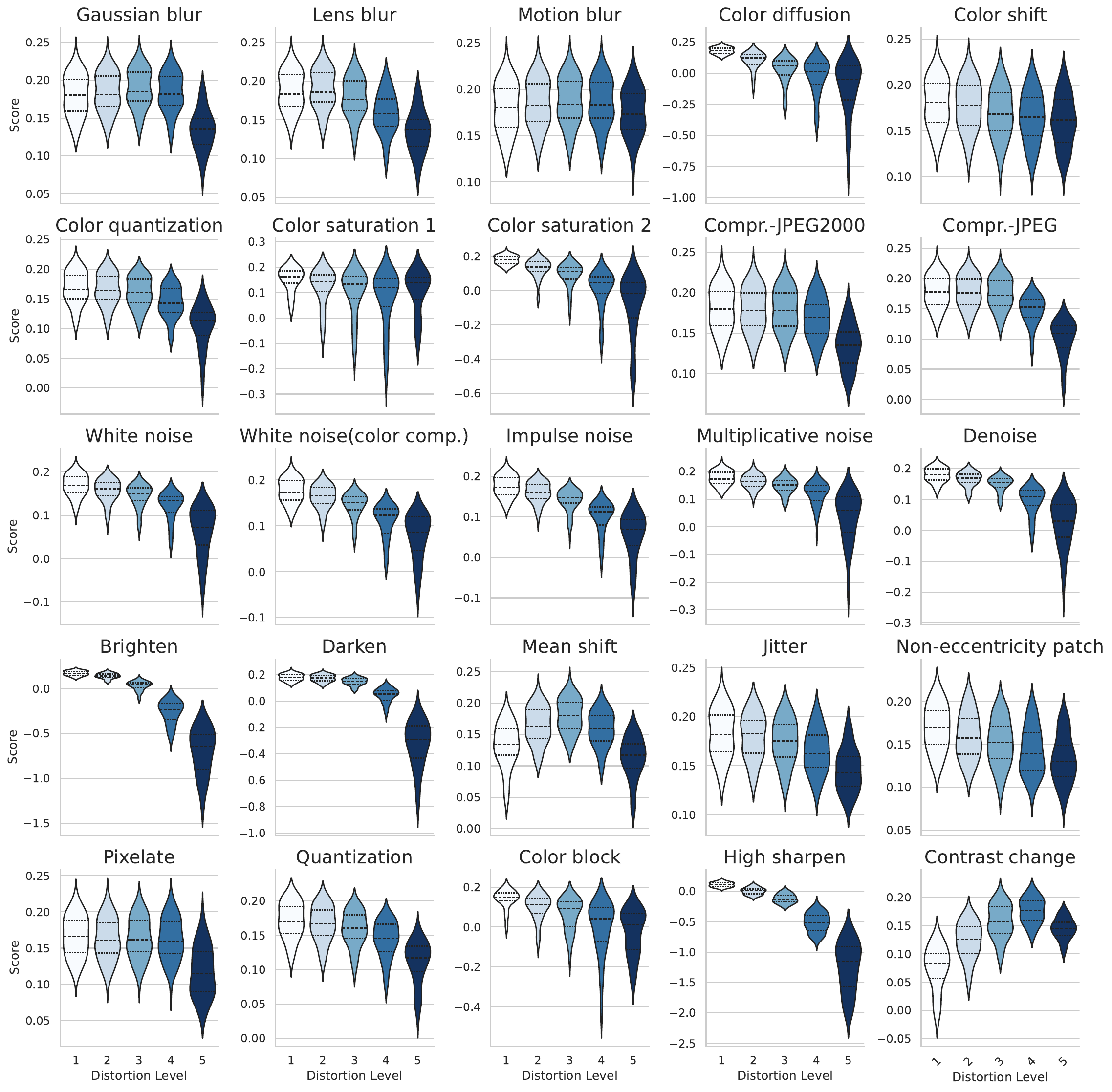}
            \caption{SUSSPieApp-RH Results}
        \end{subfigure}
        \begin{subfigure}[b]{0.44\textwidth}
            \centering
            \includegraphics[width=\textwidth]{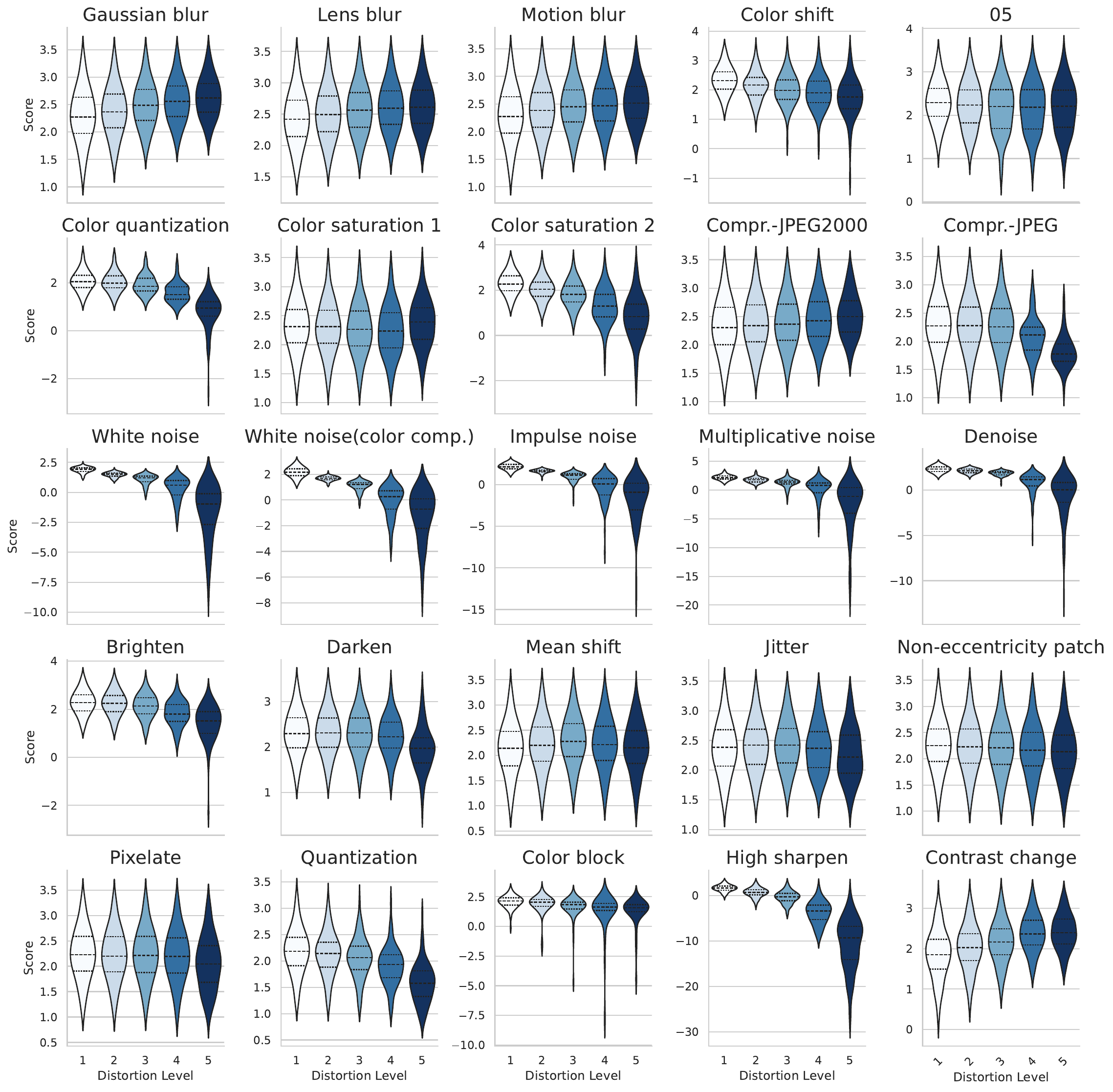}
            \caption{SUSSBase Results}
        \end{subfigure}
        \begin{subfigure}[b]{0.33\textwidth}
            \centering
            \includegraphics[width=\textwidth]{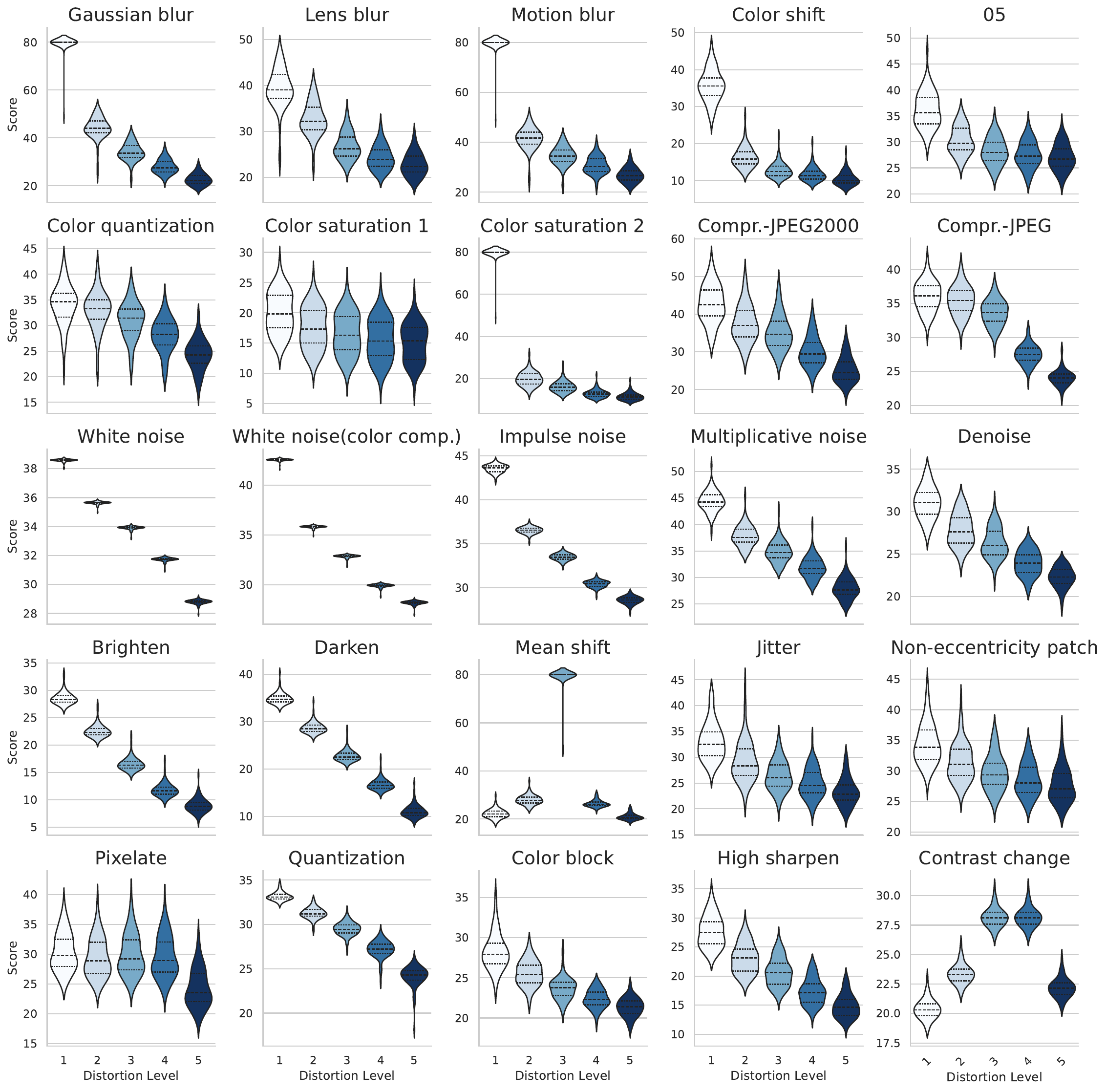}
            \caption{PSNR}
        \end{subfigure}
        \begin{subfigure}[b]{0.33\textwidth}
            \centering
            \includegraphics[width=\textwidth]{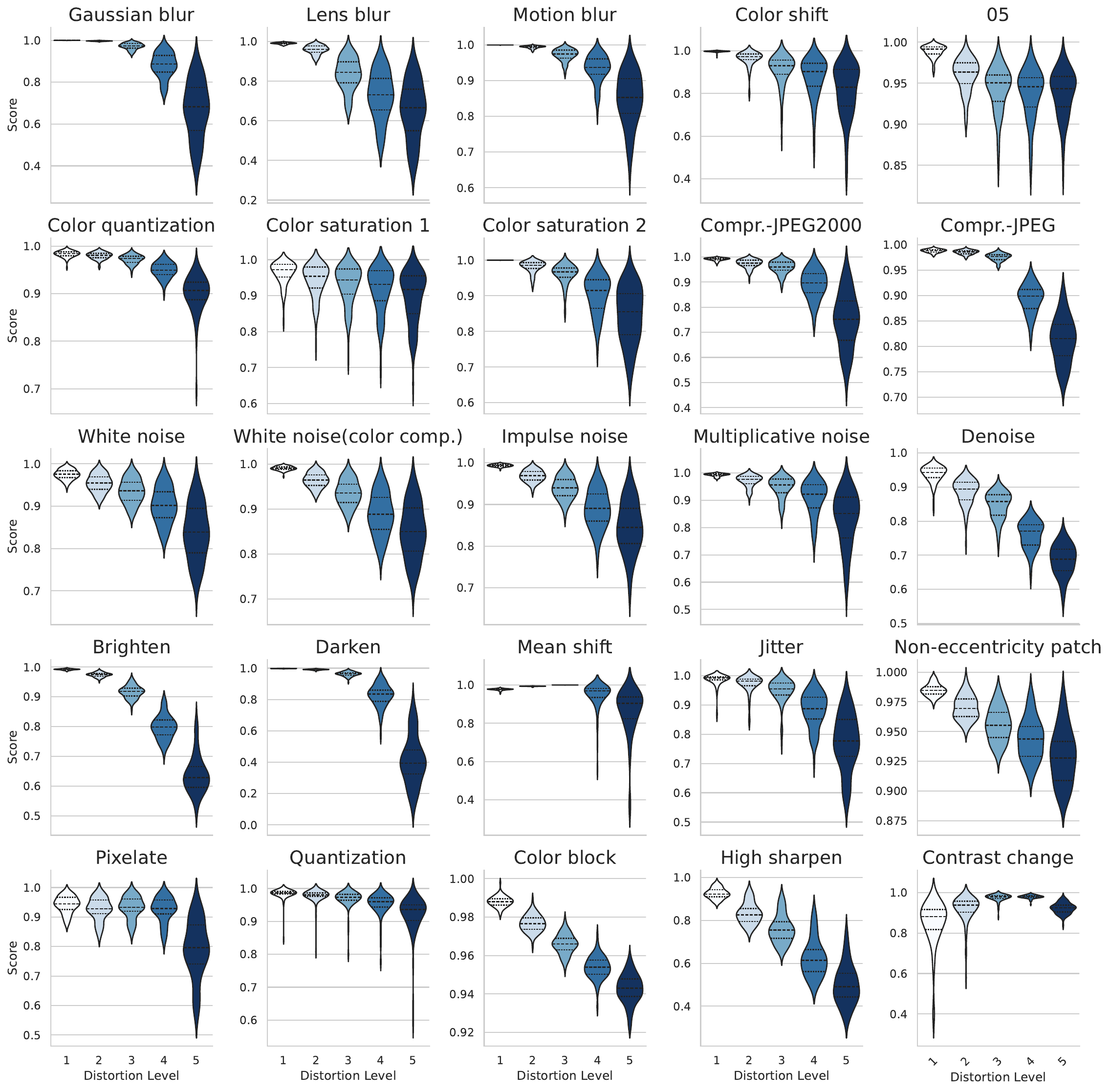}
            \caption{SSIM}
        \end{subfigure}
        \begin{subfigure}[b]{0.33\textwidth}
            \centering
            \includegraphics[width=\textwidth]{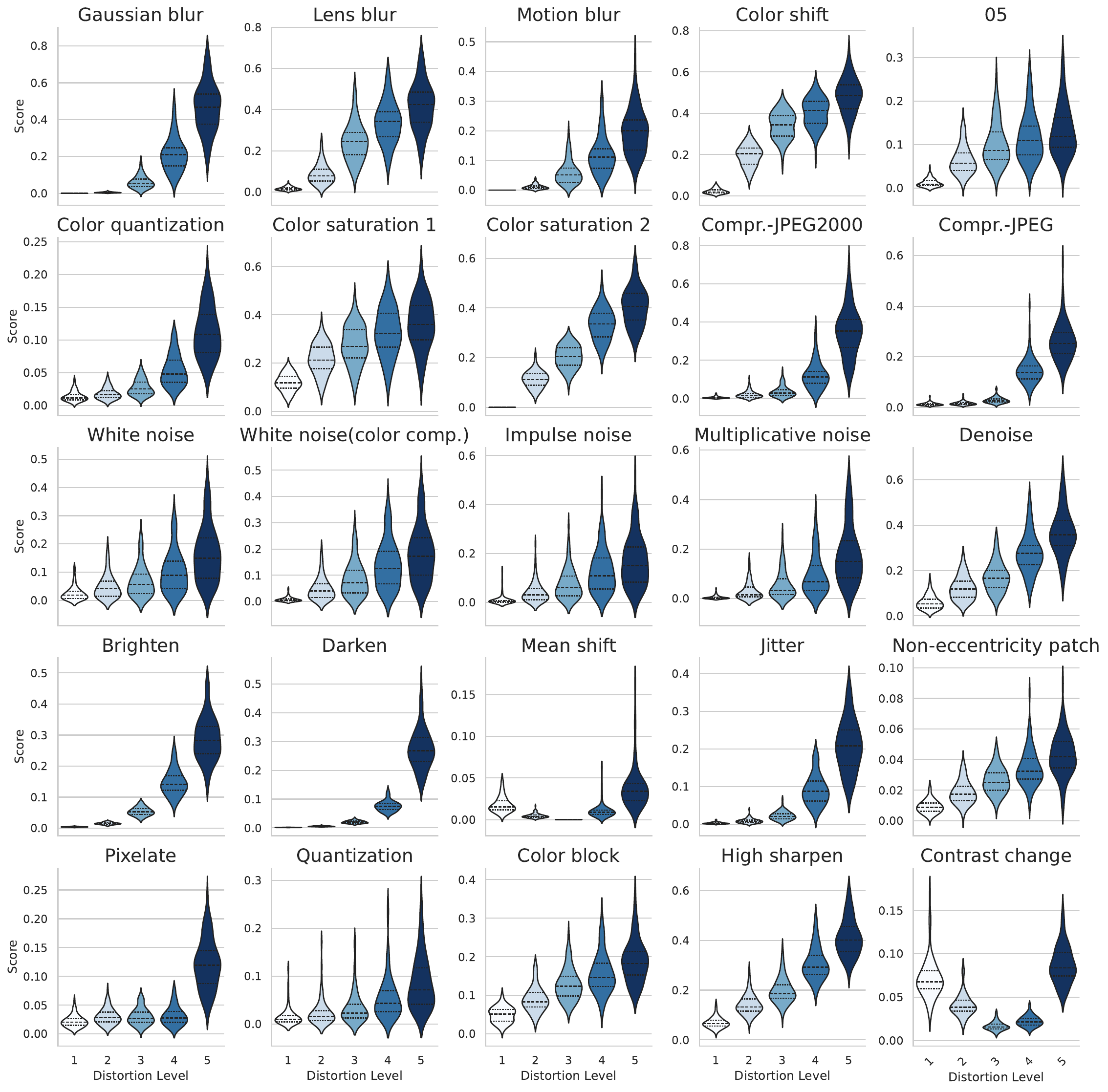}
            \caption{LPIPS}
        \end{subfigure}
        \begin{subfigure}[b]{0.33\textwidth}
            \centering
            \includegraphics[width=\textwidth]{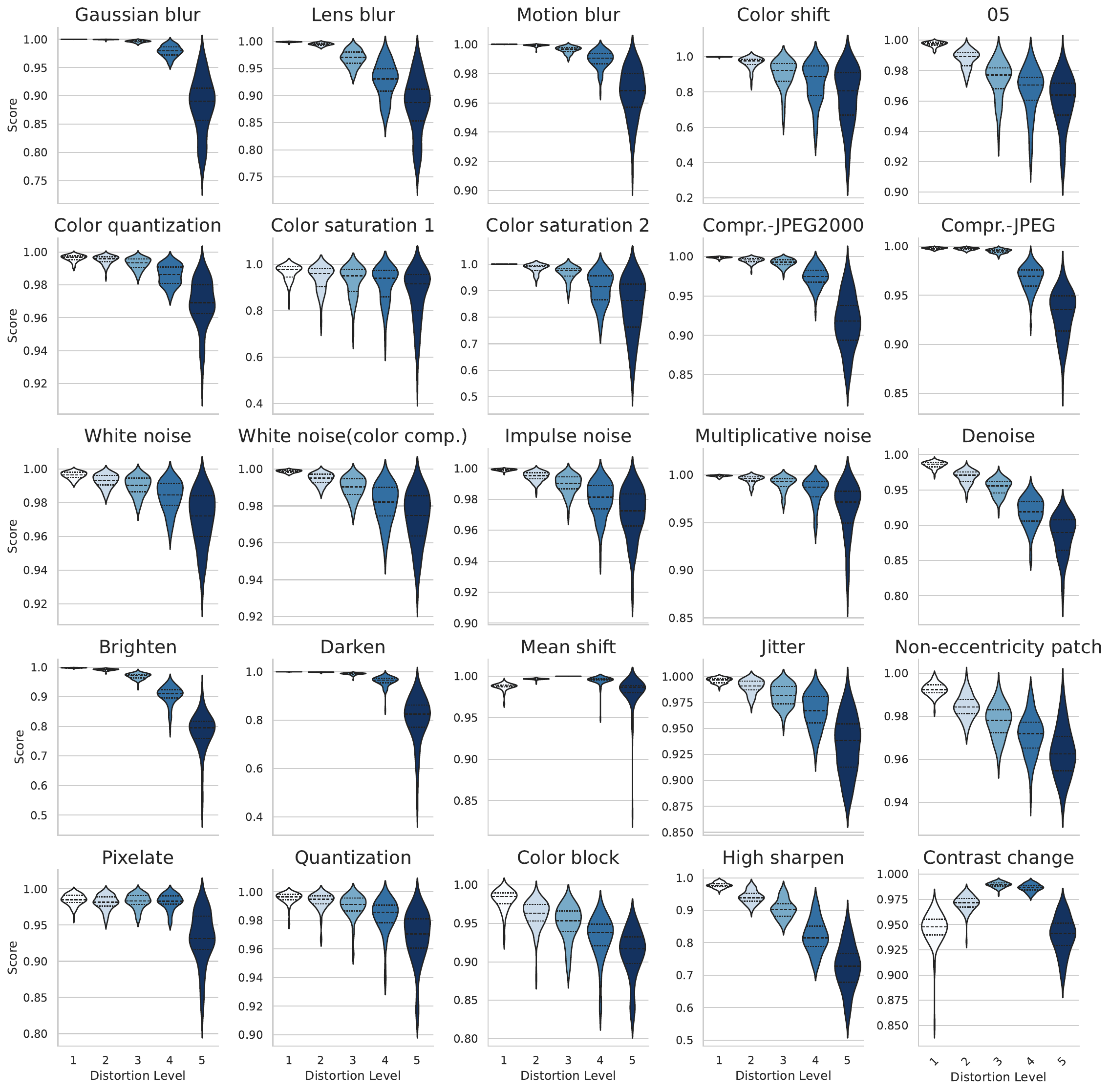}
            \caption{MS-SSIM}
        \end{subfigure}
        \begin{subfigure}[b]{0.33\textwidth}
            \centering
            \includegraphics[width=\textwidth]{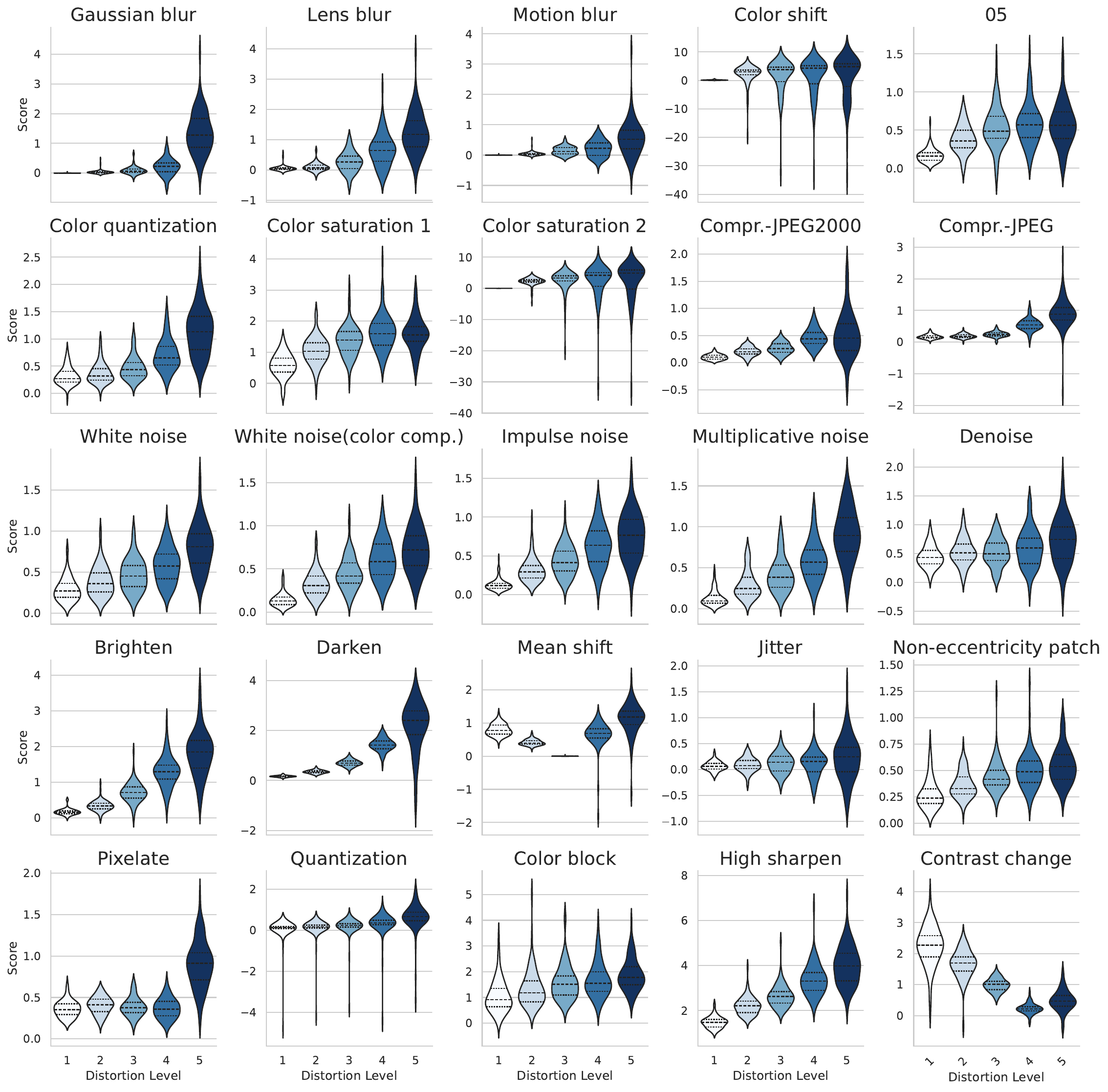}
            \caption{PieApp}
        \end{subfigure}
    \end{minipage}
    \caption{Results on the KADID dataset across various distortion types. The results show the expected linear progression of metric values as distortion increases, with some variation depending on the metric.}
    \label{fig:kandid}
\end{figure*}

 \clearpage
\section{Super Resolution}
\label{App:sr}
\begin{figure*}[t]
    \centering
    \includegraphics[width=\textwidth]{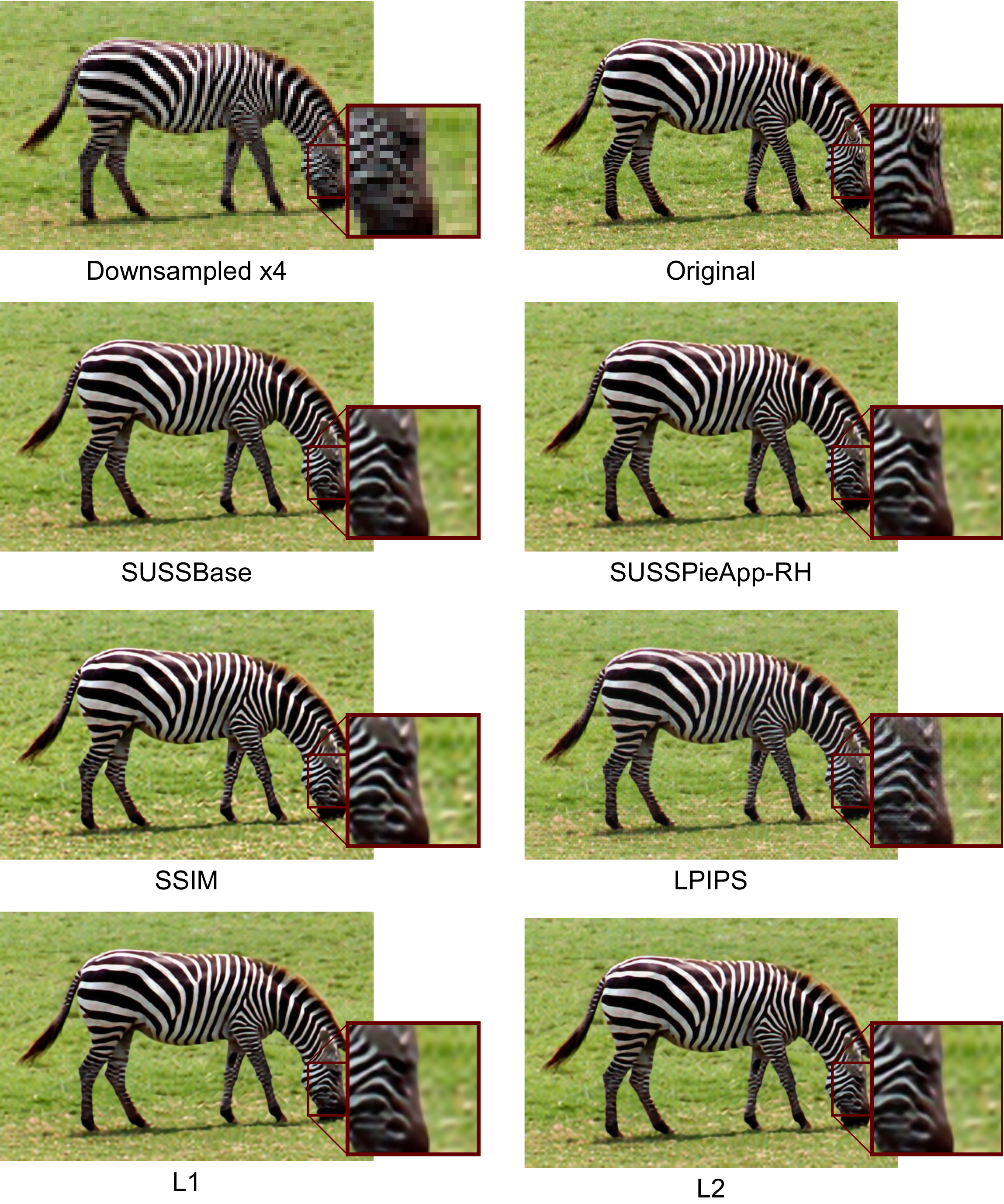}
    \caption{SR results for "zebra example" from test dataset for models finetuned on different objective functions.}
    \label{fig:sr1}
\end{figure*}
\begin{figure*}[t]
    \centering
    \includegraphics[width=0.8\textwidth]{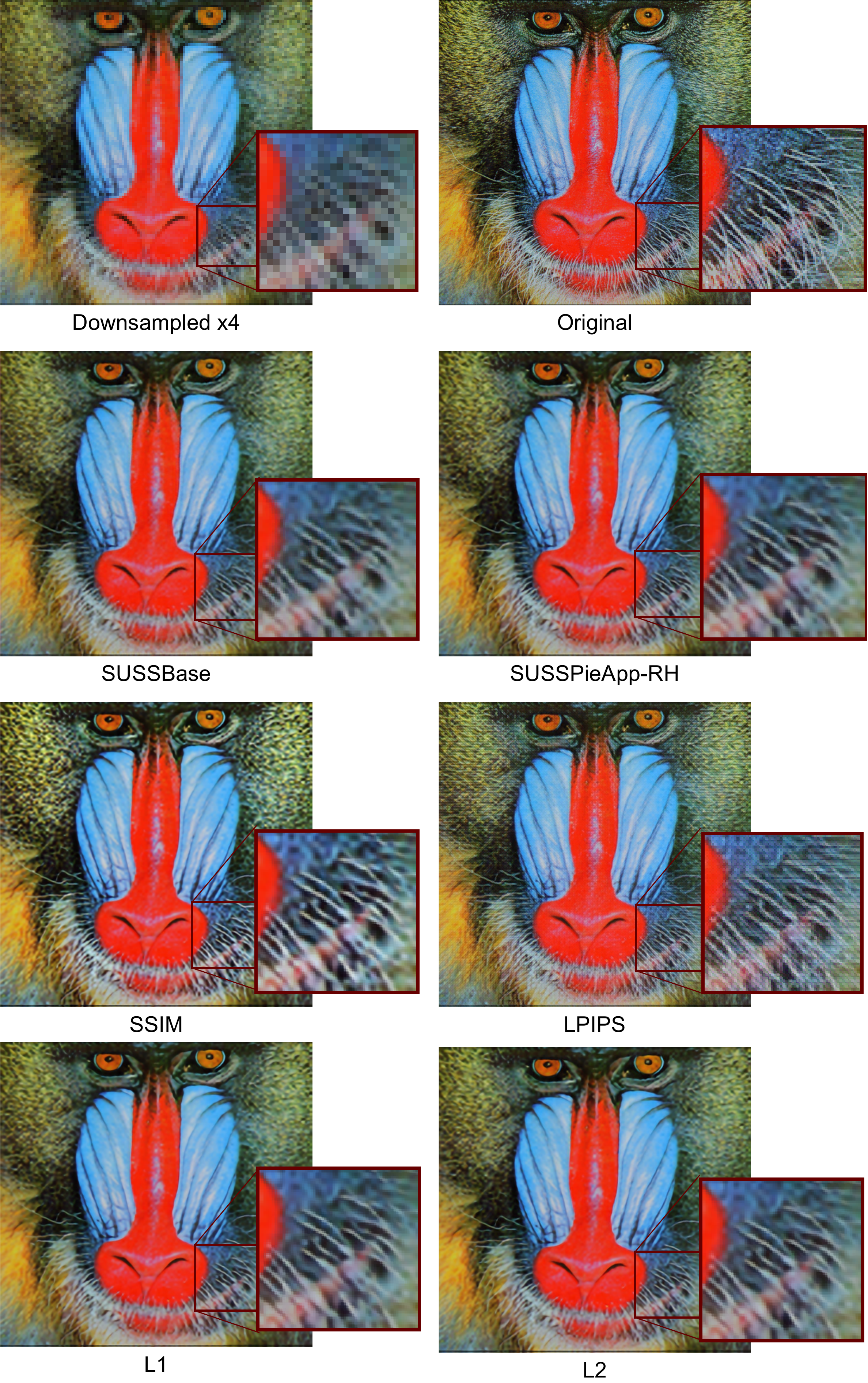}
    \caption{SR results for "baboon example" from test dataset for models finetuned on different objective functions.}
    \label{fig:sr1}
\end{figure*}
\begin{figure*}[t]
    \centering
    \includegraphics[width=\textwidth]{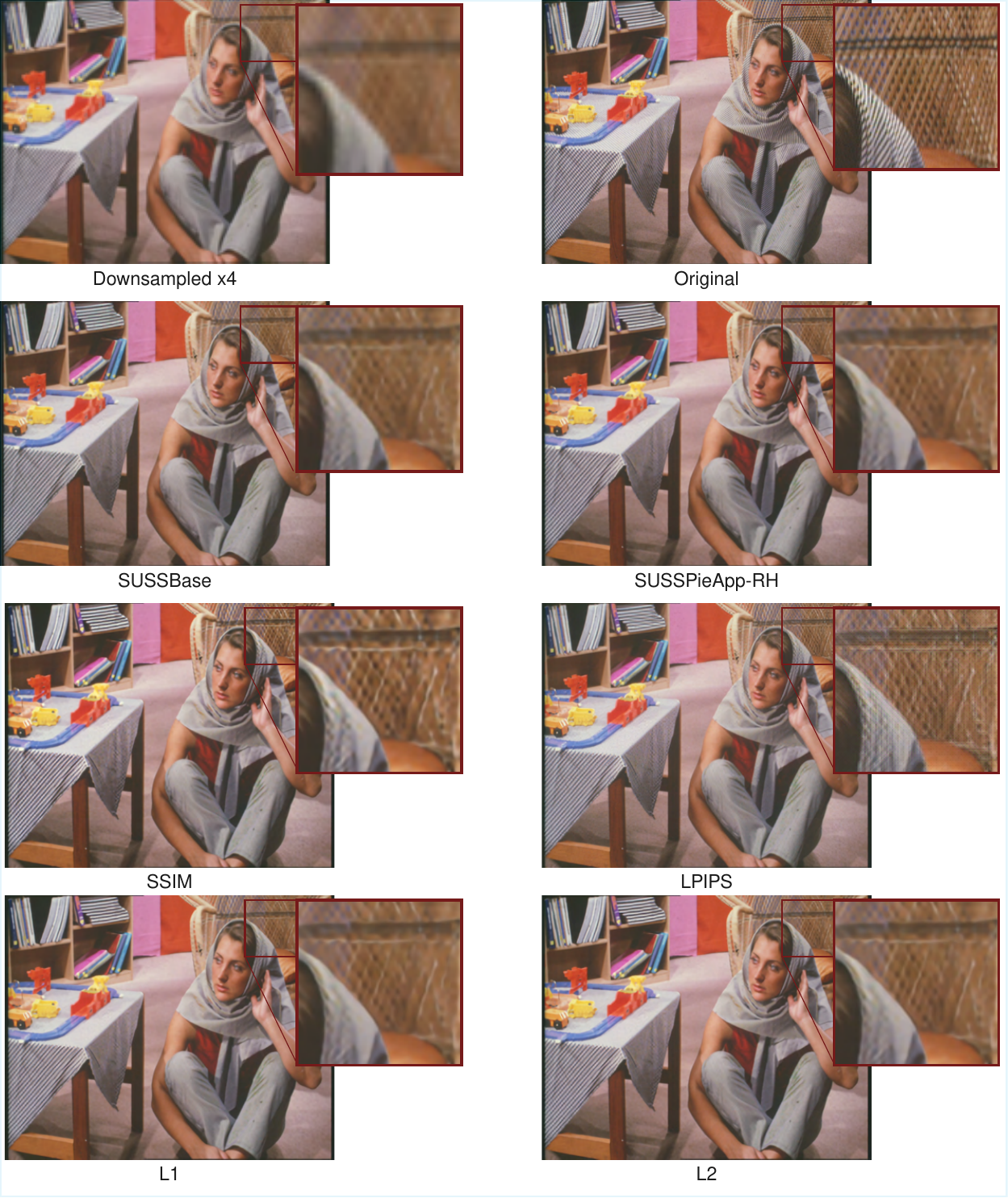}
    \caption{SR results of "Barbara example" from test dataset for models finetuned on different objective functions.}
    \label{fig:sr1}
\end{figure*}

 \clearpage
\section{Asymmetry results}
\label{App:assy}

Because of its formulation, SUSS is inherently asymmetric. We quantify this asymmetry by running an experiment on the KADID-10k dataset: for every reference image - distortion pair we compute SUSS(\textit{ref}, \textit{distorted(ref)}) and the reversed SUSS(\textit{distorted(ref)}, \textit{ref}). We measure asymmetry both in terms of absolute score difference and ranking consistency, using Pearson and Spearman correlation between the two score sets.

These are the results:
\begin{table}[h!]
\centering
\begin{tabular}{lccc}
\toprule
SUSS Model & Mean Abs. Asym. $\downarrow$ & Pearson $\uparrow$ & Spearman $\uparrow$ \\
\midrule
SUSSBase & 1.1 & 0.19 & 0.56 \\
SUSSPieApp-RH & 0.39 & 0.38 & 0.87 \\
\bottomrule
\end{tabular}
\caption{Quantitative asymmetry analysis of SUSS on KADID-10k.}
\end{table}

We observe that both models exhibit asymmetry, but to different degrees. The base model shows higher absolute asymmetry and weaker directional agreement, whereas the ranking/human-judgement fine-tuned variant (SUSSPieApp-RH) demonstrates significantly reduced magnitude asymmetry and strong rank consistency. The relatively high Spearman correlation indicates that, although absolute values change when reversing the input order, the relative ordering of image pairs remains largely preserved.

The effect of this asymmetry becomes clearer when examining SUSS maps of an example from KADID-10k, where there is some variation in the judgment (Figure~\ref{fig:asym1}). Moving from the sharp reference image to a blurred distorted version produces a different distribution of residual emphasis than the reverse direction. In particular, structural details reintroduced when comparing a blurry image to a sharper reference (e.g., ground textures in the lower region of Fig.~\ref{fig:asym1}b) are weighted differently than the loss of detail when going from sharp to blurry (Fig.~\ref{fig:asym1}a).

In practical downstream applications, where a fixed reference image is almost always available, this directional bias is unlikely to be problematic. Nevertheless, SUSS could be trivially symmetrised by averaging both directions:
\[
SUSS_{sym}(A,B) = \frac{1}{2}\big(SUSS(A,B) + SUSS(B,A)\big),
\]
which remains an interesting direction for future work.

Following shows example SUSS maps for an asymmetrical KADID-10k image pair:
\begin{figure*}[t]
    \begin{subfigure}[b]{\textwidth}
        \includegraphics[width=\textwidth]{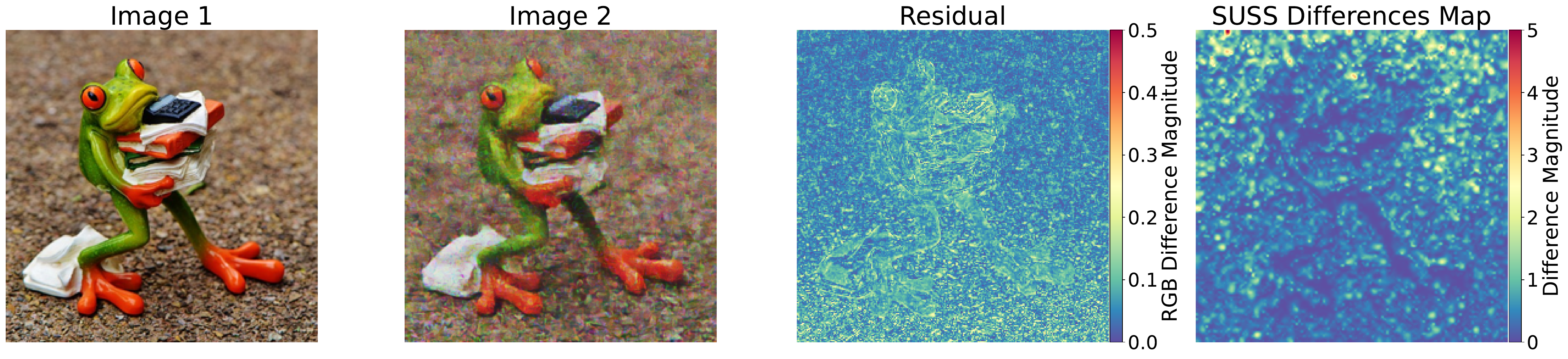}
        \caption{SUSS(\textit{Img 1}, T(\textit{Img 1})) = 2.2486}
    \end{subfigure}
    \begin{subfigure}[b]{\textwidth}
        \centering
        \includegraphics[width=\textwidth]{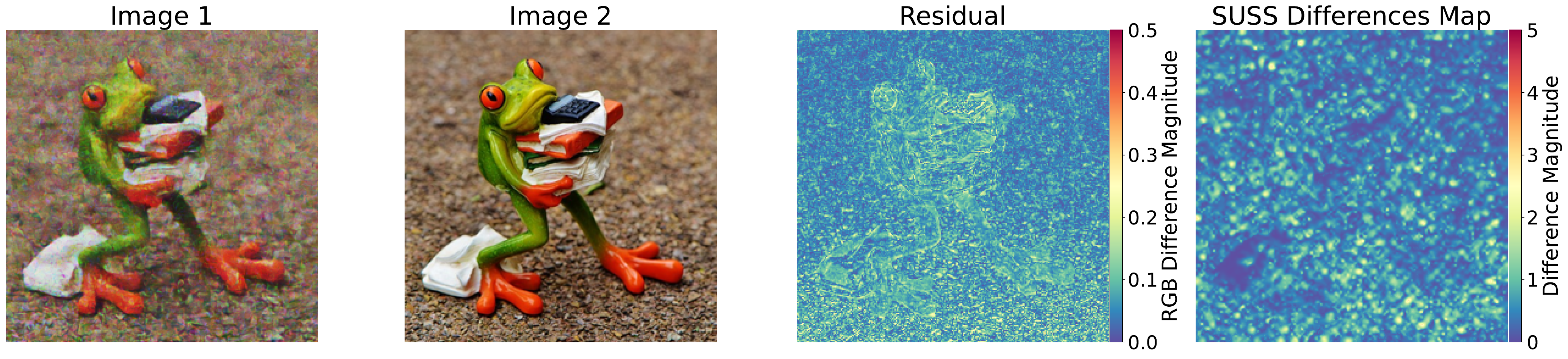}
        \caption{SUSS(T(\textit{Img 1}), \textit{Img 1}) = 1.6967}
    \end{subfigure}

    \caption{Example illustrating the asymmetry of SUSS. Reversing the reference and transformed images yields different scores. The residual maps show that the two directions emphasize different structural changes: details introduced when comparing a blurry input to a sharper, more detailed reference (e.g., ground structures at the bottom in (b)) are weighted differently than the loss of detail when going from sharp to blurry ((a)).}
    \label{fig:asym1}
\end{figure*}